%% file: main.tex
\documentclass[sigconf]{acmart}

\renewcommand\footnotetextcopyrightpermission[1]{} % removes footnote with conference information in first column
\setcopyright{none}

\input{00macros}

\begin{document}

%don't want date printed
\date{}

% make title bold and 14 pt font (Latex default is non-bold, 16 pt)
\title{On the Alignment of Group Fairness with Attribute Privacy}

\author{Jan Aalmoes}
\email{jan.aalmoes@insa-lyon.fr}
\affiliation{
  \institution{Univ Lyon, INSA Lyon, Inria, CITI}
  \city{Lyon}
  \country{France}
}
\author{Vasisht Duddu}
\email{vasisht.duddu@uwaterloo.ca}
\affiliation{%
 \institution{University of Waterloo}
 \city{Waterloo}
 \country{Canada}
}
\author{Antoine Boutet}
\email{antoine.boutet@insa-lyon.fr}
\affiliation{
  \institution{Univ Lyon, INSA Lyon, Inria, CITI}
  \city{Lyon}
  \country{France}
}

\begin{abstract}
\input{00abstract}
\end{abstract}

\maketitle
\pagestyle{plain}

\input{1introduction}
\input{2background}

\input{3problem}

\input{4setup} 
\input{5attacks}
\input{6egd}
\input{7advdebias}

\input{8related}

\input{9discussions}

\bibliographystyle{plain}
\bibliography{paper.bib}

\input{10appendix}

\end{document}

%% file: 00macros.tex
\usepackage{multirow}
\usepackage{enumitem}
\usepackage{algorithm}
\usepackage{colortbl}
\usepackage{algpseudocode}
\usepackage{algorithm}
\usepackage{dblfloatfix}
\usepackage{pifont}
\usepackage{amsthm}
\usepackage{graphicx}
\usepackage{url}
\usepackage{subfigure}
\usepackage{version}
\usepackage{xspace}
\usepackage{amsmath}
\usepackage{amsthm}
\usepackage{dsfont}
\usepackage{amsfonts}
\usepackage{tikz}
\usepackage[T1]{fontenc}
\usepackage{color, colortbl}
\usepackage{soul} %\st{striked text}
\newcommand\change[1]{{\color{purple}{#1}\xspace}}

\usetikzlibrary{shapes,arrows}
\usetikzlibrary{backgrounds, positioning, fit}
\usetikzlibrary{shapes,arrows}

\theoremstyle{definition}
\newtheorem{definition}{Definition}[section]
\newtheorem{theorem}{Theorem}[section]

\newcommand{\egd}{\textsc{EGD}\xspace}
\newcommand{\advdebias}{\textsc{AdvDebias}\xspace}
\newcommand{\targetmodel}{f_{trg}\xspace}
\newcommand{\attackmodel}{f_{att}\xspace}
\newcommand{\traindata}{\mathcal{D}_{tr}\xspace}
\newcommand{\testdata}{\mathcal{D}_{te}\xspace}
\newcommand{\auxdata}{\mathcal{D}_{aux}\xspace}
\newcommand{\auxtraindata}{\mathcal{D}_{aux}^{tr}\xspace}
\newcommand{\auxtestdata}{\mathcal{D}_{aux}^{te}\xspace}
\newcommand{\egdeo}{\textsc{EGD} + \textsc{EqOdds}\xspace}
\newcommand{\egddp}{\textsc{EGD} + \textsc{DemPar}\xspace}
\newcommand{\aia}{\textsc{AIA}\xspace}
\newcommand{\mia}{\textsc{MIA}\xspace}
\newcommand{\eo}{\textsc{EqOdds}\xspace}
\newcommand{\dempar}{\textsc{DemPar}\xspace}
\newcommand{\adaptiveAIA}{\textsc{AdaptAIA}\xspace}
\newcommand{\adaptiveAIAHard}{\textsc{AdaptAIA-H}\xspace}
\newcommand{\adaptiveAIASoft}{\textsc{AdaptAIA-S}\xspace}
\newcommand{\race}{\texttt{Race}\xspace}
\newcommand{\sex}{\texttt{Sex}\xspace}
\newcommand{\census}{\texttt{CENSUS}\xspace}
\newcommand{\baseline}{\texttt{Baseline}\xspace}
\newcommand{\theoretical}{\texttt{Theoretical}\xspace}
\newcommand{\empirical}{\texttt{Empirical}\xspace}
\newcommand{\meps}{\texttt{MEPS}\xspace}

\newcommand{\lfw}{\texttt{LFW}\xspace}
\newcommand{\compas}{\texttt{COMPAS}\xspace}

\newcommand{\adv}{\protect{$\mathcal{A}dv$}\xspace}

\makeatletter
\tikzset{
    database/.style={
        path picture={
            \draw (0, 1.5*\database@segmentheight) circle [x radius=\database@radius,y radius=\database@aspectratio*\database@radius];
            \draw (-\database@radius, 0.5*\database@segmentheight) arc [start angle=180,end angle=360,x radius=\database@radius, y radius=\database@aspectratio*\database@radius];
            \draw (-\database@radius,-0.5*\database@segmentheight) arc [start angle=180,end angle=360,x radius=\database@radius, y radius=\database@aspectratio*\database@radius];
            \draw (-\database@radius,1.5*\database@segmentheight) -- ++(0,-3*\database@segmentheight) arc [start angle=180,end angle=360,x radius=\database@radius, y radius=\database@aspectratio*\database@radius] -- ++(0,3*\database@segmentheight);
        },
        minimum width=2*\database@radius + \pgflinewidth,
        minimum height=3*\database@segmentheight + 2*\database@aspectratio*\database@radius + \pgflinewidth,
    },
    database segment height/.store in=\database@segmentheight,
    database radius/.store in=\database@radius,
    database aspect ratio/.store in=\database@aspectratio,
    database segment height=0.1cm,
    database radius=0.25cm,
    database aspect ratio=0.35,
}
\makeatother
\definecolor{LightCyan}{rgb}{0.88,1,1}

\includeversion{submit}
\excludeversion{topic}

\newtheorem{innercustomthm}{Theorem}
\newenvironment{customthm}[1]
  {\renewcommand\theinnercustomthm{#1}\innercustomthm}
  {\endinnercustomthm}

%% file: 00abstract.tex
Group fairness and privacy are fundamental aspects in designing trustworthy machine learning models. 
Previous research has highlighted conflicts between group fairness and different privacy notions. 
We are the \textit{first} to demonstrate the \textit{alignment} of group fairness with the specific privacy notion of attribute privacy in a \textit{blackbox setting}. 
Attribute privacy, quantified by the resistance to attribute inference attacks (\aia{s}), requires indistinguishability in the target model's output predictions. 
Group fairness guarantees this thereby mitigating \aia{s} and achieving attribute privacy.
To demonstrate this, we first introduce \adaptiveAIA, an enhancement of existing \aia{s}, tailored for real-world datasets with class imbalances in sensitive attributes. 
Through theoretical and extensive empirical analyses, we demonstrate the efficacy of two standard group fairness algorithms (i.e., adversarial debiasing and exponentiated gradient descent) against \adaptiveAIA. 
Additionally, since using group fairness results in attribute privacy, it acts as a defense against \aia{s}, which is currently lacking. Overall, we show that group fairness aligns with attribute privacy at no additional cost other than the already existing trade-off with model utility.  
% this last sentence sounds strange
% it is not free as there is a tradeoff between the utility, the privacy, and the fairness

%% file: 1introduction.tex
\section{Introduction}
\label{sec:introduction}

Machine learning (ML) has been adopted for several high-stakes decision-making applications, such as criminal justice and healthcare. This raises concerns about the model's discriminatory behaviour across different demographic subgroups~\cite{surveyfair} and compromising data privacy~\cite{de2020overview}. 
Several regulations require ML practitioners to train models satisfying both privacy and group fairness~\cite{ec2019ethics,nist,dpia,ico,whitehouse}.
To design such models, practitioners have to first understand the interactions between group fairness and privacy~\cite{duddu2023sok}.

Prior works have shown that group fairness conflicts with different privacy notions such as differential privacy~\cite{cummings,ijcai2022p766}, membership inference~\cite{chang2021privacy} and distribution inference~\cite{suri2023dissecting}. However, it is unclear about how group fairness interacts with privacy of sensitive attributes (hereafter referred to as attribute privacy) which ensures that an adversary cannot exploit distinguishability in target model's predictions to infer sensitive attributes. In this work, we ask: \textbf{\textit{How does group fairness relate to attribute privacy?}}
%\begin{quote}
%\small\textit{How does group fairness relate to attribute privacy?}   
%\end{quote}

\begin{figure}[!htb]
     \centering
     \includegraphics[width=1\linewidth]{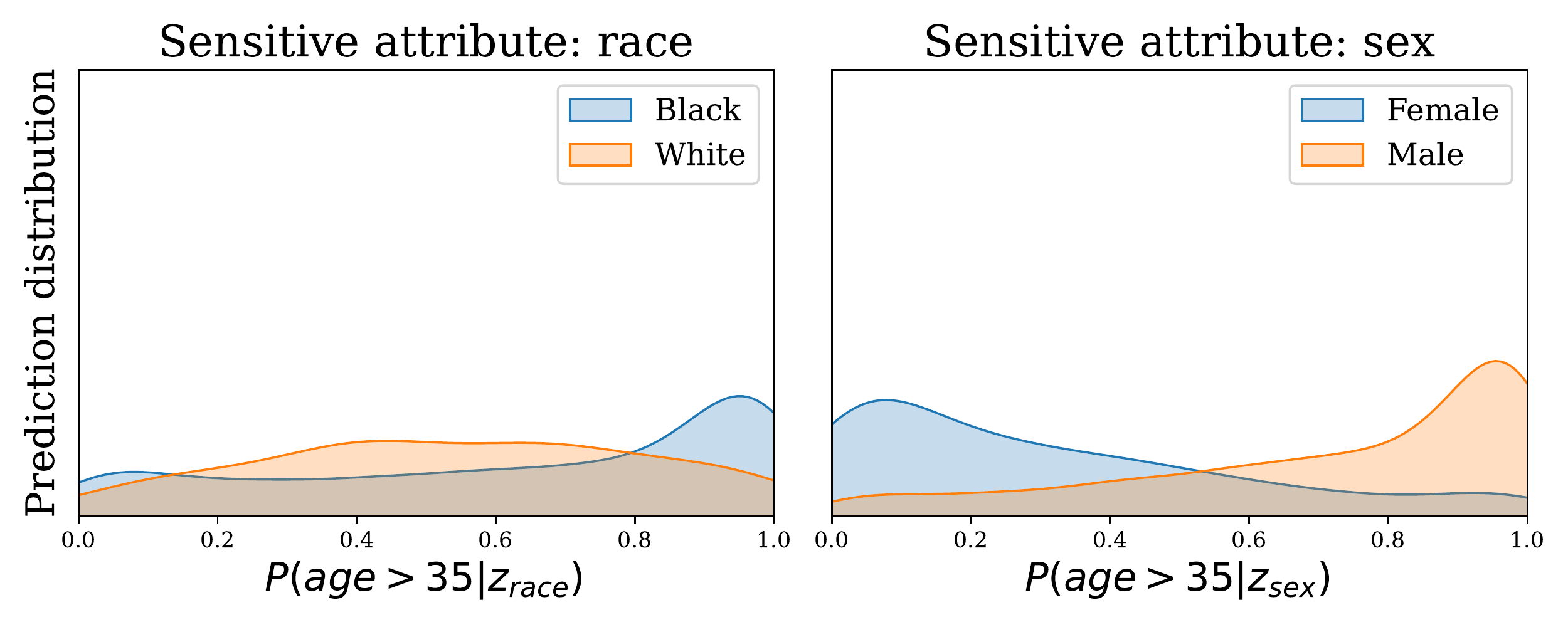}
     \caption{Privacy is violated if an  (potentially sensitive) attribute is inferred from the output predictions of a learning model if these predictions are distinguishable for different attribute values (Appendix~\ref{app:distinguishability}).}
     \label{intro}
\end{figure}

To answer this, we first quantify attribute privacy as the resistance to \textit{attribute inference attacks} (\aia{s}) where an adversary infers sensitive attributes (e.g., \race and \sex), from \textit{model observables} (e.g., predictions, intermediate outputs~\cite{fredrikson2,Mahajan2020DoesLS,yeom,Song2020Overlearning,malekzadeh2021honestbutcurious,MehnazAttInf}. These attacks pose a practical threat to data privacy and confidentiality. For example, inferring \race from a model predicting mortgage eligibility, even if not part of the training data, could violate customer privacy and potentially lead to discrimination.
We focus on representation-based \aia{s} in a practical \textit{blackbox} setting where an adversary exploits distinguishability in model observables for different sensitive attribute values~\cite{Song2020Overlearning,Mahajan2020DoesLS,malekzadeh2021honestbutcurious}. 
For instance, Figure~\ref{intro} depicts distinguishable prediction distributions for two sub-groups.

% \aia{s} are either \textit{imputation-based} (exploit non-sensitive attributes, predictions, and background information)~\cite{yeom,fredrikson2,MehnazAttInf,jayaraman2022attribute}, or \textit{representation-based} (exploit distinguishability in predictions or intermediate outputs for different sensitive attribute values~\cite{Song2020Overlearning,Mahajan2020DoesLS,malekzadeh2021honestbutcurious}. However, existing imputation-based attacks are similar to data imputation and do not pose a privacy risk~\cite{jayaraman2022attribute}.

Ensuring attribute privacy requires the success of \aia{s} to be close to random guessing, meaning model observables should be indistinguishable for different sensitive attribute values. Group fairness achieves this by ensuring indistinguishability in predictions.
Implementations of such group fairness include adversarial debiasing (\advdebias)~\cite{debiase,NIPS2017_48ab2f9b} and exponentiated gradient descent (\egd)~\cite{reductions}). Hence, we conjecture that group fairness is \textit{aligned} with attribute privacy and can be used as a defense against \aia{s}, which is currently lacking.

\noindent\textbf{\underline{Contributions.}} We are the \textit{first} to show that the objective of group fairness \textit{aligns} with attribute privacy. In other words, using group fairness results in attribute privacy at no additional cost except for the already existing trade-off with utility.
To demonstrate this, we first propose a state-of-the-art \aia, \adaptiveAIA, designed to infer sensitive attributes while accounting for class imbalance (Section~\ref{sec:adaptiveAIA}). Finally, we demonstrate, theoretically and through extensive empirical evaluation, the effectiveness of \egd and \advdebias against \adaptiveAIA (Sections~\ref{sec:EGD} and~\ref{sec:advdebias}). 
Our source code will be publicly available upon publication.
% add key results

%% file: 2background.tex
\section{Background}
\label{sec:background}

We present a background on ML and notations used in the rest of the paper (Section~\ref{back-ml}), followed by a discussion on \aia{s} (Section~\ref{sec:back-attribute}), and group fairness (Section~\ref{sec:back-fairness}).

\subsection{Machine Learning Classifiers}
\label{back-ml}

\input{New/tables/tab_summary}

\noindent\textbf{Training.} ML classifiers are functions $\targetmodel^{\theta}$ (omit $\theta$ for simplicity) parameterized by $\theta$ that map inputs with corresponding classification labels. $\theta$ is updated using a training dataset ($\traindata$) with the objective to minimize the loss incurred on predicting the classification label for inputs from $\traindata$. We remove sensitive attributes such as \race or \sex from $\traindata$ to censor them~\cite{Song2020Overlearning}. Consequently, $\targetmodel$ is trained on non-sensitive attributes.
Formally, consider a probability space $(\Omega, \mathcal{T}, \mathcal{P})$, measurable spaces $(E, \mathcal{U})$, $(\{0,1\},\mathcal{P}(\{0,1\}))$ and $([0,1], \mathcal{B})$ where $\mathcal{B}$ is the Borel tribe on $[0,1]$. We define random variables $X$ for the input data, $Y$ for the classification labels and $S$ for the sensitive attributes:
\begin{itemize}
    \item $X:(\Omega, \mathcal{T}, P) \longrightarrow (E, \mathcal{U}),$
    \item $Y:(\Omega, \mathcal{T}, P) \longrightarrow (\{0,1\},\mathcal{P}(\{0,1\})),$
    \item $S:(\Omega, \mathcal{T}, P) \longrightarrow (\{0,1\},\mathcal{P}(\{0,1\})),$
\end{itemize}
Then, $\targetmodel$ is a measurable function $\targetmodel:(E, \mathcal{U}) \longrightarrow ([0,1], \mathcal{B})$ which is used to build the statistic approaching $Y$ by updating the parameters $\theta$ on $\traindata$. 
The prediction of $\targetmodel$ on $X$ is a random variable: $\hat{Y}_h =1_{[\tau,1]}\circ \hat{Y}_s$  where $\hat{Y}_s = \targetmodel\circ X$ and $\tau\in [0,1]$. 
%We consider a specific instance of random variables ($X(\omega), Y(\omega)$, $S(\omega)$) where $\omega\in\Omega$ is a parameter to generate different instances. 
We present a background on probability and measured spaces in Appendix~\ref{app:notations}.

\noindent\textbf{Inference.} Once training is completed, $X(\omega)$ is passed to $\targetmodel$ to obtain a prediction score $\targetmodel(X(\omega))$ (aka soft labels).
%$X(\omega)$, 
The attributes during inference, are sampled from an unseen test dataset $\testdata$ disjoint from $\traindata$ to evaluate how well $\targetmodel$ generalizes. We refer to $\targetmodel$'s final predictions and intermediate outputs as \textit{model observables}.
Sensitive attributes, although available for different data records, play no role in training or inference. They are reserved solely for designing and evaluating attacks.

\subsection{Attribute Inference Attacks}\label{sec:back-attribute}

An attack constitutes a privacy risk if \adv learns something about $\traindata$ or the inputs which would be impossible to learn without access to $\targetmodel$. This differentiates between a privacy risk and simple statistical inference~\cite{cormode}. 
\aia{s} infer the specific value of a sensitive attribute for a specific input to ML model given some model observables and background information~\cite{fredrikson2,Mahajan2020DoesLS,yeom,Song2020Overlearning,malekzadeh2021honestbutcurious,MehnazAttInf}. \adv has access to auxiliary data $\auxdata$ which is sampled from the same distribution as $\traindata$, a standard assumption across all \aia{s}.
Based on \adv's knowledge and access, \aia{s} can be categorized into (a) imputation-based and (b) representation-based attacks. 

\noindent\textbf{\underline{Imputation-based attacks}} assume \adv has access to non-sensitive attributes and  background information (e.g., marginal prior over sensitive attribute and confusion matrix) in addition to model's predictions. Fredrikson et al.~\cite{fredrikson2}, Yeom et al.~\cite{yeom} and Mehnaz et al.~\cite{MehnazAttInf} assume that $S$ is part of the input of $\targetmodel$ and the targeted data point belongs to $\traindata$.
%is of the form $X_s = (\cdots,S,\cdots)$ where $S$ is included in the input and $\traindata$, and $X_s(\omega)$ belongs to $\traindata$.
Fredrikson et al.\cite{fredrikson2} and Mehnaz et al.\cite{MehnazAttInf} for a targeted data point, compute $\targetmodel$ for different values of the sensitive attribute to find the most likely one. 
%the value of $S$ to generate multiple inputs and choose the most likely value. 
Yeom et al.~\cite{yeom} predict $S$ using the output of a membership oracle or assuming it follows some distribution.
However, these attacks perform no better than data imputation and does not pose an actual privacy risk~\cite{jayaraman2022attribute}. Jayaraman and Evans~\cite{jayaraman2022attribute} propose a whitebox \aia which is a privacy risk in the setting where \adv has limited knowledge. 
We omit a comparison with this work due to difference in threat model.

\noindent\textbf{\underline{Representation-based attacks}} exploit the distinguishability in model observables for different values of sensitive attributes~\cite{Song2020Overlearning,Mahajan2020DoesLS,malekzadeh2021honestbutcurious}. For instance, the distribution of $\targetmodel\circ X$ for $S=$\textit{males} is different from $S=$\textit{females}. \textit{Song et al.~\cite{Song2020Overlearning} / Mahajan et al.~\cite{Mahajan2020DoesLS}} assume that $S$ is not in the input. \adv only observes $\targetmodel\circ X$. \adv trains an ML attack model $\attackmodel$ to map the output predictions $\targetmodel(X(\Omega))$ to $S(\Omega)$. \textit{Malekzadeh et al.~\cite{malekzadeh2021honestbutcurious}} assume that \adv can actively introduce a ``backdoor'' and train $\targetmodel$ to explicitly encode information about $S$ in $\targetmodel\circ X$. We omit comparison with Malekzadeh et al.~\cite{malekzadeh2021honestbutcurious} due to difference in threat model and focus on Song et al.~\cite{Song2020Overlearning} / Mahajan et al.~\cite{Mahajan2020DoesLS}.

We empirically measure attribute privacy using the resistance to \aia{s} which exploit distinguishability in model predictions for different values of sensitive attributes. Specifically, $\targetmodel$ satisfies attribute privacy if the success of \aia is random guess.

\subsection{Group Fairness}
\label{sec:back-fairness}

Generally, data records in the minority subgroup, identified by some sensitive attribute (e.g., \race or \sex), face unfair prediction behaviour compared to data records in the majority subgroup. For instance in criminal justice, \race plays a non-negligible role in predicting the likelihood of them re-offending~\cite{fairjustice}. Group fairness algorithms add constraints during training such that different subgroups (i.e., $S:\Omega\rightarrow\{0,1\}$) are treated equally (e.g., \advdebias~\cite{debiase} and \egd~\cite{reductions}). $S$ is either \sex or \race (i.e., $S(\omega)$ is 0 for woman and 1 for man, or 0 for black and 1 for white). There are different definitions of group fairness which have been introduced in prior work. We discuss two well-established definitions: demographic parity (\dempar) and equalized odds (\eo). 

\begin{definition}
\label{def:dp}
$\hat{Y}_h$ satisfies \dempar for $S$ if and only if: $P(\hat{Y}_h=0 | S=0) = P(\hat{Y}_h=0 | S=1)$.
\end{definition}

\dempar ensures that the number of correct predictions is the same for each subgroup.
However, this may result in different false positive (FPR) and true positive rates (TPR) if the true outcome varies with $S$~\cite{dpbad}.
\eo is a modification of \dempar to ensure that both TPR and FPR are the same for each subgroup~\cite{fairmetric2}.

\begin{definition}
    \label{def:eo}
    $\hat{Y}_h$, classifier of $Y$, satisfies \eo for $S$ if and only if: $ 
        P(\hat{Y}_h=\hat{y} | S=0,Y=y) = P(\hat{Y}_h=\hat{y} | S=1,Y=y) \quad \forall (\hat{y},y)\in\{0,1\}^2$.
\end{definition}

We consider two algorithms: (a) adversarial debiasing (\advdebias)~\cite{NIPS2017_48ab2f9b,debiase} and (b) exponentiated gradient descent (\egd)~\cite{reductions}. \textit{\advdebias} achieves fairness by training $\targetmodel$ to have indistinguishable output predictions in the presence of a discriminator network $f_{disc}$. $f_{disc}$ infers $S$ corresponding to a target data point given $\targetmodel\circ X$. $\targetmodel$ is then trained to minimize the success of $f_{disc}$. \advdebias outputs \textit{soft labels} (i.e., a probability attached to each value of the sensitive attribute). \textit{\egd} solves an under-constraint optimization problem to find a collection of optimal measurable functions $t_0, \cdots, t_{N-1}$ and threshold $(\tau_0,\cdots\tau_{N-1})\in[0,1]^N$. These are used to create the statistic for predictions $\hat{Y}_h$ to estimate $Y$. A random variable $I:\Omega\rightarrow\{0,\cdots,N-1\}$ selects one of the measurable functions and generates a randomized classifier: $\hat{Y}_h = 1_{[\tau_I, 1]}\circ t_I\circ X$. \egd can satisfy different fairness constraints (e.g., \dempar or \eo). \egd outputs \textit{hard labels} (i.e., a binary assignment to the sensitive attribute).

%% file: New/tables/tab_summary.tex
\setlength\tabcolsep{3pt}
\begin{table*}[!htb]
\caption{Comparison of prior \aia{s}: attack vector exploited (e.g., $\targetmodel(X(\omega))$, $X(\omega)$, $Y(\omega)$, distribution over $S$ ($P_S$) and confusion matrix $C(Y,\targetmodel\circ X)$), whether $S$ is censored, i.e., included in $\traindata$ and inputs, whether \aia{s} account for class imbalance in $S$, whether \adv is active or passive and whether the threat model is blackbox or whitebox.}
\begin{center}
\footnotesize
\resizebox{\textwidth}{!}{%
\begin{tabular}{ |c|c|c|c|c|c| }
 \hline
 \rowcolor{LightCyan} 
 \textbf{Literature} & \textbf{Attack Vector} & \textbf{Is $S$ censored?} & \textbf{Imbalance in $S$?} & \textbf{\adv} & \textbf{Threat Model} \\
 \hline
 \rowcolor{LightCyan} 
 \multicolumn{6}{|c|}{\textbf{Imputation-based Attacks}}\\
 \hline
 \textbf{Fredrikson et al.}~\cite{fredrikson2} & $X$, $Y$, $\targetmodel\circ X$, \textbf{$P_S$}, $C(Y,\targetmodel\circ X$) & $\checkmark$ & $\times$ &  Passive & Blackbox\\
 \textbf{Yeom et al.}~\cite{yeom} & $X$, $Y$, $\targetmodel$, \textbf{$P_S$} & $\checkmark$  & $\times$ &  Passive & Blackbox\\
  \textbf{Mehnaz et al.}~\cite{MehnazAttInf} & $X$, $Y$, $\targetmodel$, \textbf{$P_S$}, $C(Y,\targetmodel\circ X)$ & $\checkmark$ & $\times$ &  Passive & Blackbox\\
  \textbf{Jayaraman and Evans}~\cite{jayaraman2022attribute} & $X$, $Y$, $\targetmodel$, $P_S$, $C(Y,\targetmodel\circ X)$ & $\times$, $\checkmark$ & $\times$ &  Passive & Whitebox\\
  \hline
  \rowcolor{LightCyan} 
   \multicolumn{6}{|c|}{\textbf{Representation-based Attacks}}\\
  \hline
 \textbf{Song et al.}~\cite{Song2020Overlearning} & $\targetmodel\circ X$ & $\times$ & $\times$ &  Passive & Both\\
 \textbf{Mahajan et al.}~\cite{Mahajan2020DoesLS} & $\targetmodel\circ X$ & $\checkmark$ & $\times$ &  Passive & Blackbox\\
 \textbf{Malekzadeh et al.}~\cite{malekzadeh2021honestbutcurious} & $\targetmodel\circ X$ & $\times$ & $\times$ &  Active & Blackbox\\
 \textbf{Our Work} & $\targetmodel\circ X$ & $\times$, $\checkmark$ & $\checkmark$ &  Passive & Blackbox \\
 \hline
\end{tabular}
}
\end{center}
\label{tab:summary}
\end{table*}

%% file: 3problem.tex
\section{Problem Statement}
\label{sec:problem}

Understanding the interaction between group fairness and privacy notions is crucial for practitioners to address conflicts before model deployment~\cite{duddu2023sok}. This is essential for regulatory compliance, as both fairness and privacy are mandated by regulations~\cite{ec2019ethics,nist,dpia,ico,whitehouse}. Previous research highlights a conflict between group fairness and differential privacy showing an impossibility to train a differentially private model while ensuring group fairness~\cite{cummings}. Additionally, group fairness increases susceptibility to membership inference attacks~\cite{chang2021privacy} and distribution inference attacks~\cite{suri2023dissecting}. 
However, the interaction of group fairness with attribute
privacy remains unclear. 

\noindent\textit{Challenge 1:} %The interaction of group fairness with attribute privacy remains unclear.
Our goal is to better understand and evaluate the relation between group fairness and attribute privacy. We \textit{conjecture} that group fairness \textit{aligns} with attribute privacy by ensuring indistinguishability in the predictions for different values of $S(\omega)$. However, to effectively validate this, we need to address two other challenges.

\noindent\textit{Challenge 2:} We evaluate group fairness algorithms against \aia{s}. If the success of \aia{s} is random guess, it indicates that attribute privacy is satisfied. Appendix~\ref{app:distinguishability} shows that distinguishable output predictions for two sub-groups could be exploited to infer a sensitive attribute even when it is not included in $\targetmodel$’s $\traindata$. However, none of the current \aia{s} in literature are effective as they fail to account for real-world datasets with significant class imbalance in sensitive attributes. 
For instance, the fraction of males and whites in different datasets are 68\% and 90\% (\census), 81\% and 51\% (\compas), 53\% and 36\% (\meps), and 78\% and 96\% (\lfw). 
We have to design \aia{s} to account for this class imbalance to make them effective.

\noindent\textit{Challenge 3:} Group fairness algorithms can output either soft labels (probability scores indicating that an input belongs to different classes) or hard labels (most likely class from soft labels). We have to design \aia for both.

After addressing \textit{Challenge 2} and \textit{Challenge 3}, we can then revisit \textit{Challenge 1} to show that group fairness aligns with attribute privacy by mitigating our proposed \aia{s}.
We now present our threat model and assumptions about \adv's knowledge and capabilities.

\input{fig_tm2}

\noindent\textbf{\underline{Threat Model.}} We assume a blackbox \adv with no knowledge of $\targetmodel$'s parameters or architecture.
\adv can query $\targetmodel$ and obtain corresponding predictions. This is the most practical setting typically seen in ML as a service.  
Additionally, \adv has access to $\auxdata$ sampled from the same distribution as $\traindata$ similar to prior works~\cite{Mahajan2020DoesLS,Song2020Overlearning,yeom}. $\auxdata$ is split into two disjoint datasets: $\auxtraindata$ used for designing the attack and $\auxtestdata$ to evaluate the attack.
This is a strong assumption, but it was made to favor \adv since, in practice, \adv is likely to have a different distribution. We revisit the case where $\auxdata$ is sampled from a different distribution in Appendix~\ref{app:auxdist}.
\adv trains $\attackmodel$ on $\auxtraindata$ to infer $S$ using $\targetmodel\circ X$. $\attackmodel$ is then evaluated on $\auxtestdata$ (Figure~\ref{fig:tm2}). This attack is applicable in both cases when $\targetmodel$'s outputs are hard and soft labels:
\begin{enumerate}[label=\textbf{TM\arabic*},leftmargin=*,leftmargin=*,wide, labelwidth=!, labelindent=0pt]
\item\label{tm:hard} For hard labels,
%$\hat{Y}_h(\omega_0),\cdots,\hat{Y}_h(\omega_{A-1})$, 
\adv builds a statistic $\hat{S}$ to infer $S$: $ \hat{S}= \attackmodel\circ \hat{Y}_h\circ X$. 
\item\label{tm:soft} For soft labels, 
%$\hat{Y}_s(\omega_0),\cdots,\hat{Y}_s(\omega_{A-1})$, 
\adv builds a statistic $\hat{S}$ to infer $S$: $ \hat{S}=1_{[\upsilon,1]}\circ \attackmodel\circ \hat{Y}_s\circ X$. Here, $\upsilon \in [0,1]$ is a threshold which can be adapted to improve the attack.
\end{enumerate}

%% file: fig_tm2.tex
\begin{figure}[h]
\resizebox{.47\textwidth}{!}{%
\begin{tikzpicture}

    \node [rectangle,draw,thick,minimum width=1.5cm, minimum height=0.75cm] (targetmodel) {$\targetmodel$};
    
    \node[below of=targetmodel,yshift=-0.3cm,database,database radius=0.4cm,database segment height=0.2cm, label={below:\footnotesize $\traindata: (X, Y)$}] (trainingdata) {};

    \node [left of=targetmodel,minimum width=0.75cm,xshift=-0.8cm,fill=red!20,rectangle,draw,thick,label={below:\footnotesize Input}] (inputrecord) {$X'(\omega)$};
    \node [right of=targetmodel,xshift=1.5cm,rectangle,draw,thick] (outputpred) {$\targetmodel(X'(\omega))$};
    % \node [below of=outputpred,minimum width=1.5cm,rectangle,draw,thick] (explanation) {$\phi(x)$};

    \begin{scope}[on background layer]
        \node (tm1) [fit=(targetmodel) (trainingdata), fill= gray!20, rounded corners, inner sep=0.1cm, label={above:\footnotesize }] {};
    \end{scope}

    \node [right of=outputpred,rectangle,draw,thick,minimum width=1.5cm, minimum height=0.75cm,xshift=1.2cm,fill= gray!20] (attmodel) {$\attackmodel$};

    \node [right of=attmodel,xshift=0.6cm,minimum width=0.75cm,rectangle,draw,thick] (attout) {$S(\omega)$};

    \node[below of=attmodel,database,database radius=0.4cm,database segment height=0.2cm,yshift=-0.3cm, label={below:\footnotesize $\auxdata: (X', Y', S')$}] (auxdata) {};

\begin{scope}[on background layer]
    \node (models) [fit=(attmodel) (outputpred) (attout) (auxdata), fill= red!20, rounded corners, inner sep=0.1cm, label={above:\footnotesize Accessible to \adv}] {};
\end{scope}

\draw[->,ultra thick] (inputrecord.east) -- node[anchor=south, align=center] {\em\footnotesize } (targetmodel.west);
\draw[->,ultra thick] (targetmodel.east) -- node[anchor=south, align=center] {\em\footnotesize } (outputpred.west);
% \draw[->,ultra thick] (targetmodel.east) -- node[anchor=south, align=center] {\em\footnotesize } (explanation.west);

\draw[->,ultra thick, dashed] (outputpred.east) -- node[anchor=south, align=center] {\em\footnotesize } (attmodel.west);
% \draw[->,ultra thick, dashed] (explanation.east) -- node[anchor=south, align=center] {\em\footnotesize } (attmodel.west);
\draw[->,ultra thick] (attmodel.east) -- node[anchor=south, align=center] {\em\footnotesize } (attout.west);
\draw[->,ultra thick,dashed] (trainingdata.north) -- node[anchor=south, align=center,label={[yshift=-0.2cm]right:\footnotesize Train}] {\em\footnotesize } (targetmodel.south);
\draw[->,ultra thick,dashed] (auxdata.north) -- node[anchor=south, align=center,label={[yshift=-0.2cm]right:\footnotesize Train}] {\em\footnotesize } (attmodel.south);

\end{tikzpicture}
}
\caption{\adv wants to infer $S(\omega)$ for $X(\omega)$ given $\targetmodel(X(\omega))$. \adv uses $\attackmodel$ which is trained on $\auxdata$ to map $S'(\omega)$ from $\targetmodel(X'(\omega))$.} 
\label{fig:tm2}
\end{figure}

%% file: 4setup.tex
\section{Experimental Setup}
\label{sec:setup}

\noindent\textbf{\underline{Datasets:}}
\label{sec:datasets} We consider four real-world datasets covering different domains: criminal justice (\compas), income prediction (\census), healthcare (\meps), and face recognition (\lfw), to illustrate the effectiveness of the proposed \aia{s}. These have been used as benchmarks for privacy~\cite{Song2020Overlearning,MehnazAttInf,Mahajan2020DoesLS} and fairness~\cite{surveyfair}.

\census comprises 30,940 data records with 95 attributes about individuals from 1994 US Census data.
The attributes include marital status, education, occupation, job hours per week among others.
The classification task is to estimate whether an individual makes an income of 50k per annum. For $\targetmodel$, we use 24,752 data records for $\traindata$ and 6,188 data records as $\testdata$. We use 4,950 data records from $\testdata$ as $\auxtraindata$ for training $\attackmodel$ and evaluate it on 1,238 unseen data records in $\auxtestdata$.

\compas is used for commercial algorithms by judges and parole officers for estimating the likelihood of a criminal re-offending using seven attributes.
The classification task is whether a criminal will re-offend or not, and contains 6,172 criminal defendants in Florida. 
The dataset contains 7 attributes and we use 4,937 data records for training $\targetmodel$ and 1,235 data records as testing dataset. We use 988 data records from $\testdata$ as $\auxtraindata$ for training $\attackmodel$, i.e., $\auxtraindata$ and evaluate it on 247 unseen data records in $\auxtestdata$.

\meps contains 15,830 records of different patients using medical services by capturing the trips made to clinics and hospitals.
The classification task is to predict the utilization of medical resources as 'High' if the sum of the number of office based visits, outpatient visits, ER visits, inpatient nights and home health visits, is greater than ten. 
We use 12,664 data records for training $\targetmodel$ and 3,166 $\traindata$ records as testing dataset. We use 2,532 data records from $\testdata$ as $\auxtraindata$ for training $\attackmodel$ and evaluate it on 634 unseen data records in $\auxtestdata$.

\lfw has 8,212 images of people with the classification task to predict whether their age is $>$35 years. In all datasets, we consider \race and \sex as binary sensitive attributes to be inferred.

We use 80\% of the dataset as $\traindata$ and the remaining 20\% for $\testdata$.
We use $\testdata$ as $\auxdata$ and ensure that the distribution of $S$ is uniform between them. We use 80\% of $\auxdata$ for training $\attackmodel$ and 20\% for evaluation of the attack.
The impact of an adversary using auxiliary knowledge (Daux) which differs from the target model training data () is evaluated in the Appendix~\ref{app:auxdist}.
We use cross validation where each split is done five times without any overlap.
$\targetmodel$ is trained and evaluated five times and $\attackmodel$ is trained and validated ten times.
% \change{NO ANOVA YET}
We check for statistical significance for the results (i.e., p-value <0.05).

\noindent\textbf{\underline{Model Architectures:}} For all the datasets, we use neural networks with four hidden layers with the following dimensions: [32, 32, 32, 32] and ReLU activation functions.

\noindent\underline{\textbf{Metrics:}}
We use standard classification \textit{accuracy} between predicted labels and ground truth labels for evaluating $\targetmodel$'s performance. We refer to this as \textit{utility}.
We evaluate attack success using \textit{balanced accuracy} which is the average of the proportion of correct predictions of each class of the sensitive attribute individually: $\frac{1}{2}(P(\hat{S}=0|S=0) + P(\hat{S}=1|S=1))$. This metric accounts for class imbalance and simple accuracy is misleading when the datasets have significant class imbalance. An accuracy of 50\% corresponds to random guesses.
To evaluate fairness, we use \textit{\dempar\_Level} given by $|P(\hat{Y}=0 | S=0) - P(\hat{Y}=0 | S=1)|$. \dempar-level close to zero indicates that $\targetmodel$ is fair.

\noindent\underline{\textbf{Baselines:}}
To evaluate the impact of group fairness on attribute privacy, we compare the attack success of \adaptiveAIA with and without using group fairness 
(the former case is referred to as \empirical while the latter is referred to as \baseline). 
For classifiers that output hard labels (i.e., using  \adaptiveAIAHard), we also indicate the theoretical bound from Theorem (referred to as \theoretical).

%\noindent\underline{\textbf{Methodology:}}
% add a word about the lib / code you used and about the calibration of the hyperparameter that drives the utility and privacy/fairness 

%% file: 5attacks.tex
\section{\adaptiveAIA: An Effective \aia}
\label{sec:adaptiveAIA}

We first address \textit{Challenge 2} and \textit{Challenge 3} from Section~\ref{sec:problem}. Our goal is to design an \aia which accounts for class imbalance in $S$, typical of real-world applications~\cite{classIMb1,classIMb2}, and is applicable for soft and hard labels. We present \adaptiveAIA, an \aia with adaptive threshold and its two variants of the attack, \adaptiveAIASoft for soft labels, and \adaptiveAIAHard for hard labels. We describe our proposed attacks below.

\noindent\textbf{\underline{\adaptiveAIASoft.}} Recall from Section~\ref{sec:problem} that datasets have significant class imbalance in sensitive attributes. 
We conjecture that this skews $\targetmodel$'s predictions, thereby requiring us to adapt $\attackmodel$ to correctly infer the value of sensitive attributes~\cite{classIMb1,classIMb2,classIMb3}.
Hence, the default threshold of $0.5$ over $\attackmodel$'s soft labels will not result in accurate estimation of $S$. \adv's attack success can be improved by using an adaptive threshold~\cite{classIMb1,classIMb2,classIMb3}.
However, none of the prior \aia{s} account for this. 

Hence, we can improve \aia{s} with an adaptive threshold instead of using a default threshold of $0.5$ as in prior \aia{s}. \adv computes the optimized threshold $\upsilon^*$ on $\auxtraindata$ which is later used for the attack on $\auxtestdata$.
We compute $\upsilon^*$ to balance TPR and FPR. Ideally, a perfect attack would result in no FPR and only TPR.
\adv's goal is thus to approach this optimal value of TPR and FPR.
Formally, $\upsilon^*\in[0,1]$ where $\upsilon^* = \text{argmin}_{\upsilon} (1-TPR_\upsilon)^2 + FPR_\upsilon^2$.

\begin{figure}[!htb]
    \centering
    \includegraphics[width=0.6\linewidth]{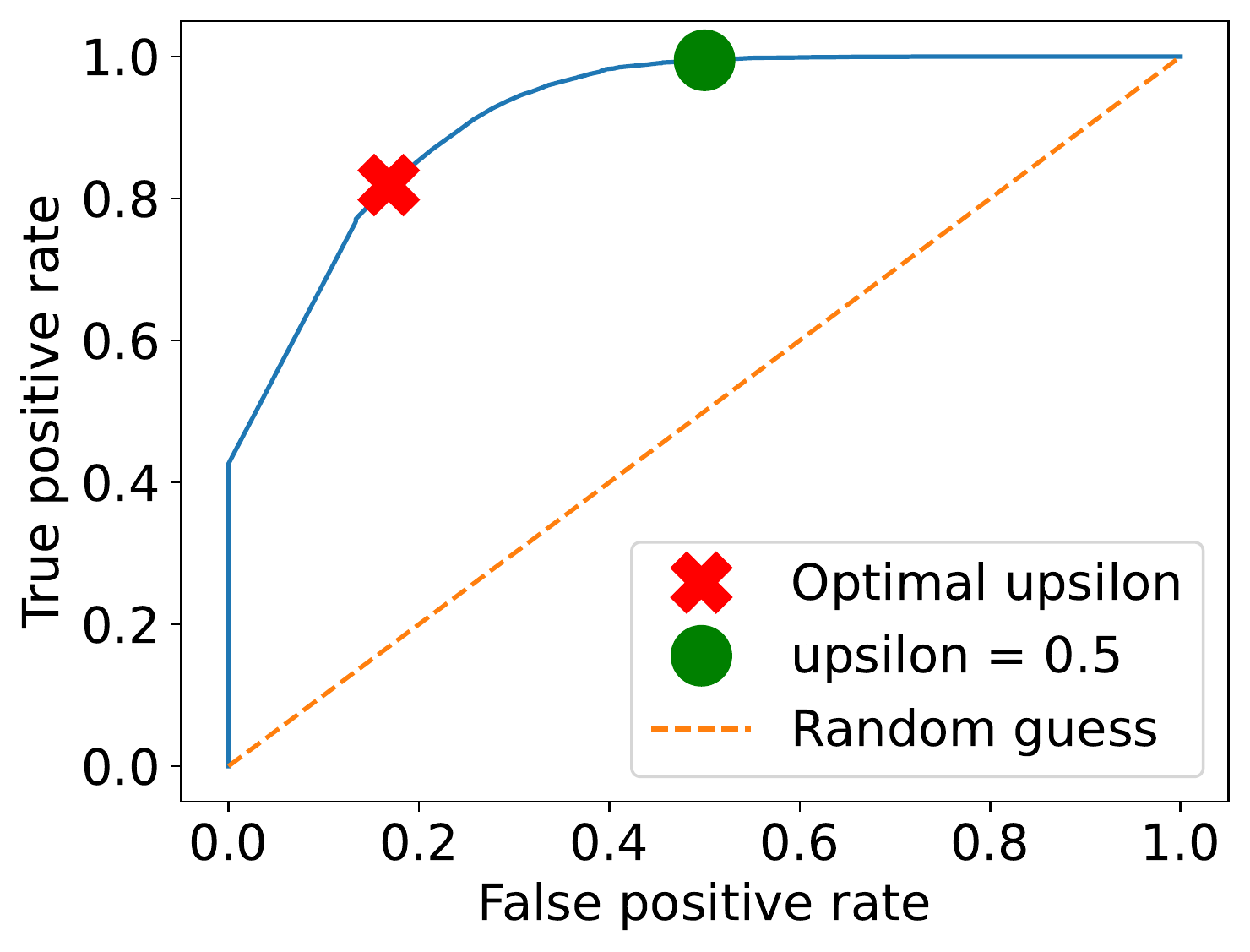}
    \caption{ROC curve: $\upsilon^*$ can lower FPR to infer \race.}
    \label{fig:curvesROC}
\end{figure}

For illustration purposes, we plot the ROC curve for inferring \race in Figure~\ref{fig:curvesROC}. 
We observe that $\upsilon^*$ does not correspond to the default classification threshold of $0.5$ used in the literature and results in lower FPR resulting in a more confident attack.

\noindent\textbf{\underline{\adaptiveAIAHard.}} For hard labels, it is not necessary to train $\attackmodel$.
Instead, we consider a set of functions from $\{0,1\}$ to $\{0,1\}$ containing four elements: $x\mapsto 0$, $x\mapsto x$, $x\mapsto 1-x$, and $x\mapsto 1$.
We optimize the attack by finding the functions which give the best balanced accuracy on $\auxtraindata$ which is used to evaluate attack success on $\auxtestdata$.

\noindent\textbf{\underline{Evaluating \adaptiveAIAHard and \adaptiveAIASoft.}} For hard labels (\ref{tm:hard}), we consider the baseline of training a neural network over hard labels to infer the value of $S$ given $X$. We then compare this with \adaptiveAIAHard. For soft labels (\ref{tm:soft}), we consider the attacks by Song et al.~\cite{Song2020Overlearning}/Mahajan et al.~\cite{Mahajan2020DoesLS} as the prior state-of-the-art baselines with the default $\upsilon=0.5$ over $\attackmodel(\targetmodel(X(\omega)))$. We then compare this with \adaptiveAIASoft. 

\begin{table}[!htb]
\caption{Comparing attack accuracy (average over ten runs) for baseline with $\upsilon=0.5$~\cite{Song2020Overlearning,Mahajan2020DoesLS} and both \adaptiveAIAHard and \adaptiveAIASoft: 
an adaptive threshold improves the success of the attack for hard and soft labels.}
\begin{center}
\footnotesize
\begin{tabular}{ l | c | c  }
\hline
\rowcolor{LightCyan}  & \multicolumn{2}{c}{\ref{tm:hard}}\\
\rowcolor{LightCyan}  \textbf{Dataset} & \textbf{Baseline ($\upsilon$=0.50)} & \textbf{\adaptiveAIAHard}\\ 
\rowcolor{LightCyan}  & \textbf{\race} | \textbf{\sex}& \textbf{\race} | \textbf{\sex}\\
\textbf{\census} & 0.50 $\pm$ 0.0 | 0.50 $\pm$ 0.0& \textbf{0.56 $\pm$ 0.01} | \textbf{0.58 $\pm$ 0.01} \\
\textbf{\compas}& \textbf{0.62 $\pm$  0.3} | 0.50 $\pm$ 0.0& \textbf{0.62 $\pm$ 0.03} | \textbf{0.57 $\pm$ 0.03} \\ 
\textbf{\meps} & 0.51 $\pm$ 0.01 | \textbf{0.55 $\pm$ 0.02} & \textbf{0.53 $\pm$ 0.01} | \textbf{0.55 $\pm$ 0.01} \\
\textbf{\lfw} & 0.59 $\pm$ 0.00  | 0.64 $\pm$ 0.15& \textbf{0.61 $\pm$ 0.11} | \textbf{0.78 $\pm$ 0.05}  \\
\hline
\rowcolor{LightCyan} & \multicolumn{2}{c}{\ref{tm:soft}} \\
\rowcolor{LightCyan}  \textbf{Dataset} & \textbf{Baseline ($\upsilon$=0.50)} & \textbf{\adaptiveAIASoft} \\ 
 \rowcolor{LightCyan} & \textbf{\race} | \textbf{\sex} & \textbf{\race} | \textbf{\sex}\\
\hline
\textbf{\census}&  0.50 $\pm$ 0.02 | 0.56 $\pm$ 0.04 & \textbf{0.61 $\pm$ 0.02} | \textbf{0.68 $\pm$ 0.01}  \\
\textbf{\compas}& \textbf{0.62 $\pm$ 0.03} | 0.50 $\pm$ 0.00 & \textbf{0.62 $\pm$ 0.03} | \textbf{0.57 $\pm$ 0.03} \\ 
\textbf{\meps} & 0.52 $\pm$ 0.02 | 0.55 $\pm$ 0.02 & \textbf{0.60 $\pm$ 0.02} | \textbf{0.62 $\pm$ 0.02}\\
\textbf{\lfw} & 0.50 $\pm$ 0.10 | \textbf{0.77 $\pm$ 0.07} & \textbf{0.61 $\pm$ 0.10} | \textbf{0.79 $\pm$ 0.05}\\
\hline
\end{tabular}
\end{center}
\label{tab:global_threshold_withoutsattr}
\end{table}

We present the comparison of the baseline attack with \adaptiveAIAHard and \adaptiveAIASoft in Table~\ref{tab:global_threshold_withoutsattr}. For all datasets, we see that \adaptiveAIAHard and \adaptiveAIASoft are significantly better on average than prior \aia{s}. Having successfully addressed \textit{Challenge 2} and \textit{Challenge 3} by designing \adaptiveAIAHard and \adaptiveAIASoft, we use the \aia{s} to evaluate the alignment with two well-established group fairness algorithms: \egd and \advdebias.

%% file: 6egd.tex
\section{Alignment of \egd}
\label{sec:EGD}

We first consider \egd under the threat model \ref{tm:hard} as
it outputs hard labels. We evaluate % to see 
if \egd is aligned with attribute privacy by mitigating \adaptiveAIAHard. We also quantify the impact on both the utility and group fairness of $\targetmodel$. %Recall from
As described in Section~\ref{sec:back-fairness}, %that 
$\targetmodel$ can be trained with \egd to satisfy either \dempar or \eo. Here, we focus on \egddp but revisit \egdeo in Appendix~\ref{app:egdeo} and demonstrate that \eo cannot mitigate \adaptiveAIAHard and by consequence \adaptiveAIASoft. 
%Section~\ref{sec:discussions}. 
We first present theoretical results followed by empirical evaluation.

\noindent\textbf{\underline{\egd: Theoretical Guarantees}} We theoretically compute the bound on the attack accuracy under \egddp (proof in Appendix~\ref{app:egd}).

\begin{theorem}
\label{th:dpgood}
The maximum attack accuracy achievable by \adaptiveAIAHard is equal to $\frac{1}{2}(1+\text{\dempar-Level of }\targetmodel)$. 
\end{theorem}
Hence, we obtain a bound for \adaptiveAIAHard without any conditions on $\targetmodel$ or datasets.
Additionally, we observe that \dempar-Level$=0$. 
Consequently, if $\targetmodel$ satisfies \dempar then no $\attackmodel$ will perform better than a random guess. Hence, \egddp satisfies attribute privacy.

\noindent\textbf{\underline{\egd: Empirical Evaluation}}
We will now empirically validate the above theoretical guarantee by evaluating $\targetmodel$ trained with \egddp against \adaptiveAIA. Due to space limitation, we only present results for \census and \meps and move the results for other datasets in Appendix~\ref{app:egd}.
%We present additional results for sanity check to show $\targetmodel$ indeed satisfies fairness along with the impact of \egddp on $\targetmodel$'s utility in Appendix~\ref{app:egd}.

%To evaluate alignment of \egddp, we compare Figure~\ref{fig:AdaptAIAEGD} the result of \adaptiveAIAHard on a classification without using group fairness (referred to as \baseline) and on a classification using \egddp. For this latter case,  %For \adaptiveAIAHard, 
%we indicate attack accuracy under \egddp from both theoretical bound from Theorem~\ref{th:dpgood} (referred to as \theoretical % ``Theory''
%) and empirically (referred to as  \empirical). %``Empirical'').

\input{New/figures/fig_egd_attack}

%We illustrate our findings in Figure~\ref{fig:AdaptAIAEGD}. 

To evaluate alignment of \egddp with attribute privacy, we compare in Figure~\ref{fig:AdaptAIAEGD} the result of \adaptiveAIAHard on a classification without using \egddp and on using \egddp.
Results show that \adaptiveAIAHard demonstrates significantly lower effectiveness (approaching random guessing, 50\%) when utilizing \egddp compared to the \baseline. %``Baseline''.
Additionally, for \adaptiveAIAHard, we note that the theoretical bound on attack accuracy matches with the empirical attack accuracy. 
The theoretical accuracy is equal to the empirical accuracy when the values are $>\frac{1}{2}$. 
But $\dempar-Level \geq 0$ implies that $(1+ \dempar-Level) \geq \frac{1}{2}$.
Hence, we observe that the theoretical accuracy is not equal to the experimental when $\attackmodel$'s attack accuracy is random guess (under $\frac{1}{2}$).
This happens when $\targetmodel$ nearly follows \dempar where \adv's $\attackmodel$ is optimal on $\traindata$ but worse than random guess for $\testdata$.
% Since $\traindata$ and $\testdata$ are sampled from the same distribution, this difference in behavior only occur when the $\attackmodel$'s accuracy is near random guess.
% In this case, the slightest difference between $\traindata$ and $\testdata$ due to random sampling can result in an optimal $\attackmodel$ on $\traindata$, to not be optimal on $\testdata$.

\begin{figure}[!htb]
    \centering
    \begin{minipage}[b]{1\linewidth}
    \centering
    \subfigure[\census]{
    \includegraphics[width=0.49\linewidth]{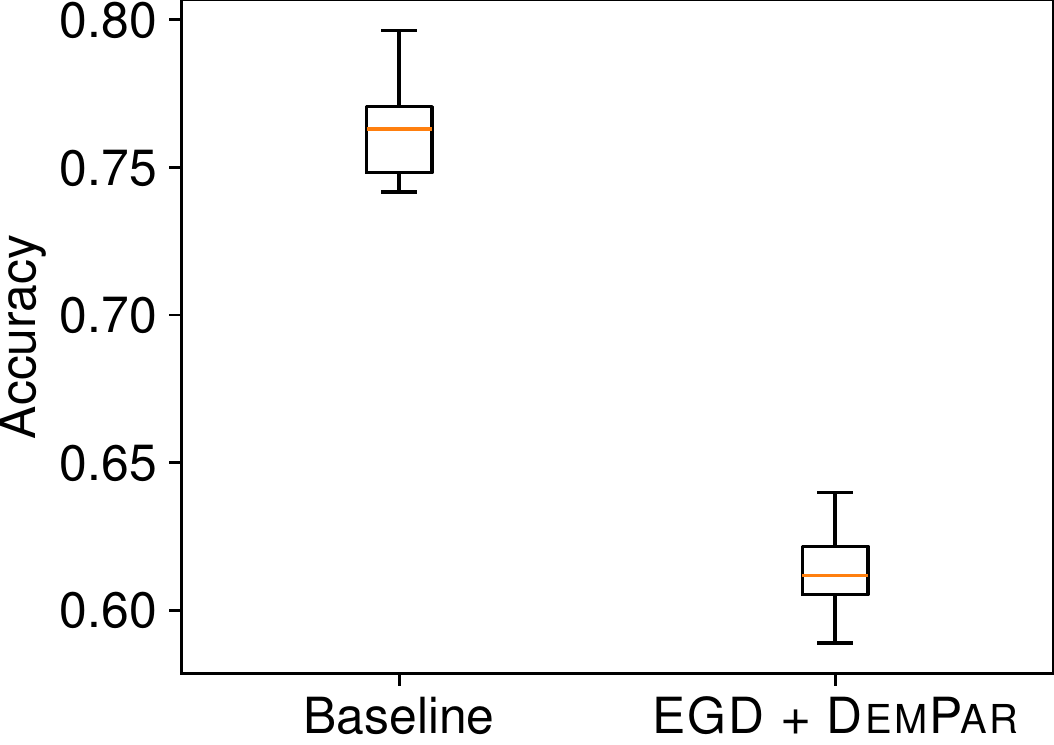}
    }%
    \subfigure[\meps]{
    \includegraphics[width=0.49\linewidth]{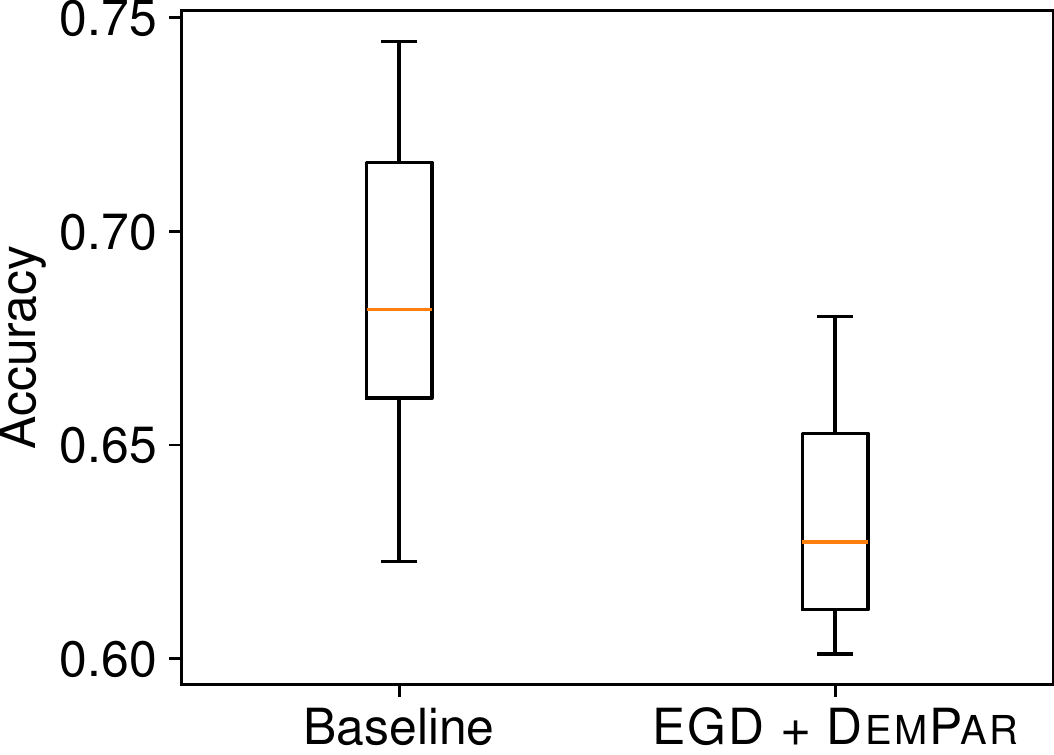}
    }
    \end{minipage}\\
    
    \caption{Imposing fairness with \egddp has a significant impact on the accuracy of $\targetmodel$, matching with the observation from prior work~\cite{reductions}}%Utility degradation for \egd: We observe a statistically significant drop in $\targetmodel$'s accuracy on using \egddp which matches the observation from prior work~\cite{reductions}.}
    \label{fig:utilityEGD}
\end{figure}

\noindent\textbf{Trade-off with $\targetmodel$'s Utility.}
Using group fairness improves the attribute privacy, but comes at the cost of $\targetmodel$'s utility~\cite{reductions}.
To quantify this impact, we compare in Figure~\ref{fig:utilityEGD} the accuracy of $\targetmodel$ with and without using \egddp.
Results show that the utility is reduced by 15\% \census and 5\% for \meps datasets. % (by x\% on average on all datasets, Appendix~\ref{}). 
This trade-off with the utility of $\targetmodel$ is inherent to the group fairness algorithms~\cite{accfairtradeoff,rodolfa2021empirical,zhai2022understanding,veldanda2022fairness}.
%for  degrades which is in line to findings from prior work~\cite{reductions}. In other words, fairness comes at the cost of accuracy.

%\noindent\textbf{Impact on Utility.} We now present the impact on $\targetmodel$'s utility on using \egddp. We present our results in Figure~\ref{fig:utilityEGD}. We indicate the model accuracy on $\testdata$ for $\targetmodel$ without \egddp as ``baseline''. We find that the utility in degrades which is in line to findings from prior work~\cite{reductions}. In other words, fairness comes at the cost of accuracy.

\begin{figure}[!htb]
    \centering
    \begin{minipage}[b]{1\linewidth}
    \centering
    \subfigure[\census (\race)]{S
    \includegraphics[width=0.49\linewidth]{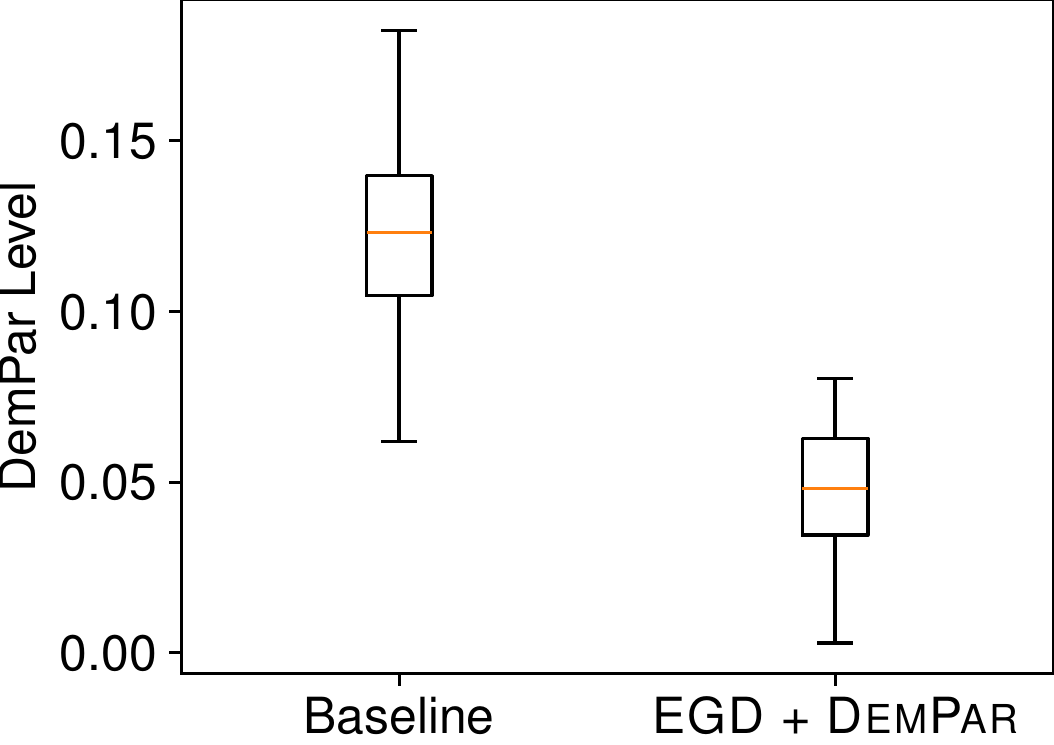}
    }%
    \subfigure[\census (\sex)]{
    \includegraphics[width=0.49\linewidth]{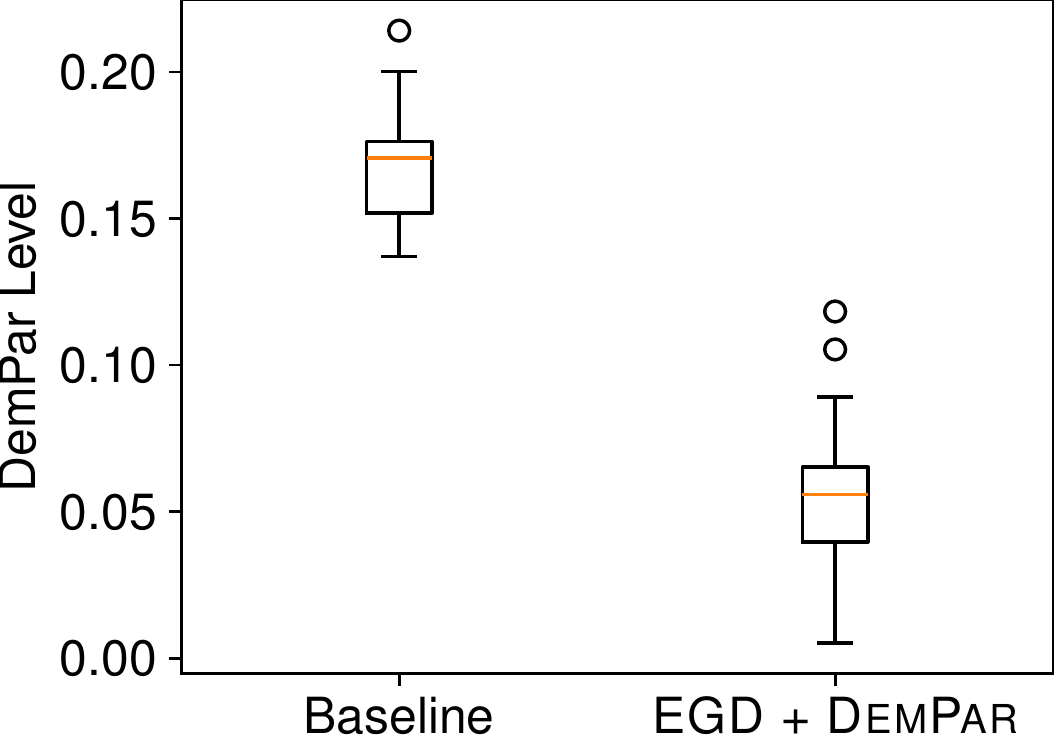}
    }
    \end{minipage}\\

    \begin{minipage}[b]{1\linewidth}
    \centering
    \subfigure[\meps (\race)]{
    \includegraphics[width=0.49\linewidth]{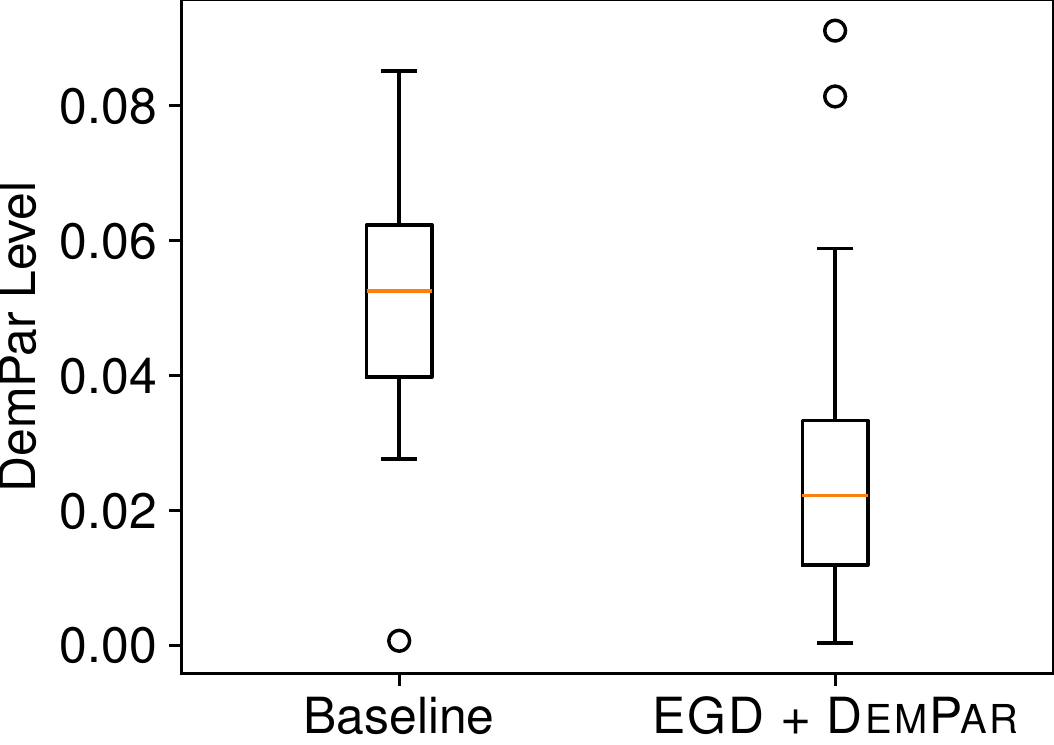}
    }%
    \subfigure[\meps (\sex)]{
    \includegraphics[width=0.49\linewidth]{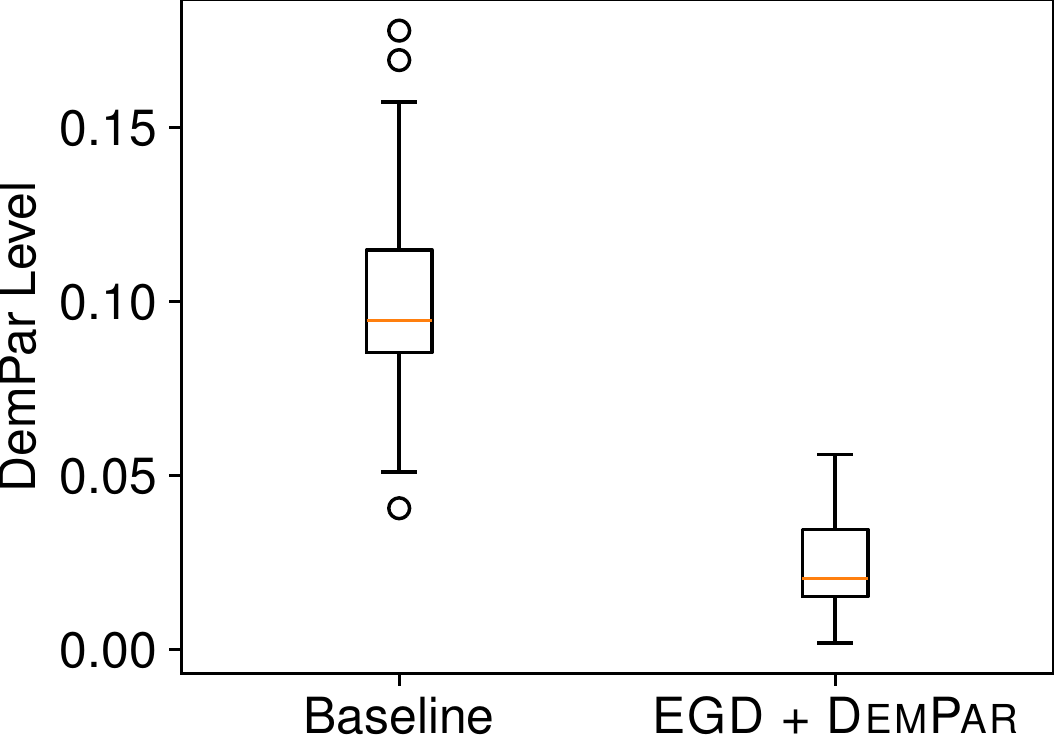}
    }
    \end{minipage}\\

    \caption{$\dempar-Level$ close to 0 
    indicates that \egddp successfully imposes 
    fairness to $\targetmodel$.}
%    We observe that \dempar-Level is lower for \egd than the baseline indicating $\targetmodel$ is fair after \egd.}
    \label{fig:DemParegd}
\end{figure}

\noindent\textbf{Sanity Check for Fairness.} We now quantify group fairness as measured by \dempar-level with and without training $\targetmodel$ with \egddp (the lower the better). %We refer to $\targetmodel$ without \egddp as ``baseline''. 
As seen in Figure~\ref{fig:DemParegd}, we observe that $\targetmodel$ with \egddp has significantly lower \dempar-level which is closer to zero as compared to baseline. Hence, \egddp is effective in achieving group fairness. 
% the lower the better

\textbf{Summary:} Considering our theoretical and empirical evaluation, we conclude that \egddp induces attribute privacy. This alignment allows us to calibrate the two dimensional trade-off between group fairness and utility while accounting attribute privacy.
%Given our theoretical and empirical evaluation, %results, 
%we conclude that \egddp aligns with attribute privacy by mitigating \adaptiveAIAHard. 

%% file: New/figures/fig_egd_attack.tex
\begin{figure}[!htb]
    \centering
    \begin{minipage}[b]{1\linewidth}
    \centering
    \subfigure[\census (\race)]{
    \includegraphics[width=0.49\linewidth]{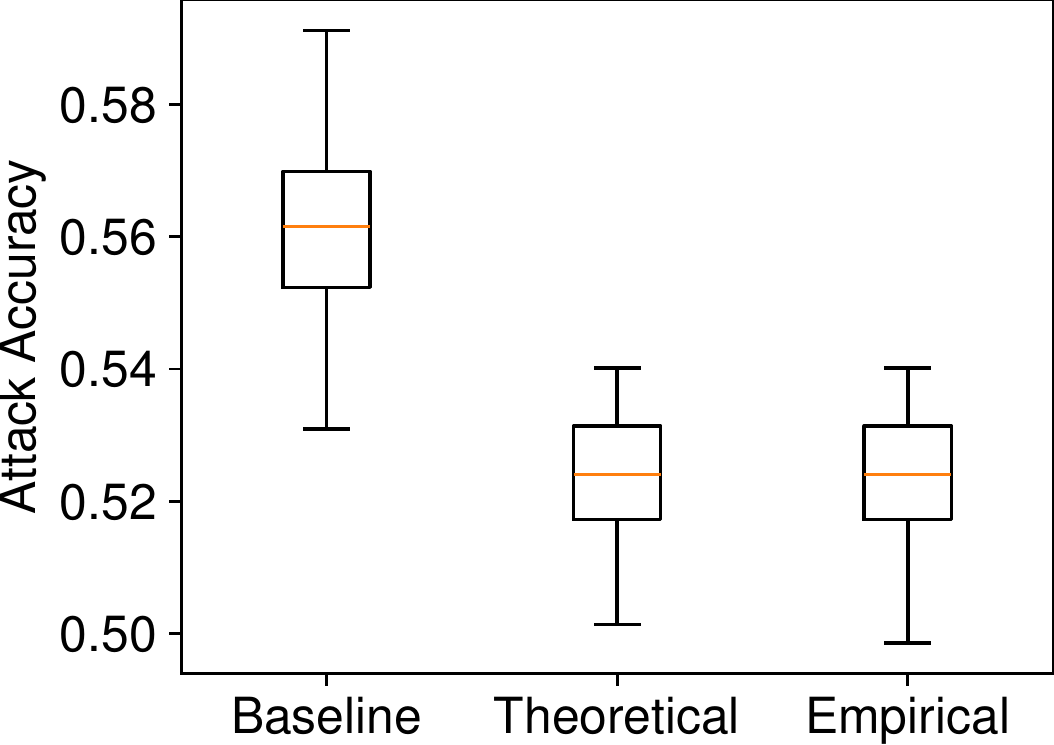}
    }%
    \subfigure[\census (\sex)]{
    \includegraphics[width=0.49\linewidth]{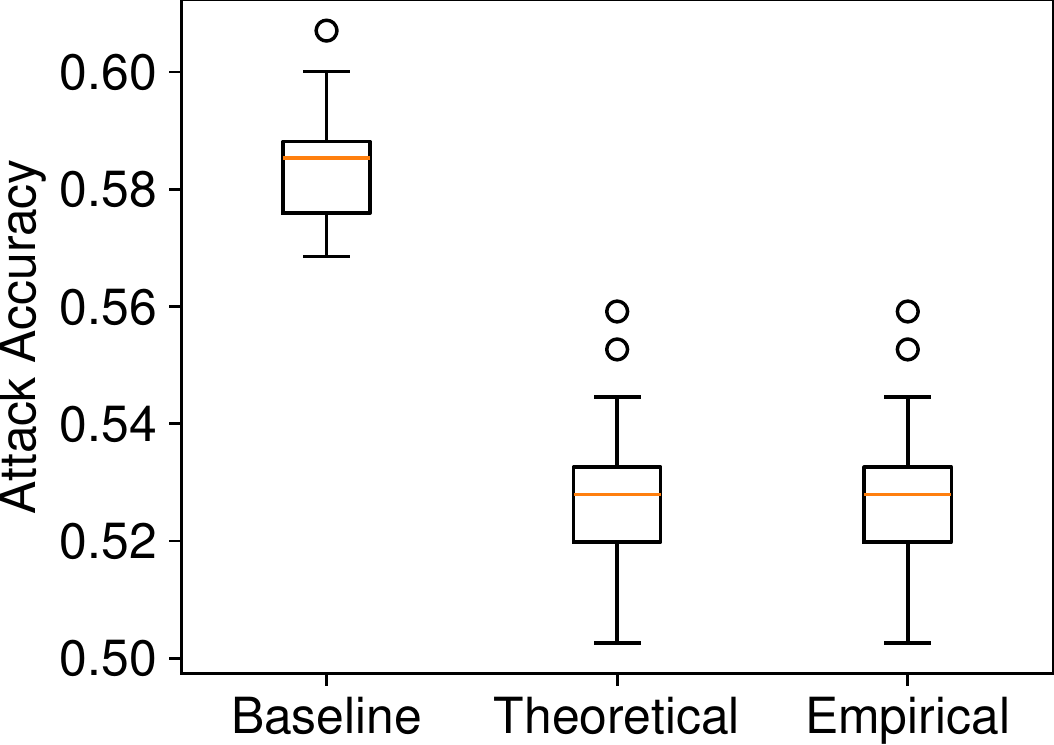}
    }
    \end{minipage}\\

%    \begin{minipage}[b]{1\linewidth}
%    \centering
%    \subfigure[\compas (\race)]{
%    \includegraphics[width=0.49\linewidth]{New/figures/egd/compas/compas_egd_attack_hard_race.pdf}
%    }%
%    \subfigure[\compas (\sex)]{
%    \includegraphics[width=0.49\linewidth]{New/figures/egd/compas/compas_egd_attack_hard_sex.pdf}
%    }
%    \end{minipage}\\

    \begin{minipage}[b]{1\linewidth}
    \centering
    \subfigure[\meps (\race)]{
    \includegraphics[width=0.49\linewidth]{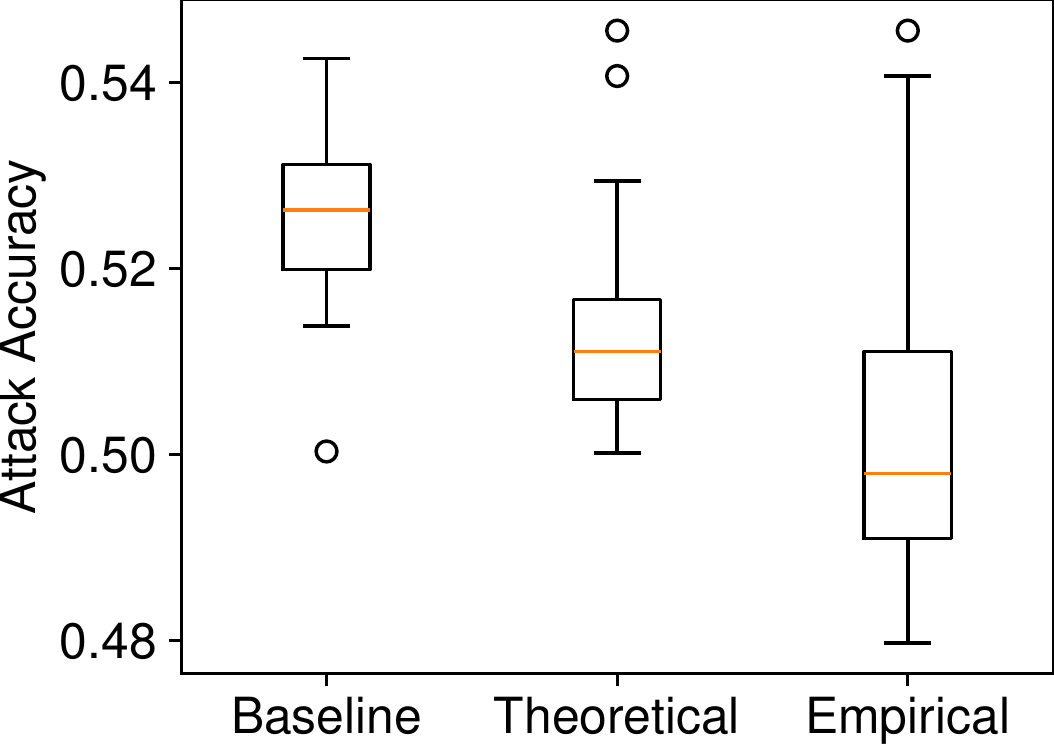}
    }%
    \subfigure[\meps (\sex)]{
    \includegraphics[width=0.49\linewidth]{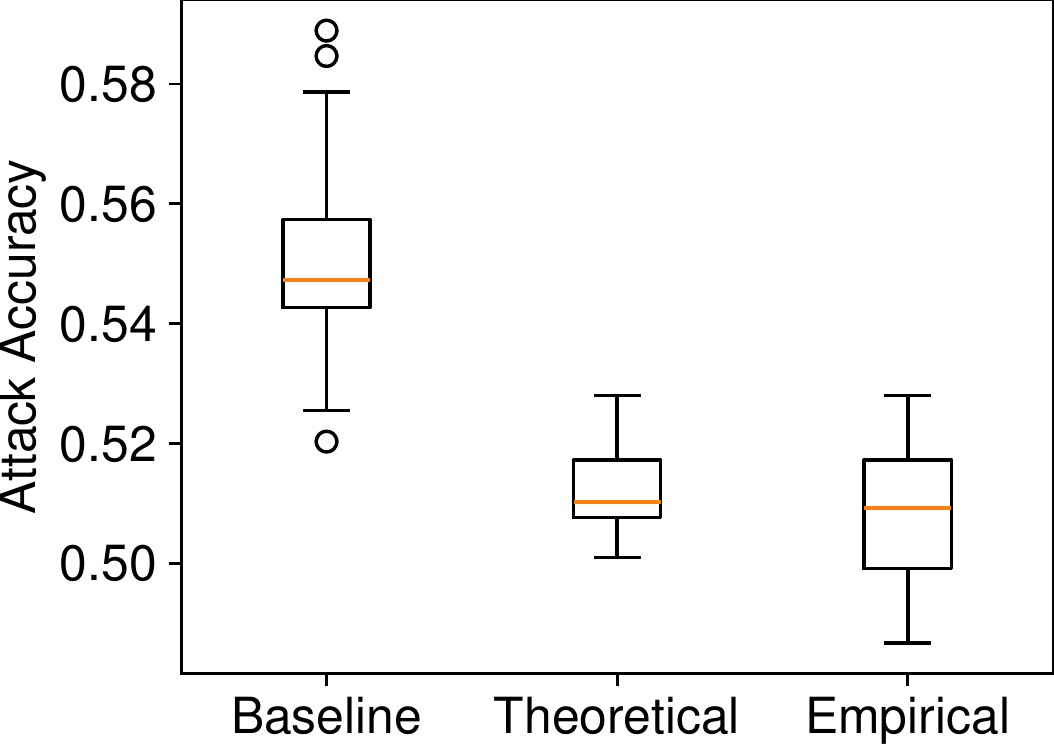}
    }
    \end{minipage}\\

%    \begin{minipage}[b]{1\linewidth}
%    \centering
%    \subfigure[\lfw (\race)]{
%    \includegraphics[width=0.49\linewidth]{New/figures/egd/lfw/lfw_egd_attack_hard_race.pdf}
%    }%
%    \subfigure[\lfw (\sex)]{
%    \includegraphics[width=0.49\linewidth]{New/figures/egd/lfw/lfw_egd_attack_hard_sex.pdf}
%    }
%    \end{minipage}

    \caption{For \adaptiveAIAHard, we observe that \egd reduces the attack accuracy to random guess ($\sim$50\%). %Additionally, the theoretical bound on attack accuracy (``Theory'') matches with the empirical results (``Empirical'').
    }
    \label{fig:AdaptAIAEGD}
\end{figure}

%\textbf{Overall, we observe that using group fairness results in attribute privacy but comes at the cost of $\targetmodel$'s utility.}

%% file: 7advdebias.tex
\section{Alignment of \advdebias}
\label{sec:advdebias}

We now consider \advdebias under threat model \ref{tm:soft} as it outputs soft labels. As described % Recall from 
Section~\ref{sec:back-fairness}, % that 
\advdebias minimizes the success of $f_{disc}$ to correctly infer the value of $S$, which makes the predictions indistinguishable. We can view $f_{disc}$ as a weak \aia without adaptive threshold. Since \advdebias accounts for $f_{disc}$, we conjecture that the success of \adaptiveAIA will decrease but not completely, thereby aligning attribute privacy. We validate this conjecture both theoretically and empirically.

 % We then empirically validate this by evaluating $\targetmodel$ trained with \advdebias against \adaptiveAIASoft and \adaptiveAIAHard.

\noindent\textbf{\underline{\advdebias: Theoretical Guarantees}} 
We first theoretically show that using \advdebias bounds the balanced attack accuracy to random guess (proof in Appendix~\ref{app:advdebias}).

\begin{theorem}\label{th:advdebias}
    The following propositions are equivalent:
    \begin{itemize}
        %\item $\hat{Y}$ satisfies extended \dempar for $S$
        \item $\hat{Y}_s$ is independent of $S$ ;
        \item Balanced accuracy of \adaptiveAIASoft is $\frac{1}{2}$.
    \end{itemize}
\end{theorem}

\noindent\textbf{\underline{\advdebias: Empirical Evaluation}} 
We will now empirically validate the above theoretical guarantee by evaluating $\targetmodel$ trained with \advdebias against \adaptiveAIASoft.
This evaluation of \advdebias using the soft labels can be converted to hard labels for using \adaptiveAIAHard. 
Due to space constraints, this section only reports the results of \adaptiveAIASoft, leaving the results of \adaptiveAIAHard in Appendix~\ref{app:advdebias}. Similarly to Section~\ref{sec:EGD}, we only report results for \census and \meps and move the results for other datasets in Appendix~\ref{app:advdebias}.

%We will now empirically validate the above theoretical guarantee by evaluating $\targetmodel$ trained with \advdebias against \adaptiveAIASoft and \adaptiveAIAHard. Unlike \egd, we can use both \adaptiveAIASoft and \adaptiveAIAHard for evaluation as the soft labels from \advdebias can be converted to hard labels. We present additional results for sanity check to show $\targetmodel$ indeed satisfies fairness along with the impact of \advdebias on $\targetmodel$'s utility in Appendix~\ref{app:advdebias}.

%To evaluate alignment of \advdebias, we compare the \baseline of without using group fairness and on using \advdebias. For \adaptiveAIAHard, we indicate attack accuracy under \advdebias from both theoretical bound from Theorem~\ref{th:dpgood} (referred to as \theoretical) and empirically (referred to as \empirical).

\input{New/figures/fig_advdebias_attacc}

To evaluate the alignment of the fairness constraint imposed by \advdebias with attribute privacy, we compare in Figure~\ref{fig:AdaptAIADebias} the attack success of \adaptiveAIASoft on the target model trained with and without group fairness using \advdebias.
%
%We illustrate our findings in Figure~\ref{fig:AdaptAIADebias}. Since, \advdebias outputs soft labels we present the results against \adaptiveAIASoft. 
%
Results show that for all datasets \advdebias reduces the attack accuracy close to random guess (i.e., 50\%) regardless of the value of the attack accuracy without group fairness (i.e., \baseline).
%In all datasets, \adaptiveAIASoft demonstrates significantly lower effectiveness (approaching random guessing, 50\%) when utilizing \advdebias compared to \baseline. 

%\textbf{Given our theoretical and empirical results, we conclude that \advdebias aligns with attribute privacy by mitigating \adaptiveAIA.}

\begin{figure}[!htb]
    \centering
    \begin{minipage}[b]{1\linewidth}
    \centering
    \subfigure[\census]{
    \includegraphics[width=0.49\linewidth]{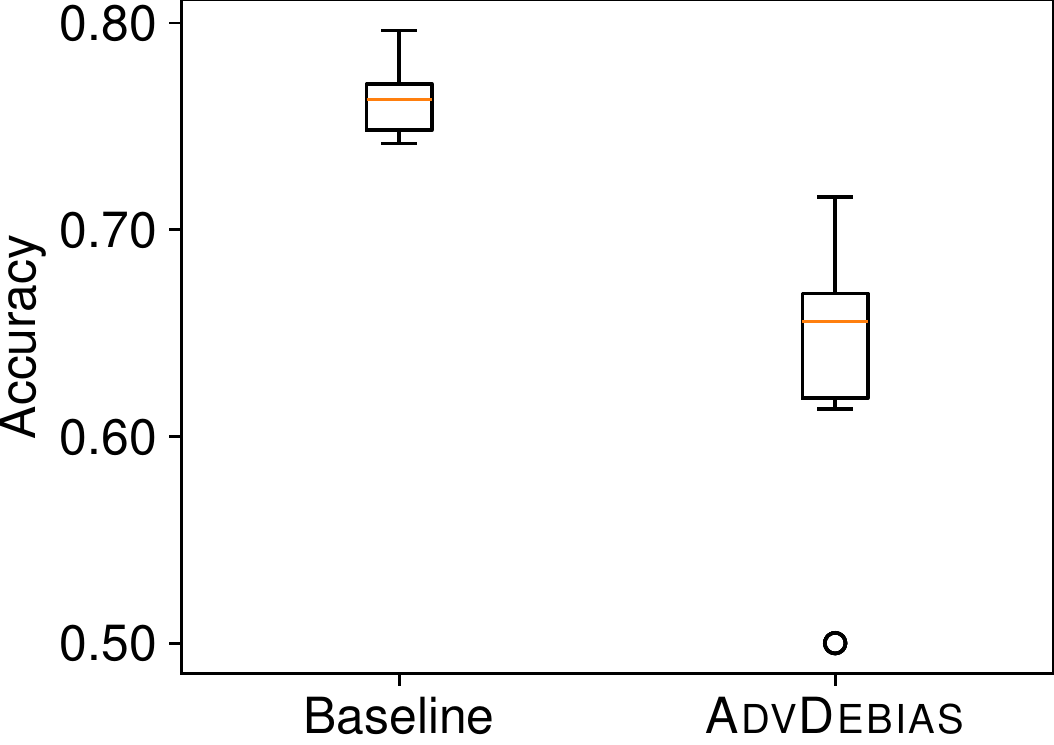}
    }%
    \subfigure[\meps]{
    \includegraphics[width=0.49\linewidth]{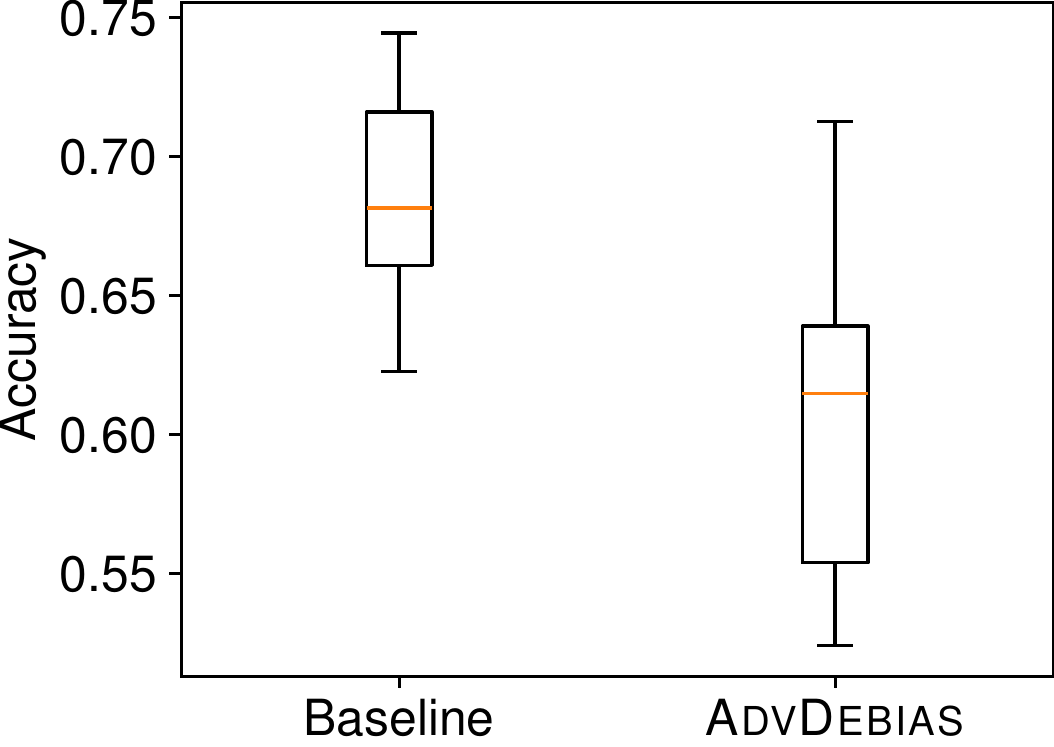}
    }
    \end{minipage}\\
   \caption{By imposing a constraint of fairness, \advdebias introduces a significant drop in $\targetmodel$'s accuracy, which is consistent with prior works% which rely on adversarial training
   ~\cite{debiase,NEURIPS2020_6b8b8e3b}.
   %Utility degradation for \advdebias: We observe a statistically significant drop in $\targetmodel$'s accuracy on using \advdebias. This is consistent with prior works which rely on adversarial training~\cite{debiase,NEURIPS2020_6b8b8e3b}.
   }
    \label{fig:utilityAdvDebias}
\end{figure}

\noindent\textbf{Trade-off with $\targetmodel$'s Utility.} We now quantify the impact on $\targetmodel$'s utility on using \advdebias (Figure~\ref{fig:utilityAdvDebias}). We report the model accuracy on $\testdata$ for $\targetmodel$ with and without \advdebias.
Results that there is a significant decrease in utility on using \advdebias across all datasets.
%
%indicate the model accuracy on $\testdata$ for $\targetmodel$ without \advdebias as \baseline. We find that there is a significant decrease in utility on using \advdebias across all datasets. 
However, this trade-off between group fairness and utility is consistent with prior works~\cite{debiase,NEURIPS2020_6b8b8e3b}.

\begin{figure}[!htb]
    \centering
    \begin{minipage}[b]{1\linewidth}
    \centering
    \subfigure[\census (\race)]{
    \includegraphics[width=0.49\linewidth]{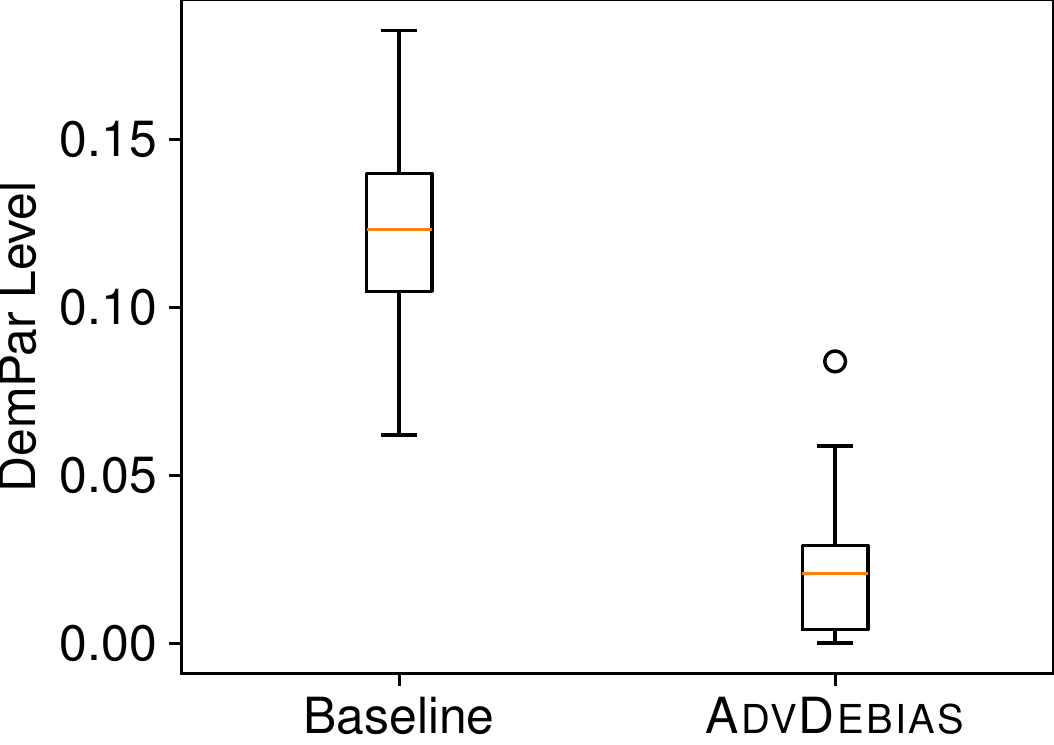}
    }%
    \subfigure[\census (\sex)]{
    \includegraphics[width=0.49\linewidth]{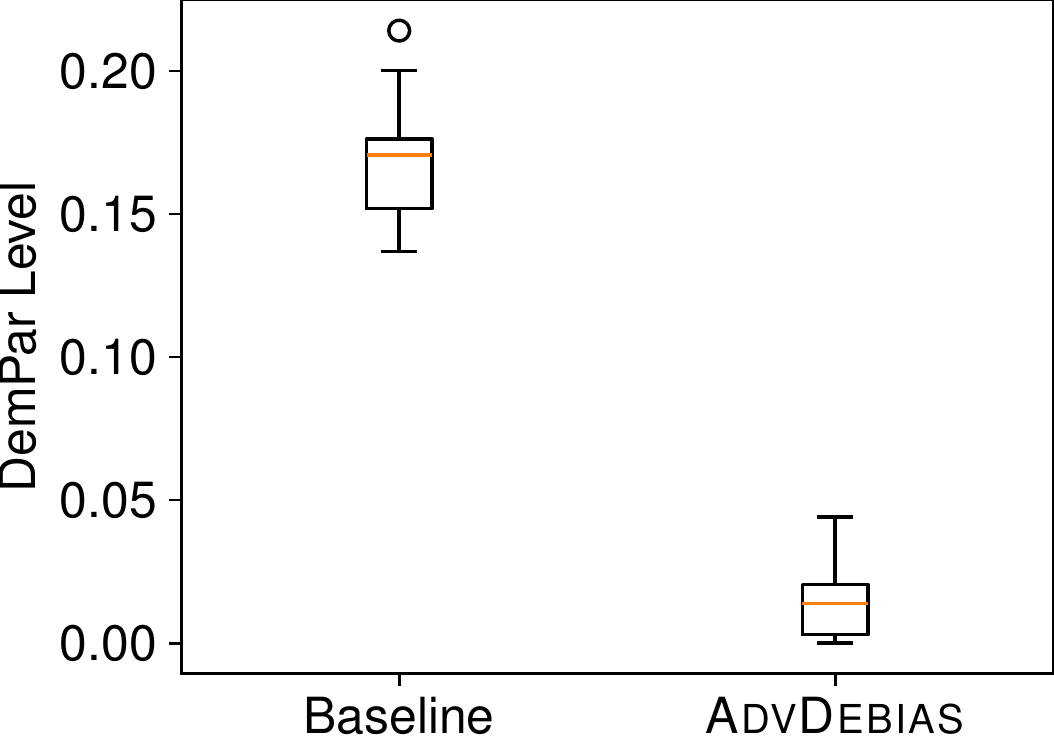}
    }
    \end{minipage}\\

    \begin{minipage}[b]{1\linewidth}
    \centering
    \subfigure[\meps (\race)]{
    \includegraphics[width=0.49\linewidth]{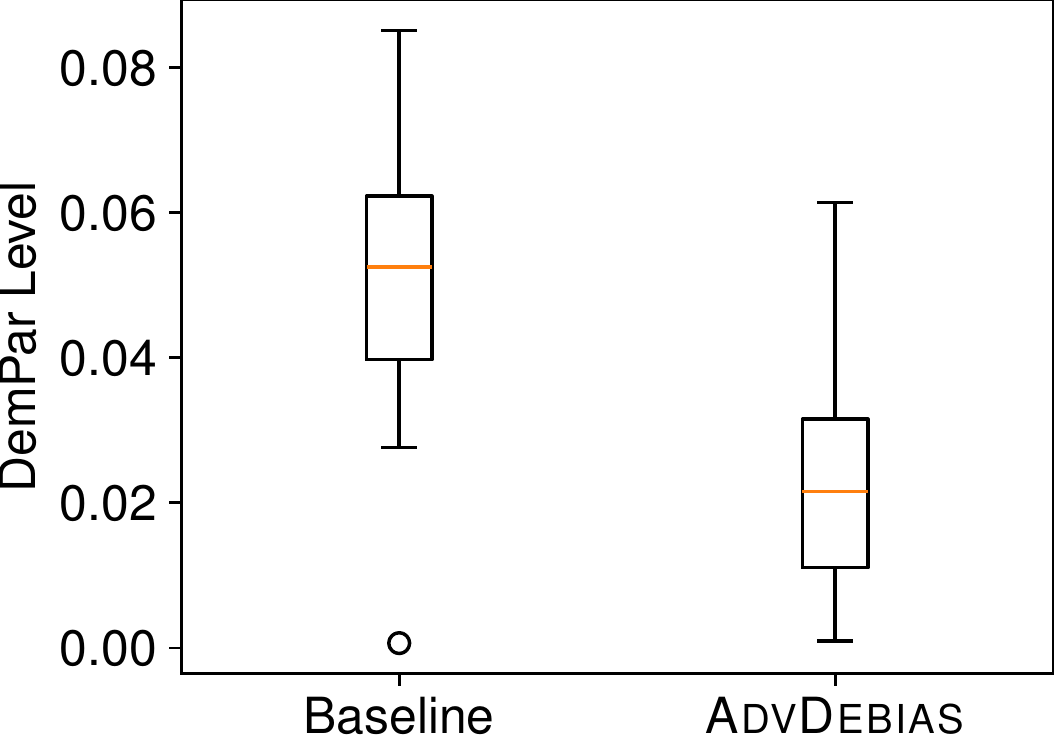}
    }%
    \subfigure[\meps (\sex)]{
    \includegraphics[width=0.49\linewidth]{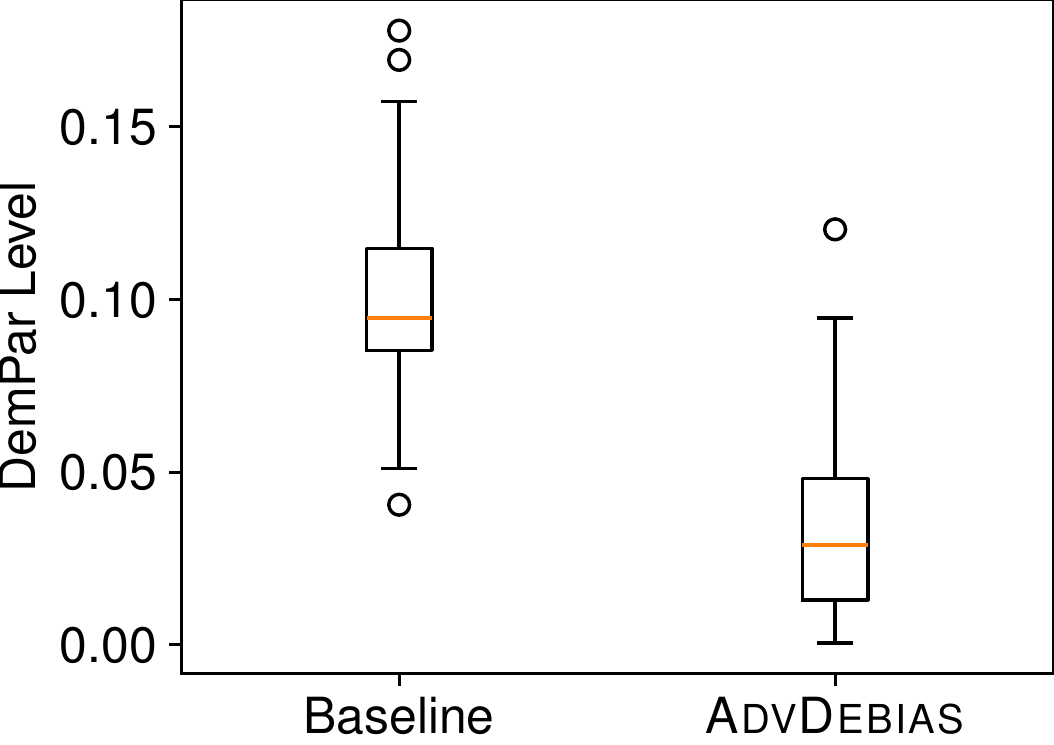}
    }
    \end{minipage}\\

    \caption{By reducing \dempar-Level close to 0 indicates that $\targetmodel$ is fair after \advdebias.
    %\dempar-Level for \advdebias: We observe that \dempar-Level is lower for \advdebias than the baseline indicating $\targetmodel$ is fair after \advdebias.
    }
    \label{fig:DemParAdvDebias}
\end{figure}

\noindent\textbf{Sanity Check for Fairness.} We now quantify the fairness as measured by \dempar-level with and without training $\targetmodel$ with \advdebias (Figure~\ref{fig:DemParAdvDebias}). Results show that $\targetmodel$ with \advdebias has significantly lower \dempar-level which is closer to zero as compared to baseline. Hence, \advdebias is effective for group fairness.

\noindent\textbf{Summary:} Similarly to \egd, considering our theoretical and empirical evaluation, we conclude that \advdebias induces attribute privacy. %This alignment allows us to calibrate the two dimensional trade-off between group fairness and utility.
This alignment allows us to account attribute privacy while only calibrating the two dimensional trade-off between group fairness and utility.

%% file: New/figures/fig_advdebias_attacc.tex
\begin{figure}[!htb]
    \centering
    \begin{minipage}[b]{\linewidth}
    \centering
    \subfigure[\census (\race)]{
    \includegraphics[width=0.48\linewidth]{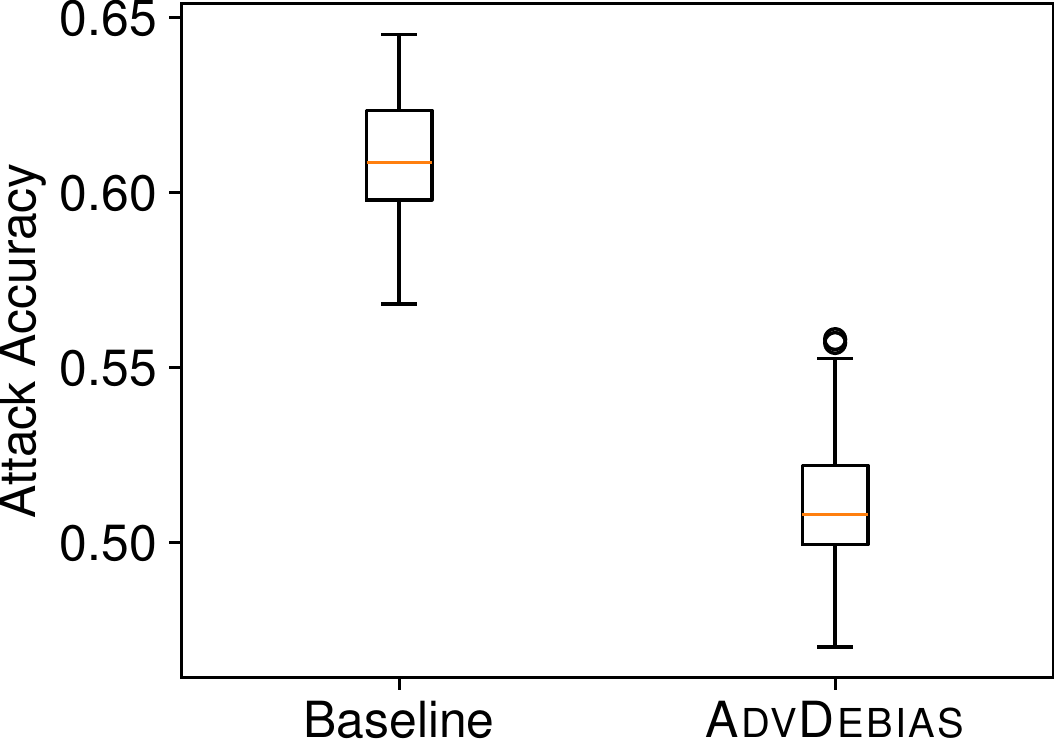}
    }%
    \subfigure[\census (\sex)]{
    \includegraphics[width=0.48\linewidth]{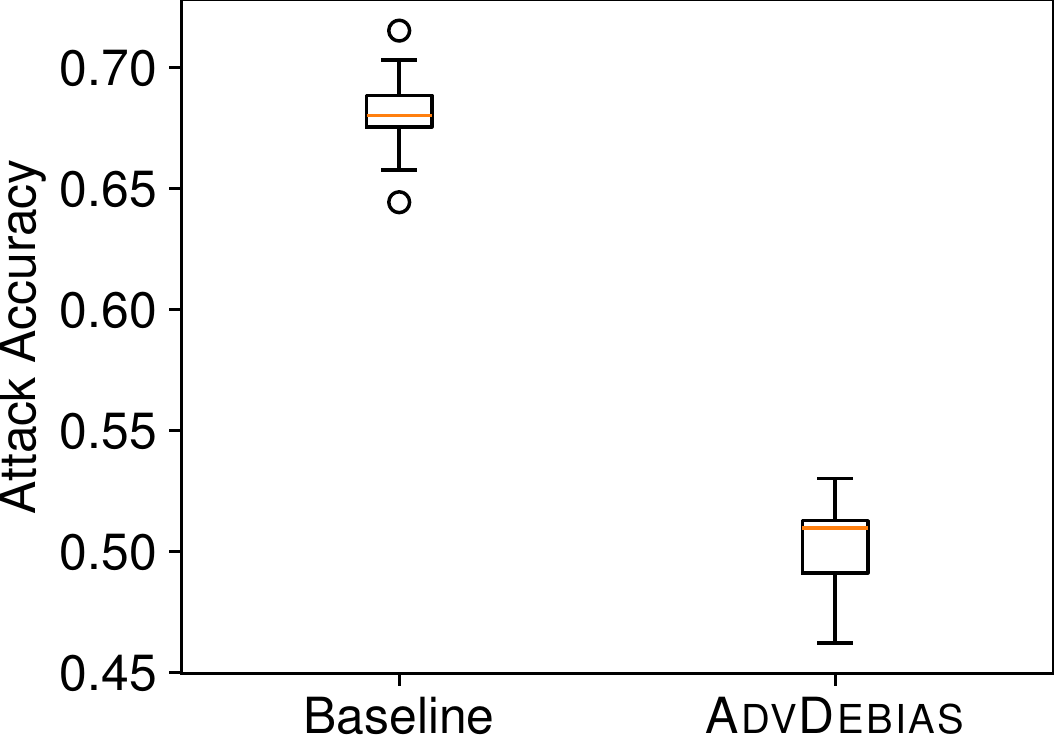}
    }
    \end{minipage}%

%    \begin{minipage}[b]{\linewidth}
%    \centering
%    \subfigure[\compas (\race)]{
%    \includegraphics[width=0.48\linewidth]{New/figures/advdebias/compas/compas_advdeb_attack_soft_experimental_race.pdf}
%    }%
%    \subfigure[\compas (\sex)]{
%    \includegraphics[width=0.48\linewidth]{New/figures/advdebias/compas/compas_advdeb_attack_soft_experimental_sex.pdf}
%    }
%    \end{minipage}%
    
    \begin{minipage}[b]{\linewidth}
    \centering
    \subfigure[\meps (\race)]{
    \includegraphics[width=0.48\linewidth]{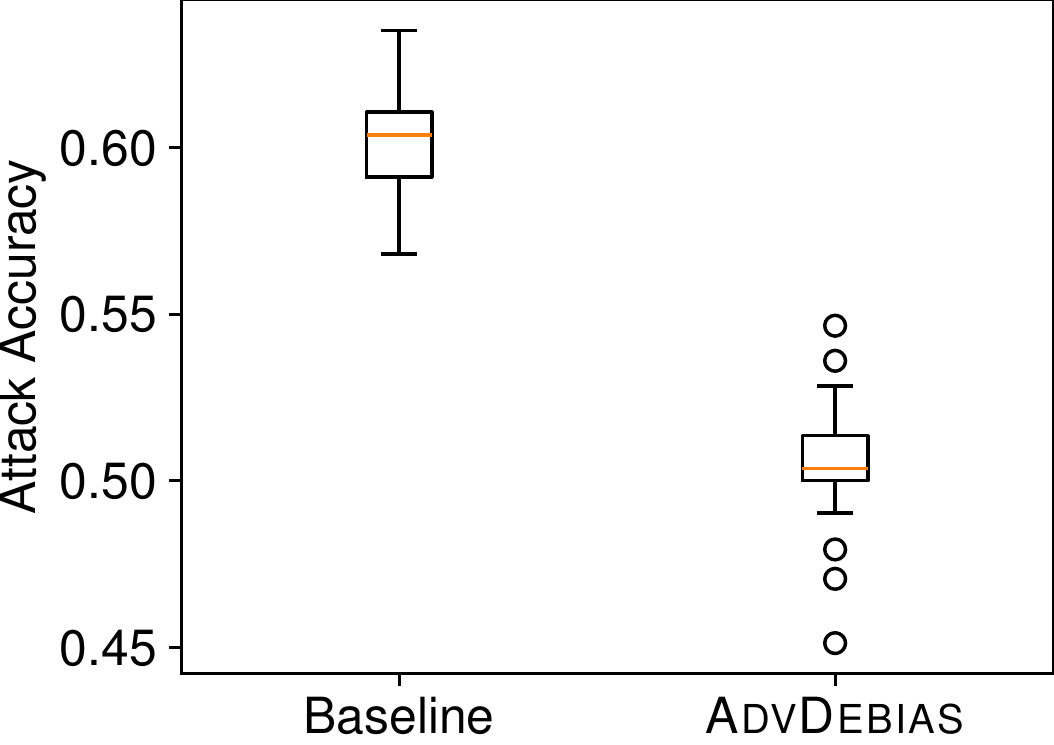}
    }%
    \subfigure[\meps (\sex)]{
    \includegraphics[width=0.48\linewidth]{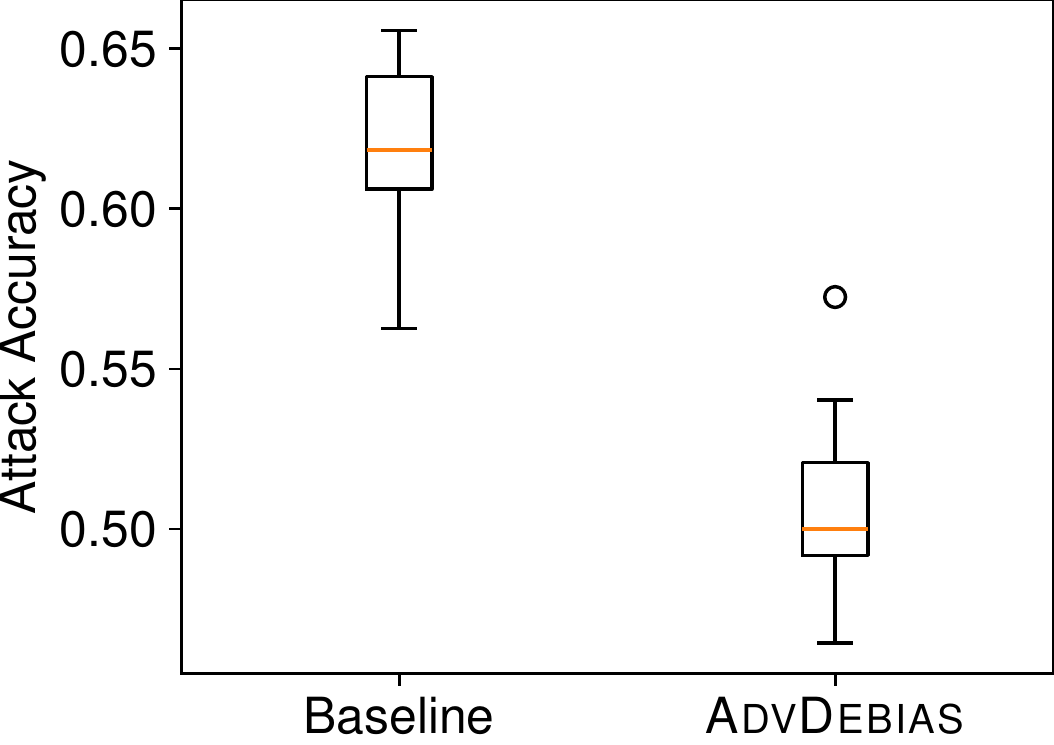}
    }
    \end{minipage}%

%   \begin{minipage}[b]{\linewidth}
%    \centering
%    \subfigure[\lfw (\race)]{
%    \includegraphics[width=0.48\linewidth]{New/figures/advdebias/lfw/lfw_advdeb_attack_soft_experimental_race.pdf}
%    }%
%    \subfigure[\lfw (\sex)]{
%    \includegraphics[width=0.48\linewidth]{New/figures/advdebias/lfw/lfw_advdeb_attack_soft_experimental_sex.pdf}
%    }
%    \end{minipage}%
    
    \caption{%For both \adaptiveAIASoft and \adaptiveAIAHard, w
    We observe that \advdebias reduces the attack accuracy to random guess ($\sim$50\%). %Additionally for \adaptiveAIAHard, the theoretical bound on attack accuracy (``Theory'') matches with the empirical results (``Empirical'').
    }
    \label{fig:AdaptAIADebias}
\end{figure}

%% file: 8related.tex
\section{Related Work}
\label{sec:related}

\noindent\textbf{\aia{s}} have been extensively studied in the context of online social media where \adv can leverage user's behavioural information~\cite{attinfSocial1,attinfSocial2,attinfSocial3,attinfSocial4,attinfSocial5,attinfSocial6}. In ML, \aia{s} are approached as data imputation challenges~\cite{fredrikson2,yeom,MehnazAttInf} and by leveraging intermediate model representations~\cite{Song2020Overlearning,Mahajan2020DoesLS,malekzadeh2021honestbutcurious}, which we extensively cover in Section~\ref{sec:back-attribute}.
Moreover, \adv can execute \aia using membership inference attacks as a subroutine~\cite{yeom}. Here, the success of \aia mirrors that of membership inference attacks, hinging on the degree of $\targetmodel$ overfitting. Nevertheless, Zhao et al.\cite{memprivNattpriv} establish this link only holds when membership inference attacks are highly effective. If not, \aia{s} offer minimal advantage, despite the model's vulnerability to membership inference attacks. Additionally, \aia{s} are effective against model explanations, presenting a compromise between attribute privacy and transparency~\cite{duddu2022inferring}.

\noindent\textbf{Defenses against \aia{s}} have limited prior literature. Attriguard~\cite{attriguard} safeguards sensitive attributes in social networks by introducing adversarial noise to the output predictions, compelling $\attackmodel$ to make incorrect classifications. For ML, adversarial training has proven effective in diminishing the inference of sensitive attributes within predictions~\cite{NEURIPS2020_6b8b8e3b}. This technique can additionally disentangle sensitive features from model representations~\cite{10.5555/3122009.3208010,10.5555/3327546.3327583,10.5555/3294771.3294827,pmlr-v80-madras18a,censoringadv}. An alternative avenue, variational autoencoders, was explored for minimizing \aia{s}\cite{10.5555/3042817.3042973}; however, Song et al.\cite{Song2020Overlearning} found these defences to be ineffective.

\noindent\textbf{Interactions between Privacy and Fairness} have been explored in prior work~\cite{chang2021privacy,indivfairness,dpaccdisp,dpfair,fioretto2022differential,pate2021fairness}. Chang et al.~\cite{chang2021privacy} demonstrate that applying in-processing fairness algorithms escalates overfitting, heightening vulnerability to \mia{s}, particularly affecting minority subgroups. Additionally, differential privacy and individual fairness share a common goal, with individual fairness encompassing differential privacy as a broader concept~\cite{indivfairness}. However, differential privacy and group fairness are at odds when there's performance discrepancy between minority and majority subgroups~\cite{dpaccdisp,dpfair,fioretto2022differential,pate2021fairness}. Ferry et al.~\cite{10136153} introduced \aia aimed at predicting sensitive attributes from a fair model. Their approach assumes that the group fairness algorithm for $\targetmodel$ is known to \adv, enabling calibration of the attacks. However, in practice, revealing model training specifics for confidentiality is improbable. In distinction from their approach, our attacks function within the context of \adv's blackbox understanding of $\targetmodel$, and the defenses we present effectively counter such attacks.

%% file: 9discussions.tex
%\section{Discussion}
\section{Discussion and Conclusion}
\label{sec:discussions}

This paper shows that there are no conflicts between group fairness and the specific notion of attribute privacy, which is lacking in the literature. Specifically, through an extensive empirical evaluation and theoretical guarantees, we show that group fairness imposed through the use of \advdebias and \egd satisfying \dempar is aligned with attribute privacy.
This alignment means that ensuring group fairness also ensures a protection against the attribute inference attack, which is lacking in the literature. 
However, ensuring fairness remains at the cost of model utility. 
To perform our extensive evaluation, we also propose new \aia{s} which outperform prior works. 
%Further, group fairness can act as a defense against these \aia{s} which is lacking in the literature.

Output indistinguishability is a general framework which can be applied to satisfy group fairness instead of explicitly optimizing for different fairness metrics~\cite{outIndist}. Any algorithm which falls into the category of output indistinguishability with respect to sensitive attributes is likely to satisfy attribute privacy as it will reduce the ability of \adv to perform accurate representation-based \aia{s} in a blackbox setting.

%\noindent\textbf{\egddp vs. \egdeo.} Recall that in Section~\ref{sec:EGD}, we use \dempar as the fairness constraint. We now theoretically show that \eo cannot mitigate \adaptiveAIAHard and by consequence \adaptiveAIASoft. 
%\begin{theorem}
%\label{th:eoo}
%If $\hat{Y}$ satisfies \eo for $Y$ and $S$ then the balanced accuracy of \adaptiveAIAHard is $\frac{1}{2}$ iff $Y$ is independent of $S$ or $\hat{Y}$ is independent of $Y$.
%\end{theorem}
%We prove the theorem in Appendix~\ref{app:egdeo}.
%Those two conditions are unlikely to happen with real-world.
%The condition of $Y$ being independent of $S$ was not observed for our datasets. We evaluate $|P(Y=0|S=0) - P(Y=0|S=1)|$ where a high value indicates $Y$ and $S$ are dependent. For \race and \sex, we found these values to be 0.05 and 0.27 (\compas), 0.20 and 0.13 (\census) and 0.07 and 0.13 (\meps) respectively.
%Further, the independence between $\hat{Y}$ and $Y$ means that $\targetmodel$ has random guess utility. 
%Hence, in practice, \eo aligns by reducing the risk to \aia{s} but does not \textit{perfectly align} as seen in \dempar by reducing \aia{s} to random guessing.
%The choice of fairness metric is important for \egd for perfect alignment. 

%\noindent\textbf{Choice of the fairness metric:}
%blablabla ... The choice of fairness metric is important for \egd for perfect alignment.

%\noindent\textbf{Non-binary Attributes:} 
Following the setting of prior \aia{s} and fairness~\cite{fredrikson2,Mahajan2020DoesLS,yeom,Song2020Overlearning,malekzadeh2021honestbutcurious,MehnazAttInf,debiase,NIPS2017_48ab2f9b,reductions}, we focus on the case where the sensitive attributes are binary. However, for \adaptiveAIASoft, the $\attackmodel$ can be trained to learn to infer non-binary attributes as well. For \adaptiveAIAHard, if we have $q$ classes and there are $p$ sensitive attributes, then there are $p^q$ functions to select from. While this is reasonable for small values, efficiently finding functions for large values is left as future work.

%\noindent\textbf{Relation to Output Indistinguishability~\cite{outIndist}:} Output indistinguishability is a general framework which can be applied to satisfy group fairness instead of explicitly optimizing for different fairness metrics~\cite{outIndist}. Any algorithm which falls into the category of output indistinguishability with respect to sensitive attributes is likely to satisfy attribute privacy as it will reduce the ability of \adv to perform accurate representation-based \aia{s} in a blackbox setting.

%\noindent\textbf{Summary:} With extensive empirical evaluation and theoretical guarantees, we show that group fairness is aligned with attribute privacy but at the cost of model utility. We propose new \aia{s} which outperform prior works. Further, group fairness can act as a defense against these \aia{s} which is lacking in the literature.

%% file: 10appendix.tex
\appendix

\section*{Appendix}

We provide more details about the notations (Appendix~\ref{app:notations}, plots for distinguishability in predictions (Section~\ref{app:distinguishability}), proofs and empirical results for \egd (Appendix~\ref{app:egd}) and \advdebias (Appendix~\ref{app:advdebias}), proof for showing \egdeo is not aligned (Appendix~\ref{app:egdeo}), and finally impact of \adv's distribution on attack success.

\section{Notations}\label{app:notations}

We introduce some notations from probability theory which use in the paper. For a set $A$, the set of subsets of $A$ is given by $\mathcal{P}(A)$. 
Each element $a \in \mathcal{P}(A)$ is such that $a \subset A$.
A tribe $\mathcal{A}$ is a subset of $\mathcal{P}(A)$ that contains: $\emptyset$, $A$, and is stable by complementary and countable union.
Then, we call $(A,\mathcal{A})$ a \textit{measure space}.
A measure $d$ is a function $d$:$\mathcal{A}$ $\rightarrow$ $[0,+\infty]$ such that $d(\emptyset) = 0$ and $d\left(\bigcup_{i\in \mathbb{N}} A_i\right) = \sum_{i\in \mathbb{N}}d(A_i)$ for any $(A_1, A_2, \cdots) \in \mathcal{A}^\mathbb{N} $ with $\forall (i,j) A_i\cap A_j = \emptyset$.
We then call $(A, \mathcal{A}, d)$ a \textit{measurable space}.
Any function mapping $A$ to $B$ is called a \textit{measurable function} if $\forall b\in\mathcal{B}$~$f^{-1}(b)\in\mathcal{A}$ and we note $f:(A, \mathcal{A})\rightarrow (B, \mathcal{B})$ or $f:(A, \mathcal{A},d)\rightarrow (B, \mathcal{B})$

In the special case where $d(A) = 1$ we call $d$ a \textit{probability measure}.
We then call $(A,\mathcal{A},d)$ a \textit{probability space} and the measurable functions on this space are \textit{random variables}.

$\hat{S} = 1_{[0.5,1]}\circ \change{\attackmodel}\circ \change{\targetmodel}\circ X$ translates to $S$ is equal to the composition of four functions: the random variable $X$, $\targetmodel$, $\attackmodel$ and the indicator function of the set $[0.5,1]$.
We summarize all important notations used in our paper in Table~\ref{tab:notations}.

\begin{table}[!htb]
\caption{Summary of notations used in the paper.}
\begin{center}
\footnotesize
\resizebox{\columnwidth}{!}{%
\begin{tabular}{ | c | c | }
\hline
\textbf{Notation} & \textbf{Meaning}\\
\hline
 \adv & Adversary \\  
 $\targetmodel$ & Target model being attacked \\
 $\targetmodel\circ X$ or $\targetmodel(X\change{(\omega)})$ & Predictions\\
 $\hat{Y}_h(\omega)$ & Prediction (Hard label) \\
 $\hat{Y}_s(\omega)$ & Prediction (Soft label) \\
 $\attackmodel$ & \adv's Attack model \\  
 $f_{disc}$ & Discriminator network for \advdebias \\
 $\traindata$ & Data used to train the target model \\  
 $\testdata$ & Data used to test the target model \\  
 $\auxdata$ & Auxiliary data available to the adversary \\
 $\auxtraindata$ & \adv's dataset to train attack model\\
 $\auxtestdata$ & \adv's dataset to evaluate attack model\\
 $\upsilon$ & Threshold on $\attackmodel$'s output\\
 $X$ & Random variable of non-sensitive attributes\\
 $Y$ & Random variable of classification labels\\
 $S$ & Random variable of sensitive attributes\\
 $X(\omega), Y(\omega), S(\omega)$ & Specific instances parameterized by $\omega$\\
 $(\Omega, \mathcal{T}, \mathcal{P})$ & Probability space\\
 $(E, \mathcal{U})$ & Measurable space\\
 $\mathcal{B}$ & Borel tribe ($\sigma$-algebra)\\
 \hline
\end{tabular}
}
\end{center}
\label{tab:notations}
\end{table}

\section{Prediction Distinguishability}
\label{app:distinguishability}

\begin{figure}[!htb]
     \centering
     \begin{minipage}[b]{0.9\linewidth}
     \centering
     \subfigure[\census]{
     \includegraphics[width=1\linewidth]{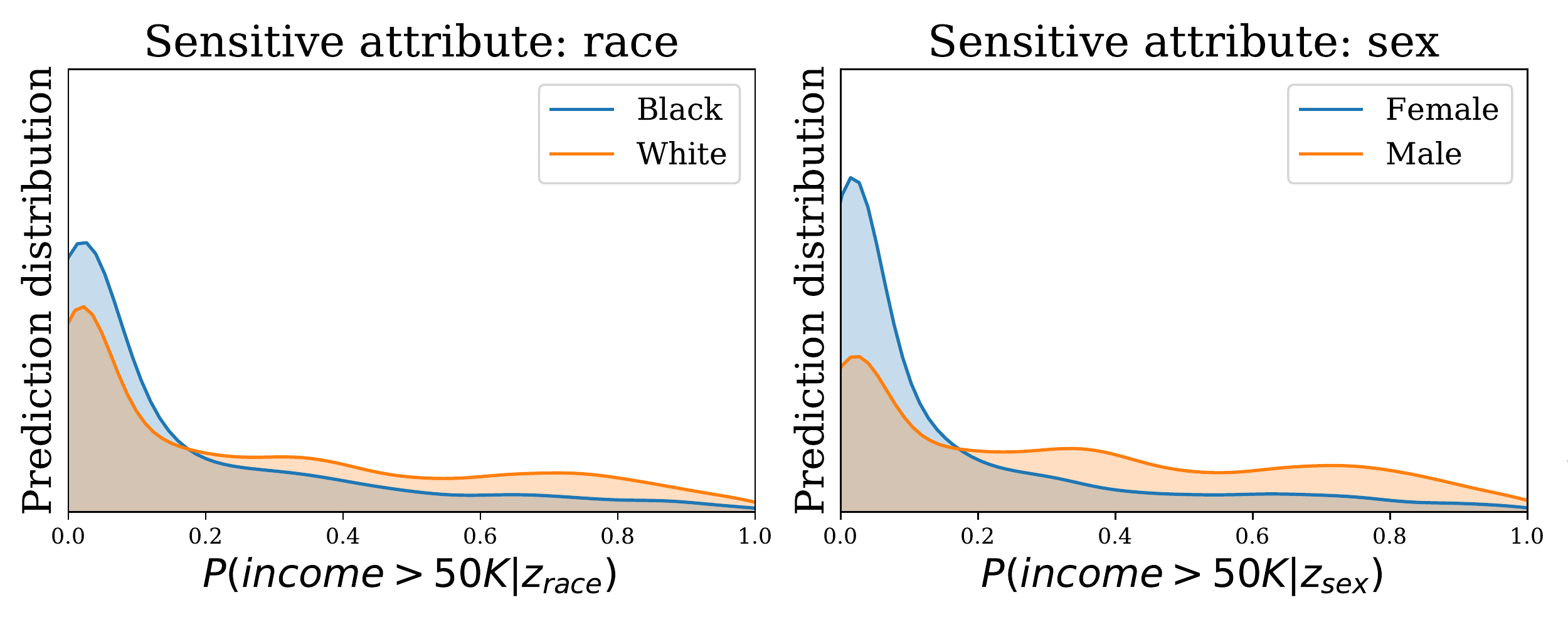}
     }
     \end{minipage}
     \begin{minipage}[b]{0.9\linewidth}
     \centering
     \subfigure[\compas]{
     \includegraphics[width=1\linewidth]{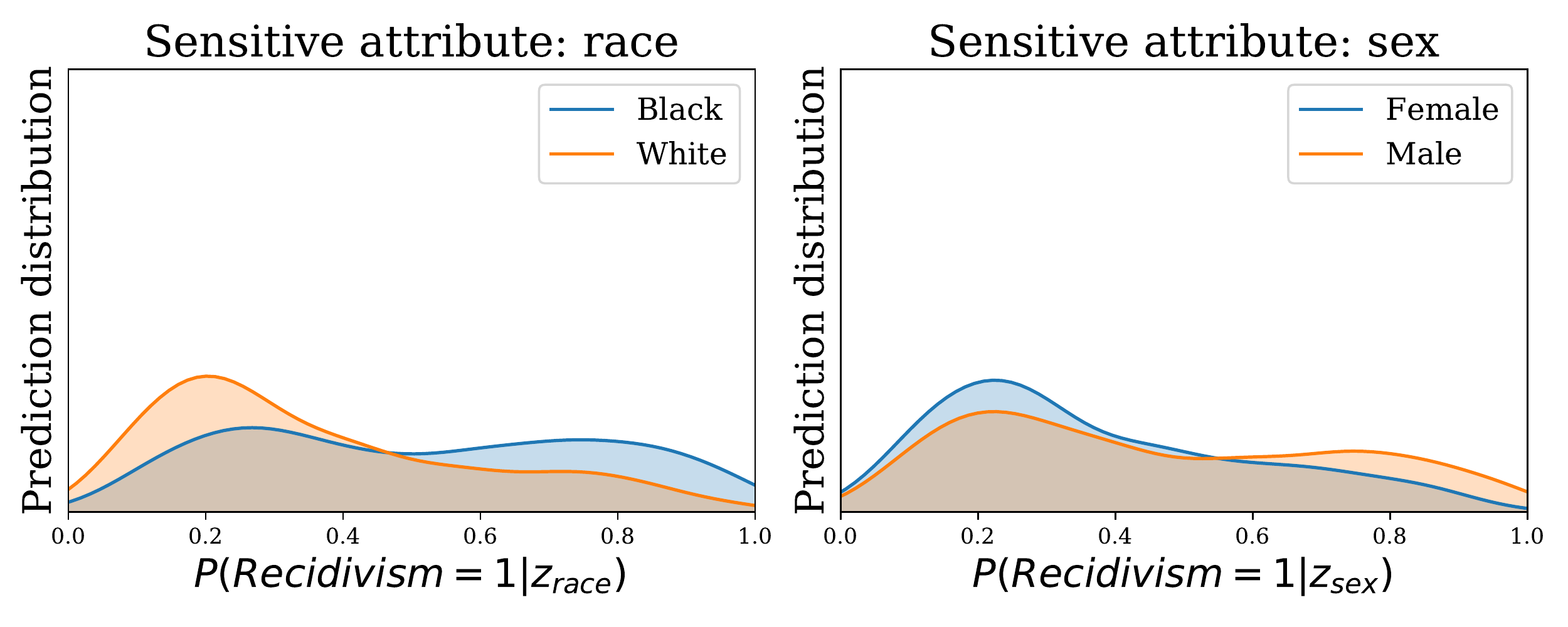}
     }
     \end{minipage}
     \begin{minipage}[b]{0.9\linewidth}
     \centering
     \subfigure[\meps]{
     \includegraphics[width=1\linewidth]{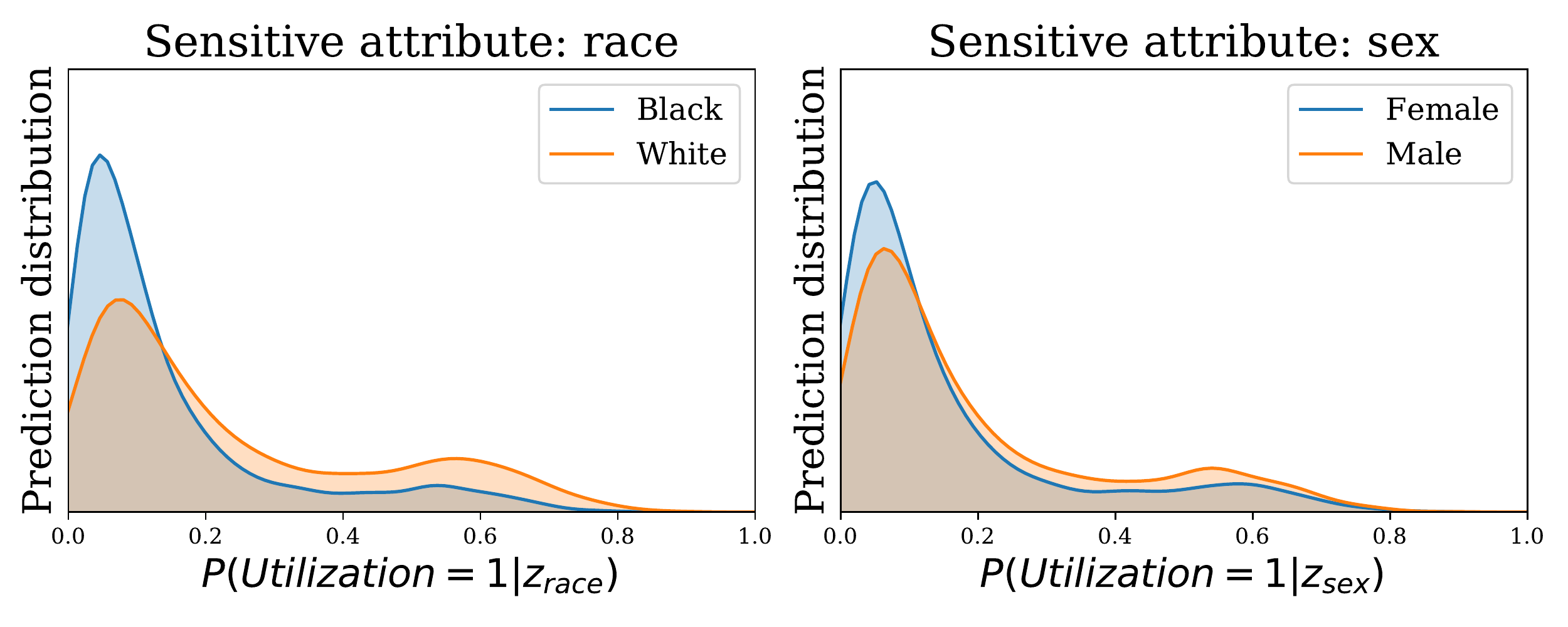}
     }
     \end{minipage}
     \begin{minipage}[b]{0.9\linewidth}
     \centering
     \subfigure[\lfw]{
     \includegraphics[width=1\linewidth]{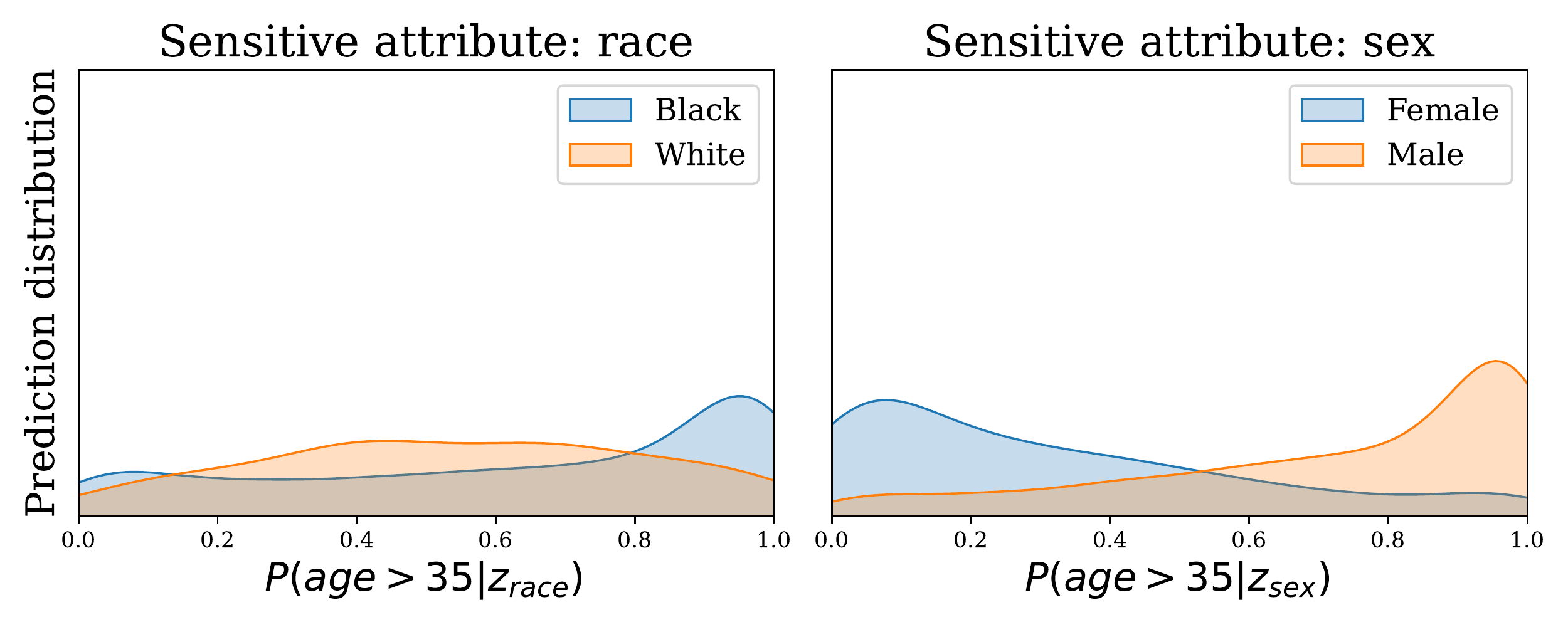}
     }
     \end{minipage}
     \caption{Distinguishable output predictions for two subgroups identified by \race and \sex. \adv exploits this to infer $S(\omega)$ for an arbitrary input data record.}
     \label{fig:distribution}
 \end{figure}

To better understand the impact of bias in $\traindata$ on predictions, we plot the conditional densities of $\targetmodel\circ X$ for \race and \sex (each taking values $S=0$ and $S=1$) in Figure~\ref{fig:distribution}. 
We observe that $\targetmodel(X(\omega))$ are distinguishable for different values of $S(\omega)$, even when it is not included in the input. For \census, the distributions for both \race and \sex corresponding to the minority group (e.g., blacks and females) have a larger peak for lower output probability values.
This indicates that the classifier is likely to predict $<$50K salary for non-white and females compared to members of the majority group.
For \compas, $\targetmodel$ is likely to predict males and blacks to re-offend more compared to females or whites.
Furthermore, in \meps the distributions look similar but the use of medical resources is lower for blacks and females.
Finally for \lfw, males and blacks are likely to be predicted as being $>$35.
To account for the skew in predictions and class imbalance, we want to use an adaptive threshold.

\section{\egddp}
\label{app:egd}

We first present the proof for the Theorem~\ref{th:dpgood}. We then present empirical results to the success of \adaptiveAIA showing that $\targetmodel$ trained with \egddp is fair compared to the baseline as a sanity check. Finally, we present the impact on $\targetmodel$'s utility on using \egddp. 

\input{New/proofs/proof_egd_dp}

\begin{figure}[!htb]
    \centering

    \begin{minipage}[b]{1\linewidth}
    \centering
    \subfigure[\compas (\race)]{
    \includegraphics[width=0.49\linewidth]{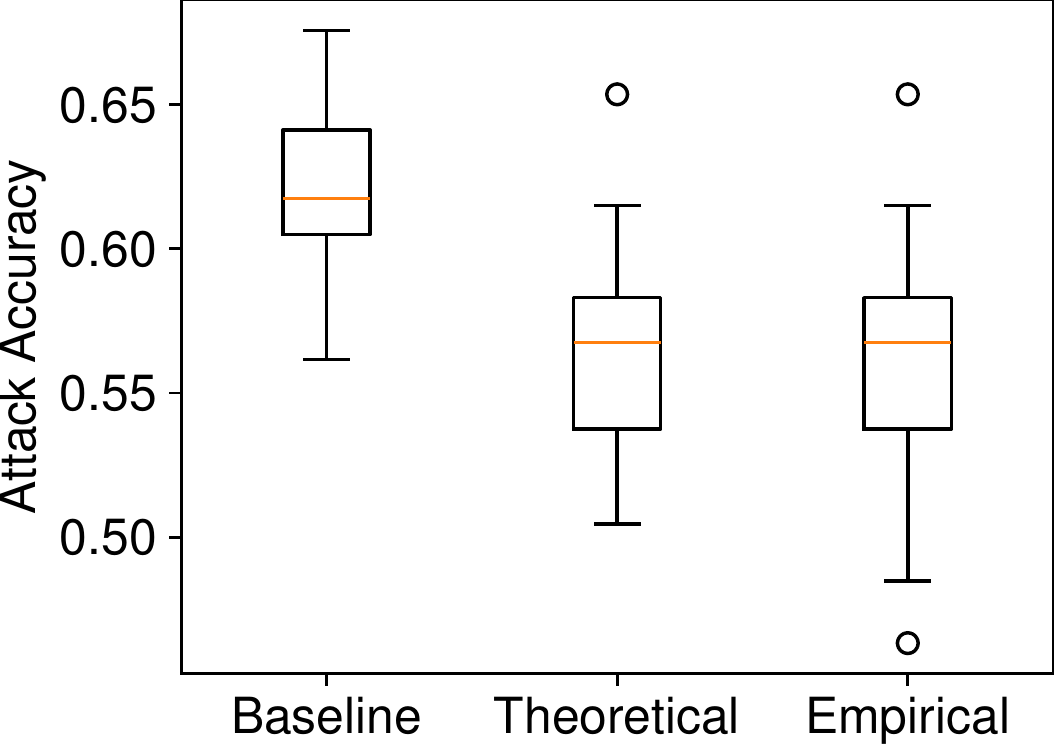}
    }%
    \subfigure[\compas (\sex)]{
    \includegraphics[width=0.49\linewidth]{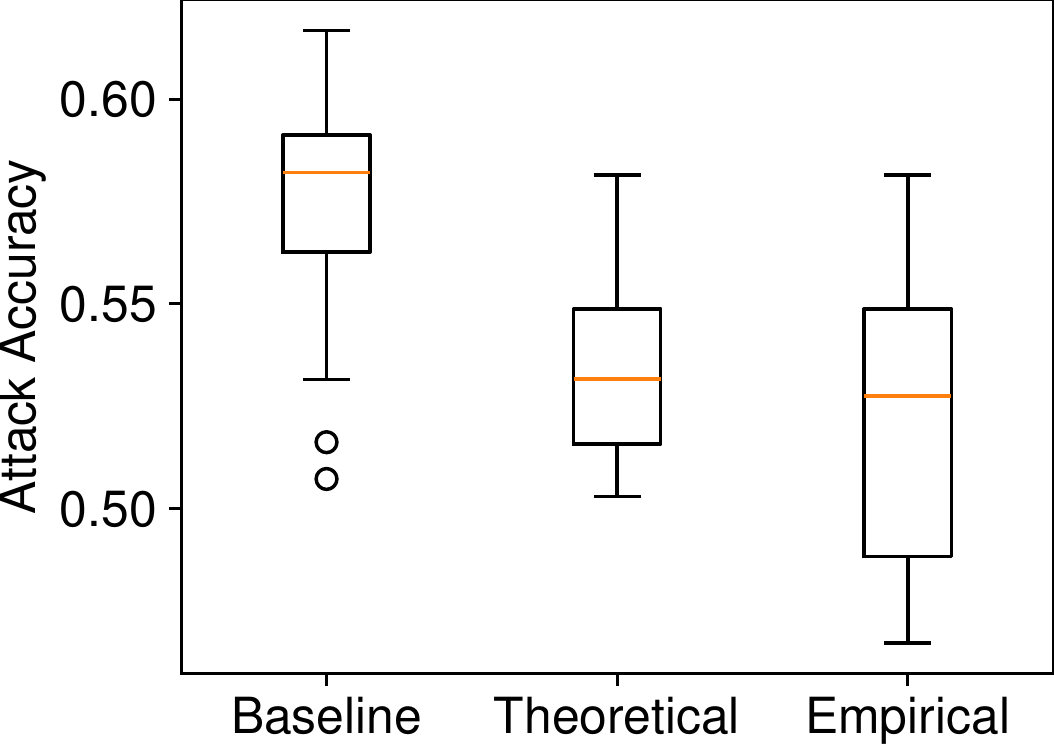}
    }
    \end{minipage}\\

    \begin{minipage}[b]{1\linewidth}
    \centering
    \subfigure[\lfw (\race)]{
    \includegraphics[width=0.49\linewidth]{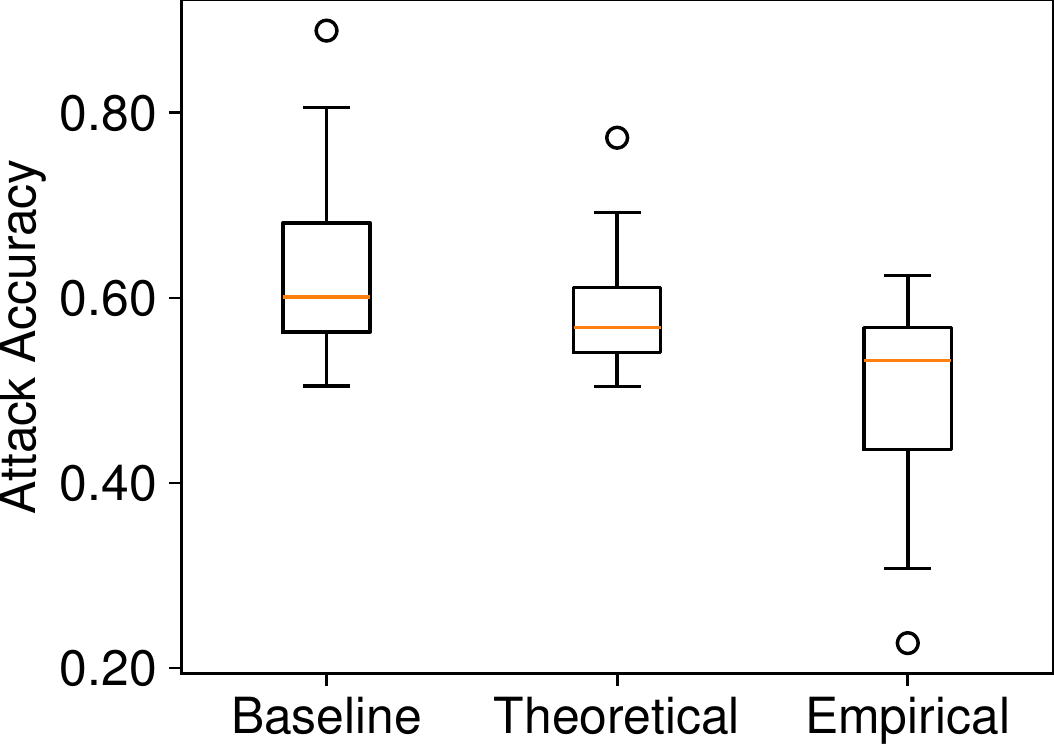}
    }%
    \subfigure[\lfw (\sex)]{
    \includegraphics[width=0.49\linewidth]{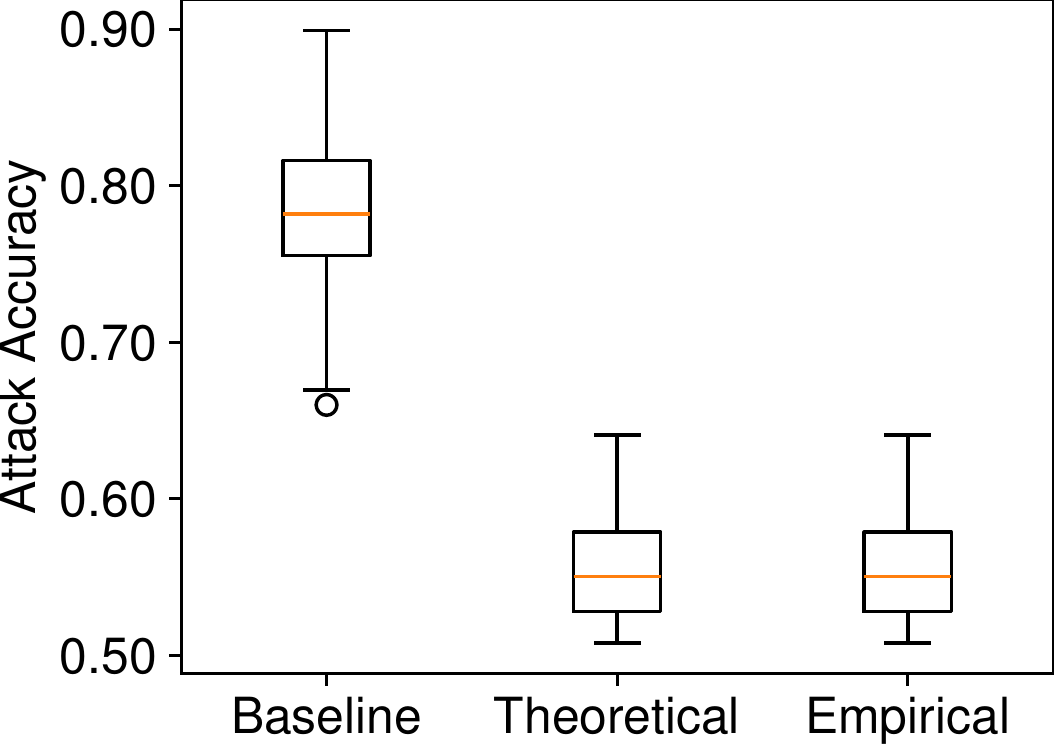}
    }
    \end{minipage}

    \caption{For \adaptiveAIAHard, we observe that \egd reduces the attack accuracy to random guess ($\sim$50\%).}
    \label{fig:AdaptAIAEGD}
\end{figure}

\noindent\textbf{Additional Results: \adaptiveAIAHard}. We present additional results for \compas and \lfw evaluating the effectiveness of \egddp against \adaptiveAIAHard. Similar to Section~\ref{sec:EGD}, we observe that the success of \adaptiveAIAHard decreases on using \egddp compared to \baseline (Figure~\ref{fig:AdaptAIAEGD}). Additionally, the theoretical bound on attack accuracy (\theoretical) matches with the empirical results (\empirical).

\noindent\textbf{Additional Results: Impact on Utility.} We now present the impact on $\targetmodel$'s utility on using \egddp for \compas and \lfw (Figure~\ref{fig:utilityEGD}). We find that the utility degrades on using \egddp which is in line with findings from prior work~\cite{reductions}. In other words, group fairness comes at the cost of accuracy.

\input{New/figures/fig_egd_utility}

\noindent\textbf{Additional Results: Fairness.} For \compas and \lfw, we present the \dempar-level with and without training $\targetmodel$ with \egddp (Figure~\ref{fig:DemParegd2}). We observe that $\targetmodel$ with \egddp has significantly lower \dempar-level which is closer to zero as compared to \baseline. Hence, \egddp is effective to achieve group fairness.

\input{New/figures/fig_egd_fair}

\section{\advdebias}
\label{app:advdebias}

We first present the proof of Theorem~\ref{th:advdebias} followed by results showing that models trained with \advdebias is indeed fair as a sanity check. Finally, we show the impact of using \advdebias on $\targetmodel$'s utility.

We present the theorem and proof which bounds \aia accuracy to random guess on using \advdebias.
\input{New/proofs/proof_advdebias}

\begin{figure}[!htb]
    \centering
    \begin{minipage}[b]{\linewidth}
    \centering
    \subfigure[\compas (\race)]{
    \includegraphics[width=0.48\linewidth]{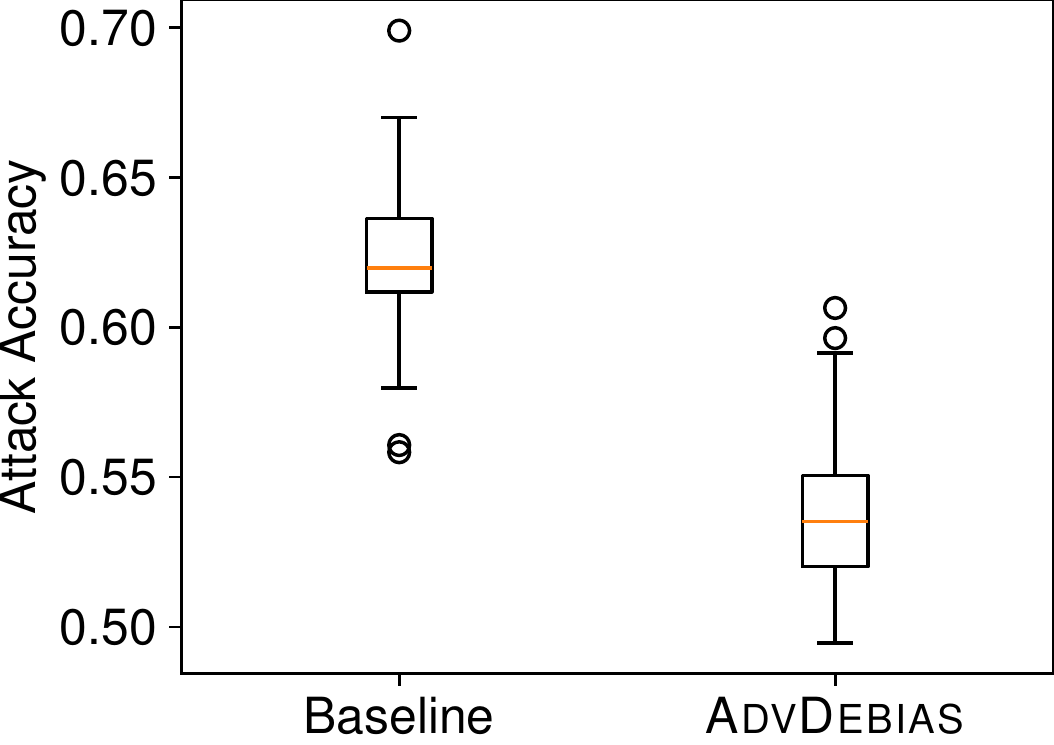}
    }%
    \subfigure[\compas (\sex)]{
    \includegraphics[width=0.48\linewidth]{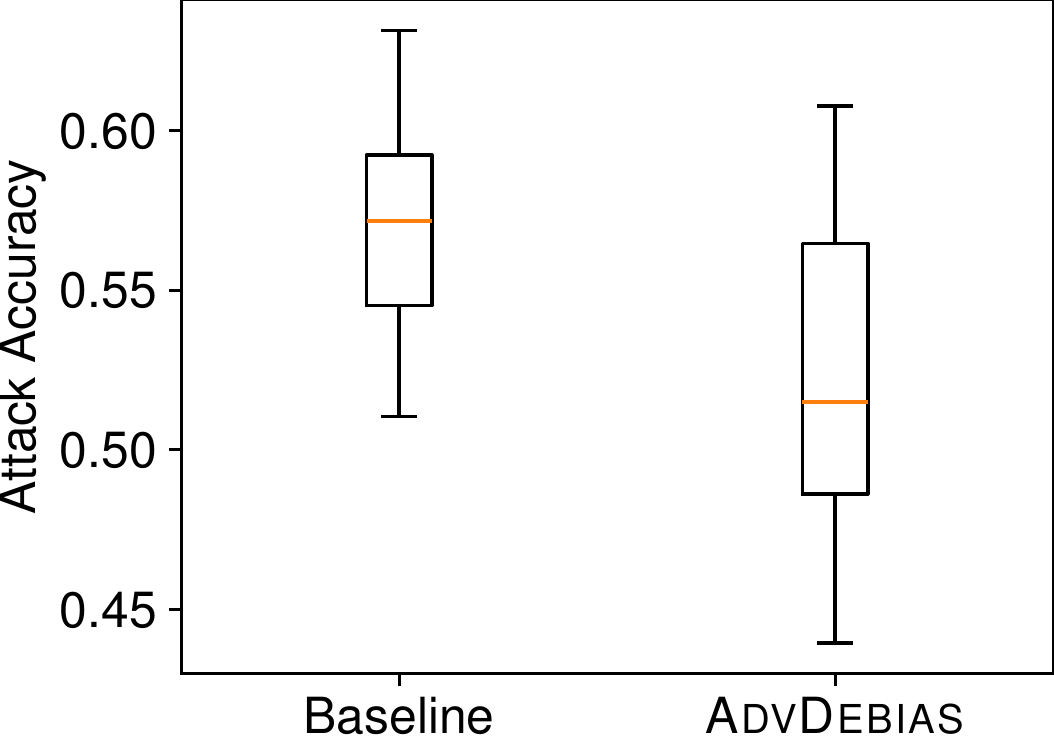}
    }
    \end{minipage}%

   \begin{minipage}[b]{\linewidth}
    \centering
    \subfigure[\lfw (\race)]{
    \includegraphics[width=0.48\linewidth]{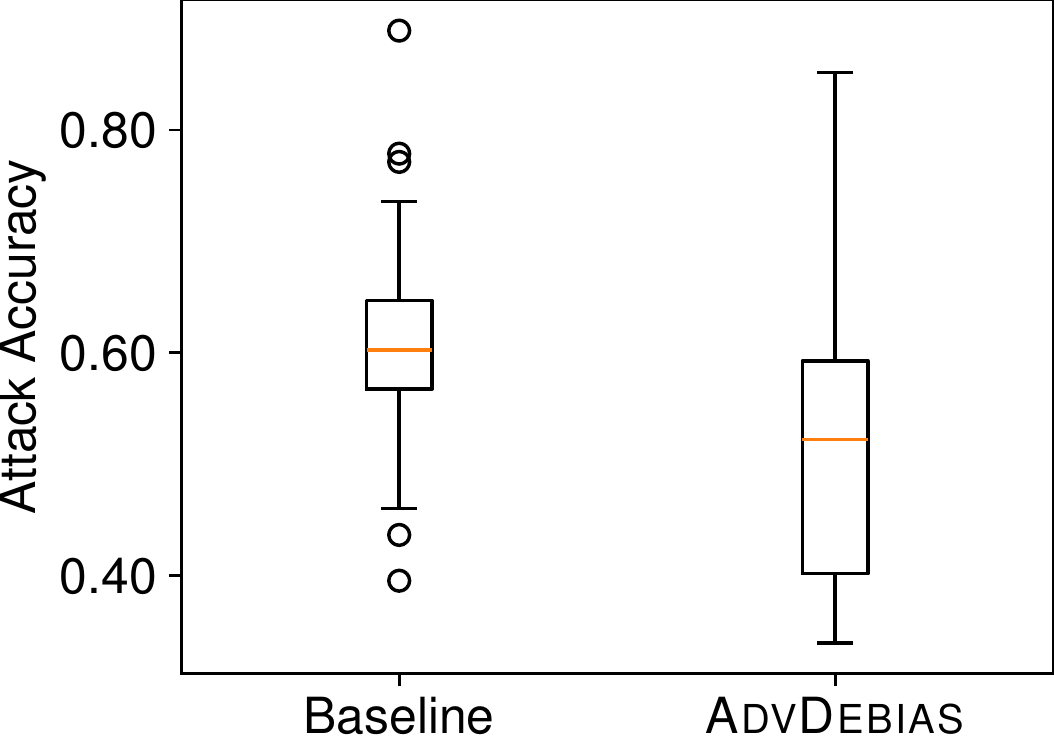}
    }%
    \subfigure[\lfw (\sex)]{
    \includegraphics[width=0.48\linewidth]{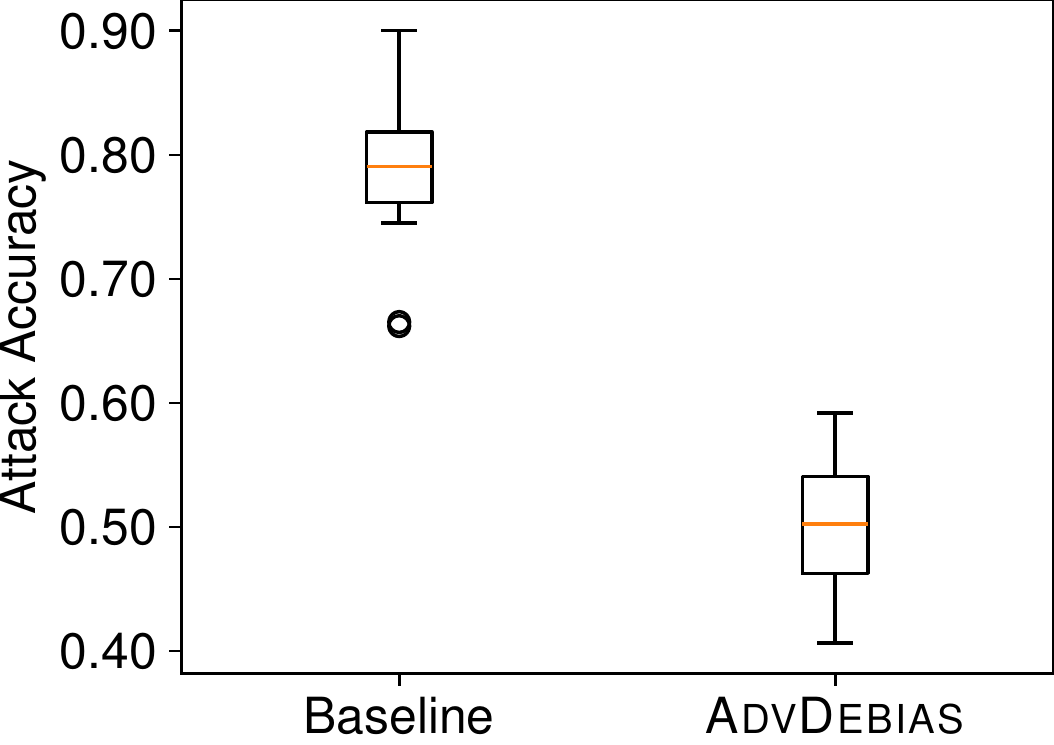}
    }
    \end{minipage}%
    
    \caption{%For both \adaptiveAIASoft and \adaptiveAIAHard, w
    We observe that \advdebias reduces the attack accuracy to random guess ($\sim$50\%). %Additionally for \adaptiveAIAHard, the theoretical bound on attack accuracy (``Theory'') matches with the empirical results (``Empirical'').
    }
    \label{fig:AdaptAIASoftDebias2}
\end{figure}

\noindent\textbf{Additional Results: \adaptiveAIASoft.} We present additional results for \compas and \lfw in Figure~\ref{fig:AdaptAIASoftDebias2} showing that \advdebias is effective to mitigate \adaptiveAIASoft consistent with the results in Section~\ref{sec:advdebias}.

\begin{figure}[!htb]
    \centering
    \begin{minipage}[b]{\linewidth}
    \centering
    \subfigure[\census (\race)]{
    \includegraphics[width=0.48\linewidth]{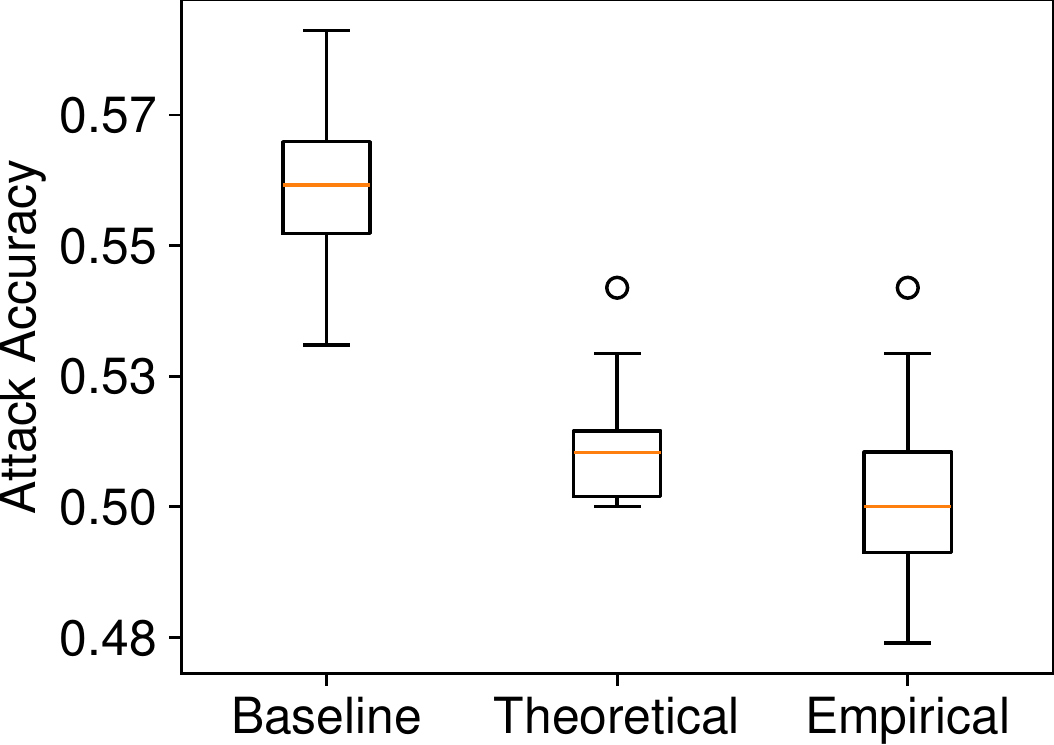}
    }%
    \subfigure[\census (\sex)]{
    \includegraphics[width=0.48\linewidth]{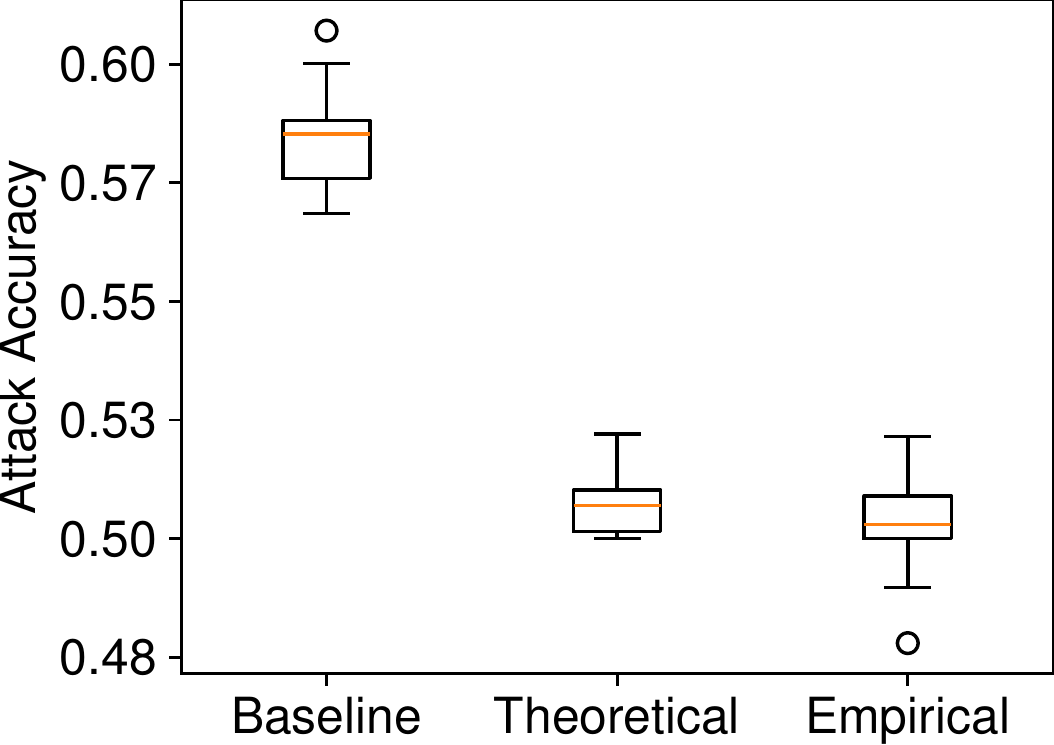}
    }
    \end{minipage}\\

    \begin{minipage}[b]{\linewidth}
    \centering
    \subfigure[\compas (\race)]{
    \includegraphics[width=0.48\linewidth]{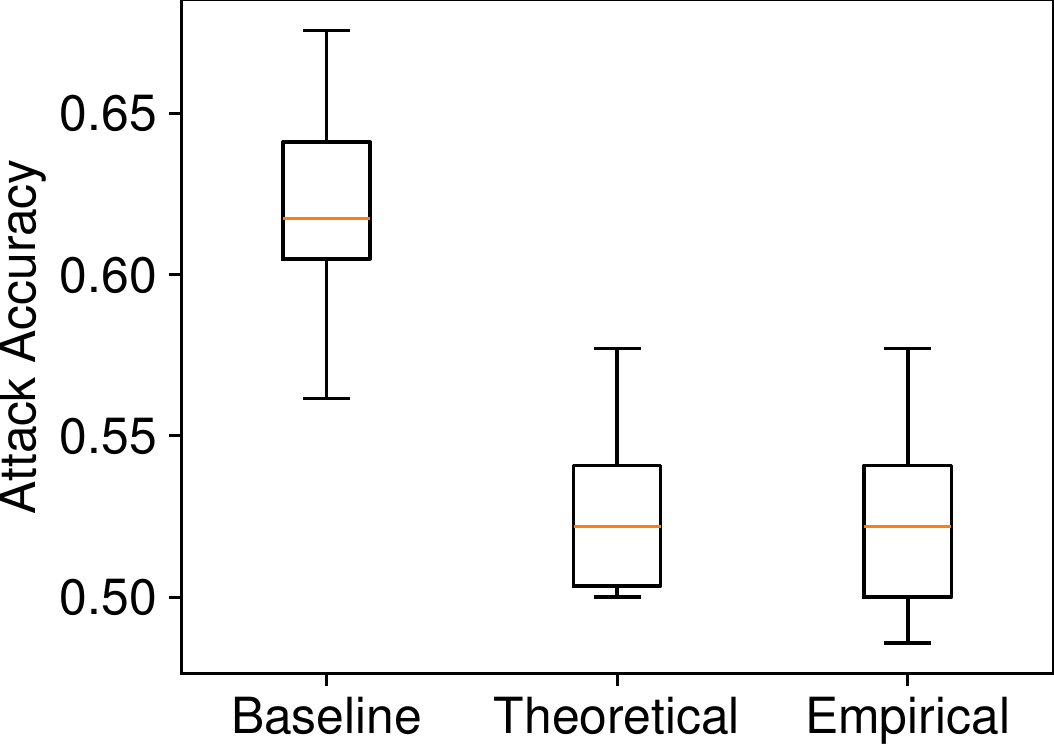}
    }%
    \subfigure[\compas (\sex)]{
    \includegraphics[width=0.48\linewidth]{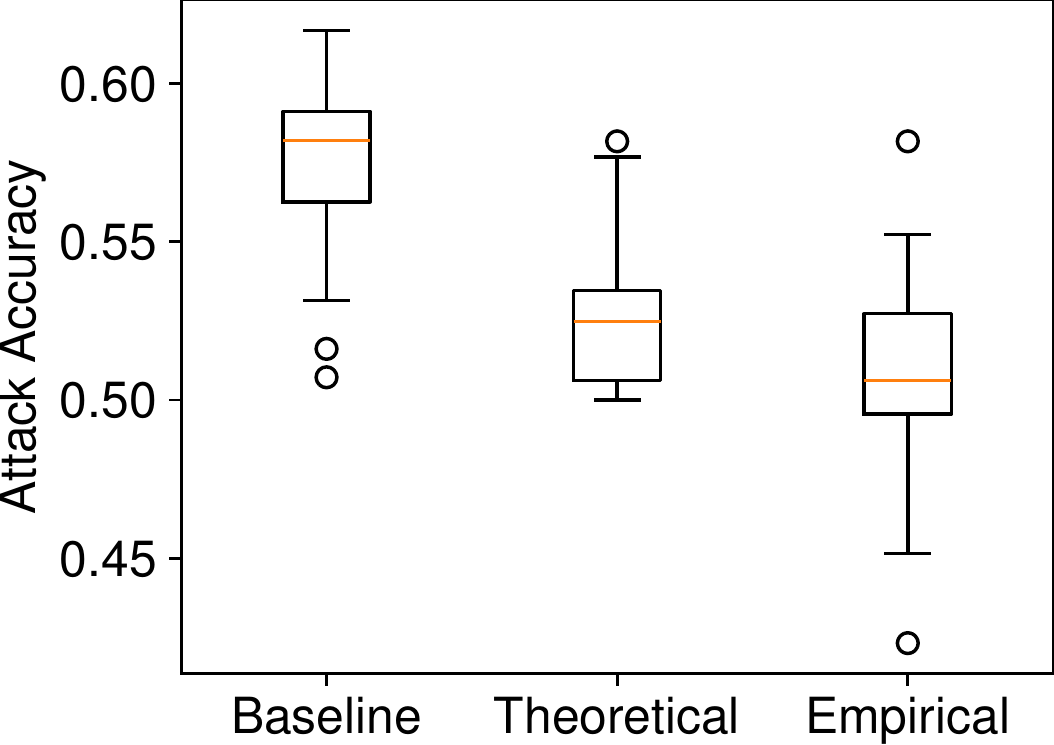}
    }
    \end{minipage}\\

    \begin{minipage}[b]{\linewidth}
    \centering
    \subfigure[\meps (\race)]{
    \includegraphics[width=0.48\linewidth]{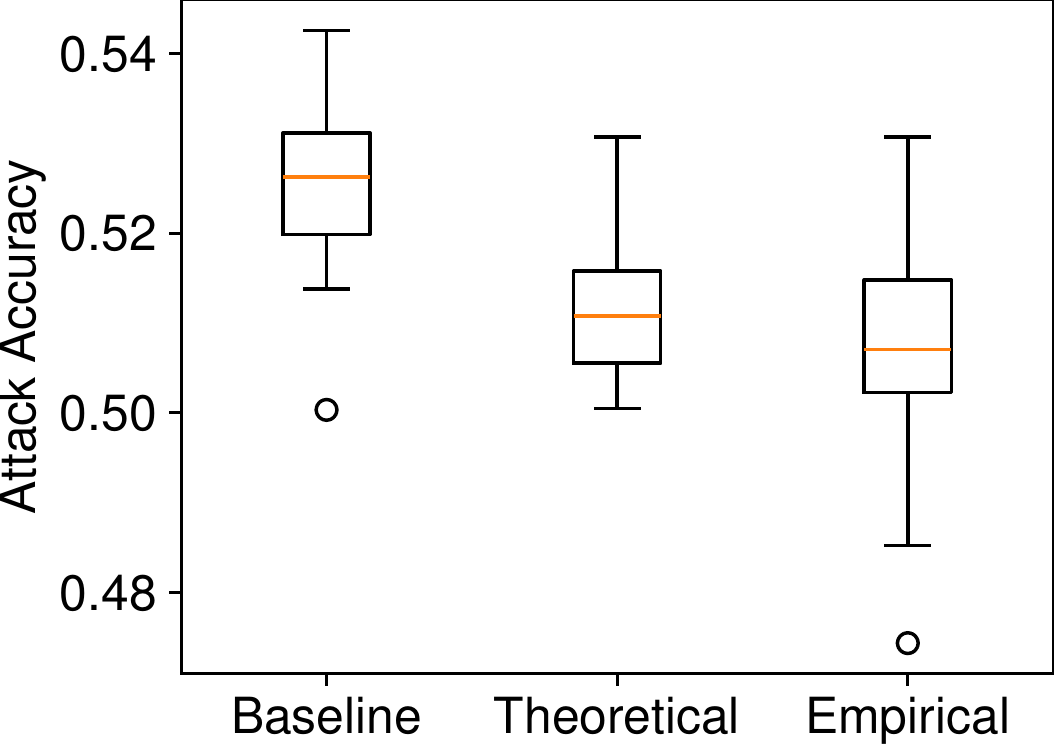}
    }%
    \subfigure[\meps (\sex)]{
    \includegraphics[width=0.48\linewidth]{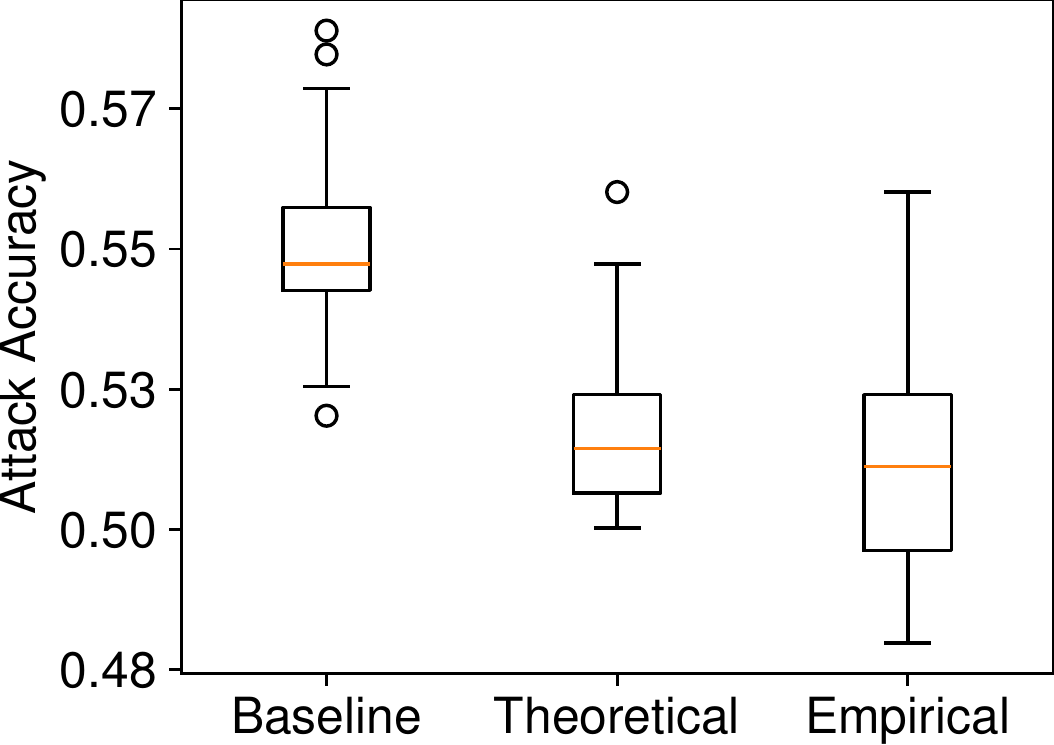}
    }
    \end{minipage}\\
    
    \begin{minipage}[b]{\linewidth}
    \centering
    \subfigure[\lfw (\race)]{
    \includegraphics[width=0.48\linewidth]{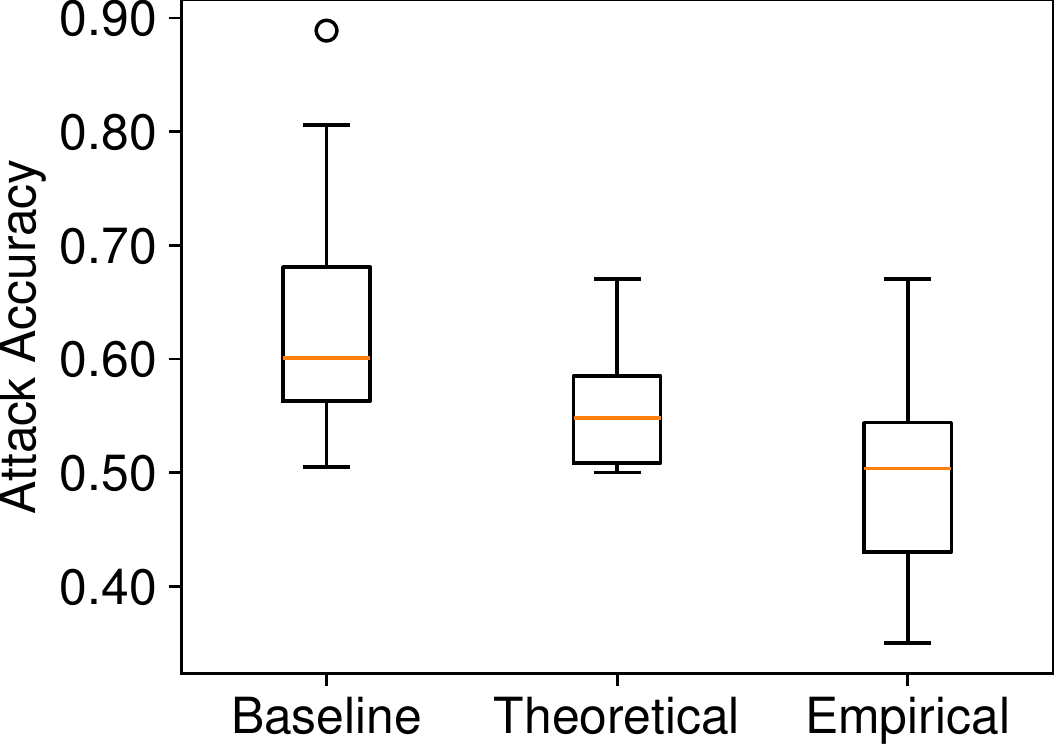}
    }%
    \subfigure[\lfw (\sex)]{
    \includegraphics[width=0.48\linewidth]{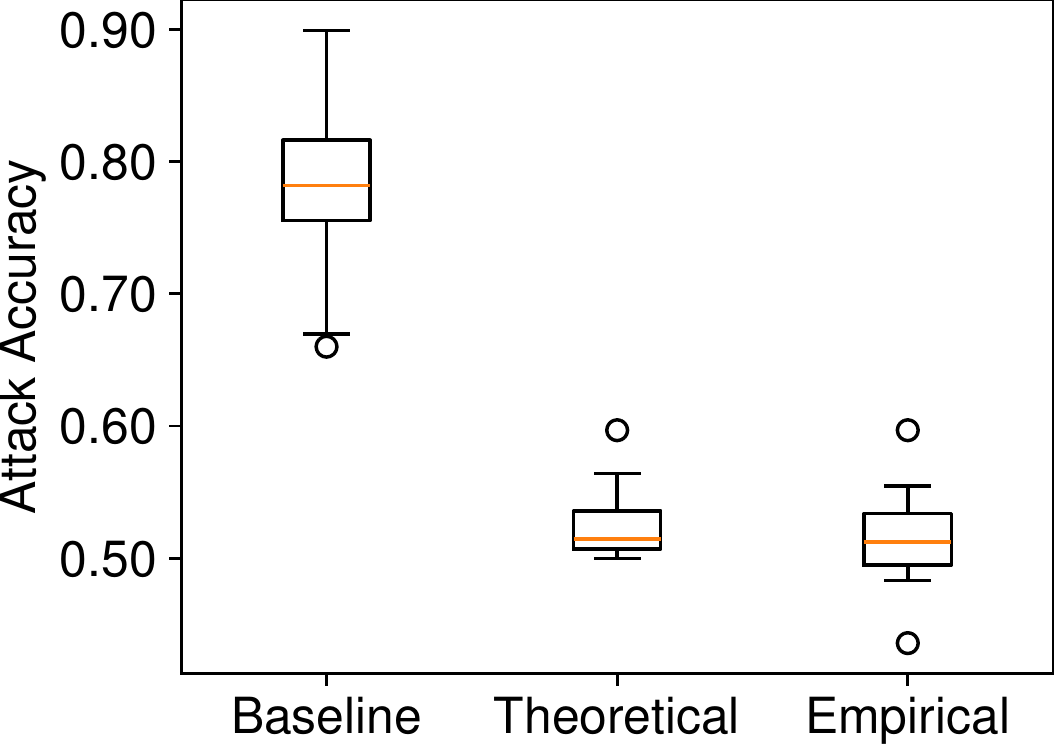}
    }
    \end{minipage}

    \caption{For \adaptiveAIAHard, we observe that \advdebias reduces the attack accuracy to random guess ($\sim$50\%).}
    \label{fig:AdaptAIAHardDebias2}
\end{figure}

\noindent\textbf{Additional Results: \adaptiveAIAHard.} Since, \advdebias reveals soft labels, we can convert them into hard labels to evaluate against \adaptiveAIAHard. We present eh evaluation for all the datasets in Figure~\ref{fig:AdaptAIAHardDebias2}. We observe that \advdebias can successfully lower the attack accuracy of \adaptiveAIAHard.
Additionally, we note that the theoretical bound on attack accuracy matches with the empirical attack accuracy.

\noindent\textbf{Additional Results: Impact on Utility.} We now present the impact on $\targetmodel$'s utility on using \advdebias on remaining datasets: \compas and \lfw. Similar to results in Section~\ref{sec:advdebias}, we observe a drop in accuracy in Figure~\ref{fig:utilityAdvDebias2} on using \advdebias which is a known trade-off observed in prior works.

\input{New/figures/fig_advdebias_utility}

\noindent\textbf{Additional Results: Fairness.} For \compas and \lfw, we measure \dempar-level with and without \advdebias (Figure~\ref{fig:DemParAdvDebias2}). Similar to results in Section~\ref{sec:advdebias}, we note that there is a decrease in \dempar-level on using \advdebias indicating that \advdebias is effective in achieving group fairness.

\input{New/figures/fig_advdebias_fair}

\section{\egdeo}
\label{app:egdeo}

%\noindent\textbf{\egddp vs. \egdeo.} Recall that in Section~\ref{sec:EGD}, we use \dempar as the fairness constraint. 
In this section, we theoretically show that \eo cannot mitigate \adaptiveAIAHard and by consequence \adaptiveAIASoft. 

\begin{theorem}
\label{th:eoo}
If $\hat{Y}$ satisfies \eo for $Y$ and $S$ then the balanced accuracy of \adaptiveAIAHard is $\frac{1}{2}$ iff $Y$ is independent of $S$ or $\hat{Y}$ is independent of $Y$.
\end{theorem}

\input{New/proofs/proof_egd_eo}

%We prove the theorem in Appendix~\ref{app:egdeo}.
Those two conditions are unlikely to happen in the real-world.
The condition of $Y$ being independent of $S$ was not observed for our datasets. We evaluate $|P(Y=0|S=0) - P(Y=0|S=1)|$ where a high value indicates $Y$ and $S$ are dependent. For \race and \sex, we found these values to be 0.05 and 0.27 (\compas), 0.20 and 0.13 (\census) and 0.07 and 0.13 (\meps) respectively.
Further, the independence between $\hat{Y}$ and $Y$ means that $\targetmodel$ has random guess utility. 
Hence, in practice, \eo aligns by reducing the risk to \aia{s} but does not \textit{perfectly align} as seen in \dempar by reducing \aia{s} to random guessing.
The choice of fairness metric is important for \egd for perfect alignment.

%We discussed in Section~\ref{sec:discussions} that \egdeo does not perfectly align with attribute privacy as it does not render \aia accuracy to random guess. We provide a proof for the stated theorem.

\section{Impact of \adv's $\auxdata$ Distribution}
\label{app:auxdist}

Recall from Section~\ref{sec:setup}, the distribution of $\auxdata$ is the same as $\traindata$. However, in practice, \adv can sample $\auxdata$ from a different distribution.
%(different ratio of $S$). 
To evaluate this, we consider \census but from different US states by using~\cite{folk}.
We evaluate the success of \adaptiveAIAHard and \adaptiveAIASoft against \egd, \advdebias and a baseline without any fairness specific algorithm for different samples of $\auxdata$.
We trained $\targetmodel$ on the \census data of Alabama and then we set $\auxdata$ to be every other state one after the other.
That way we obtain an attack balanced accuracy per state.

To evaluate the shift of distribution from Alabama' \census to the other state, we use the quality report provided by the Synthetic Data Vault~\cite{SDV}.
This score leverages multiple statistical metrics to evaluate how close two distributions are.
We then scale this score in $[0,1]$: a score of 1 indicates no shift while a score of 0 indicates the farther away state from Alabama in terms of distribution.

\begin{figure}
    \centering
    \subfigure[\baseline]{\includegraphics[width=0.69\linewidth]{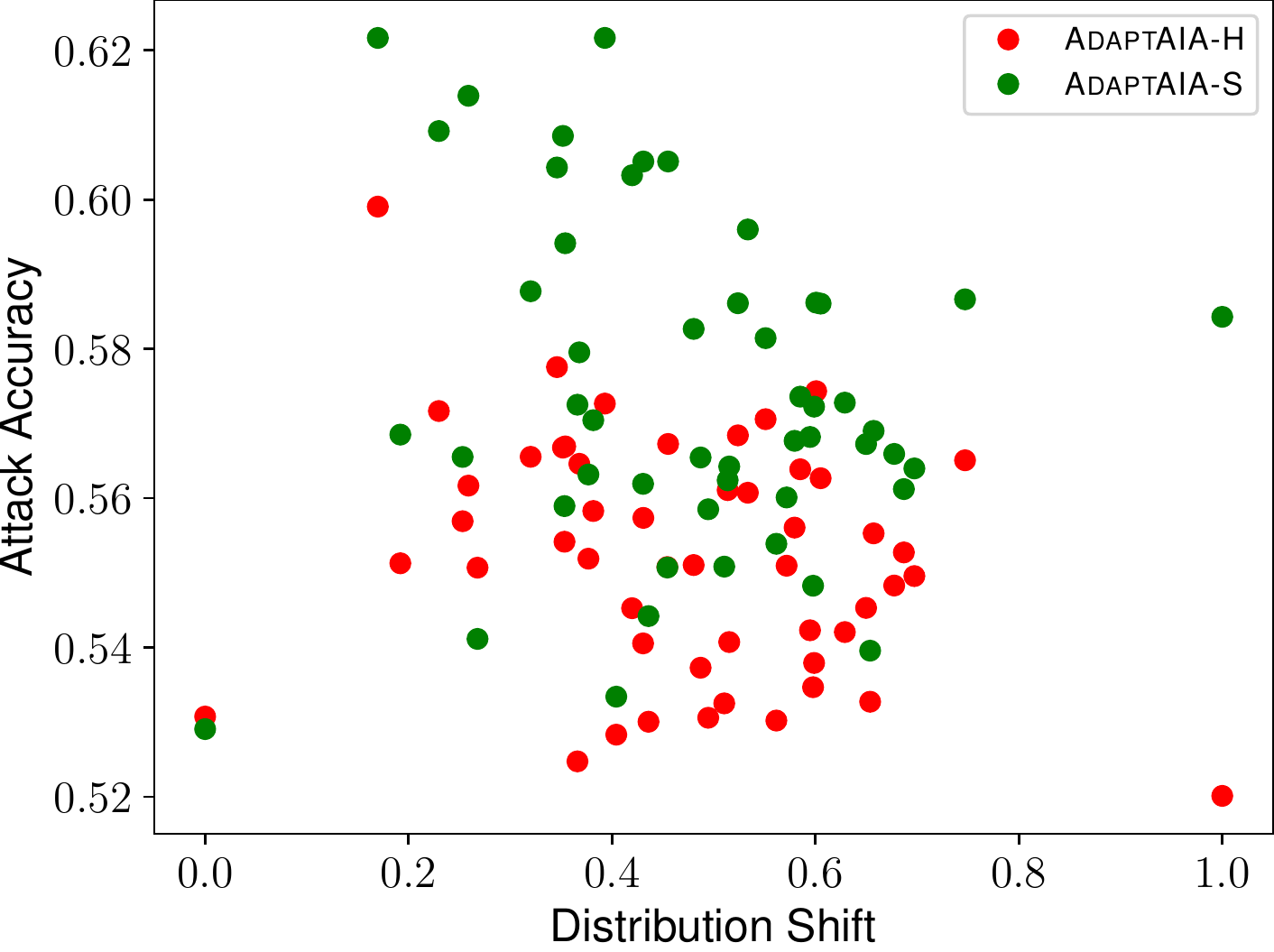}}
    
     \subfigure[\egddp]{\includegraphics[width=0.69\linewidth]{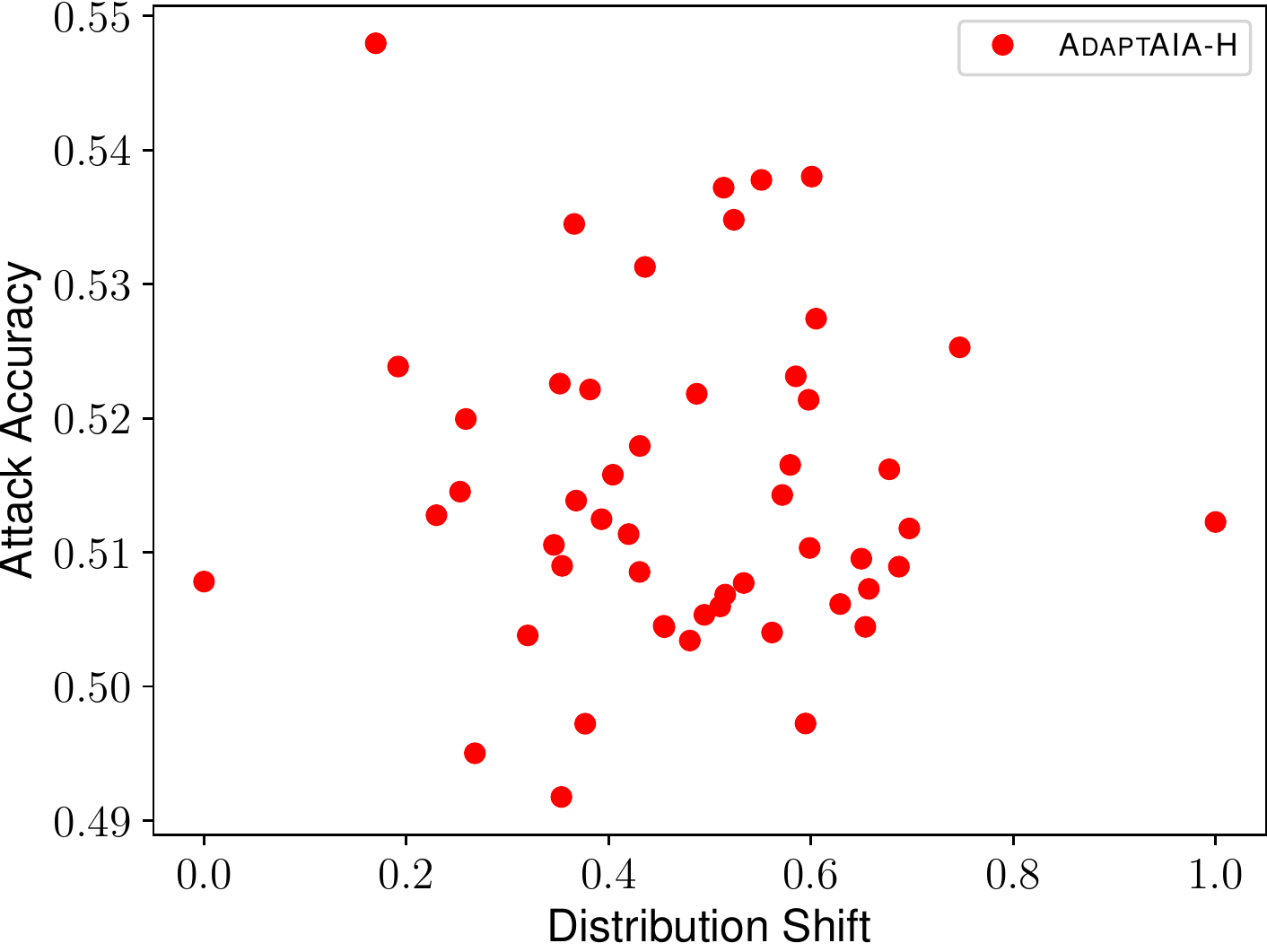}}
     
      \subfigure[\advdebias]{\includegraphics[width=0.69\linewidth]{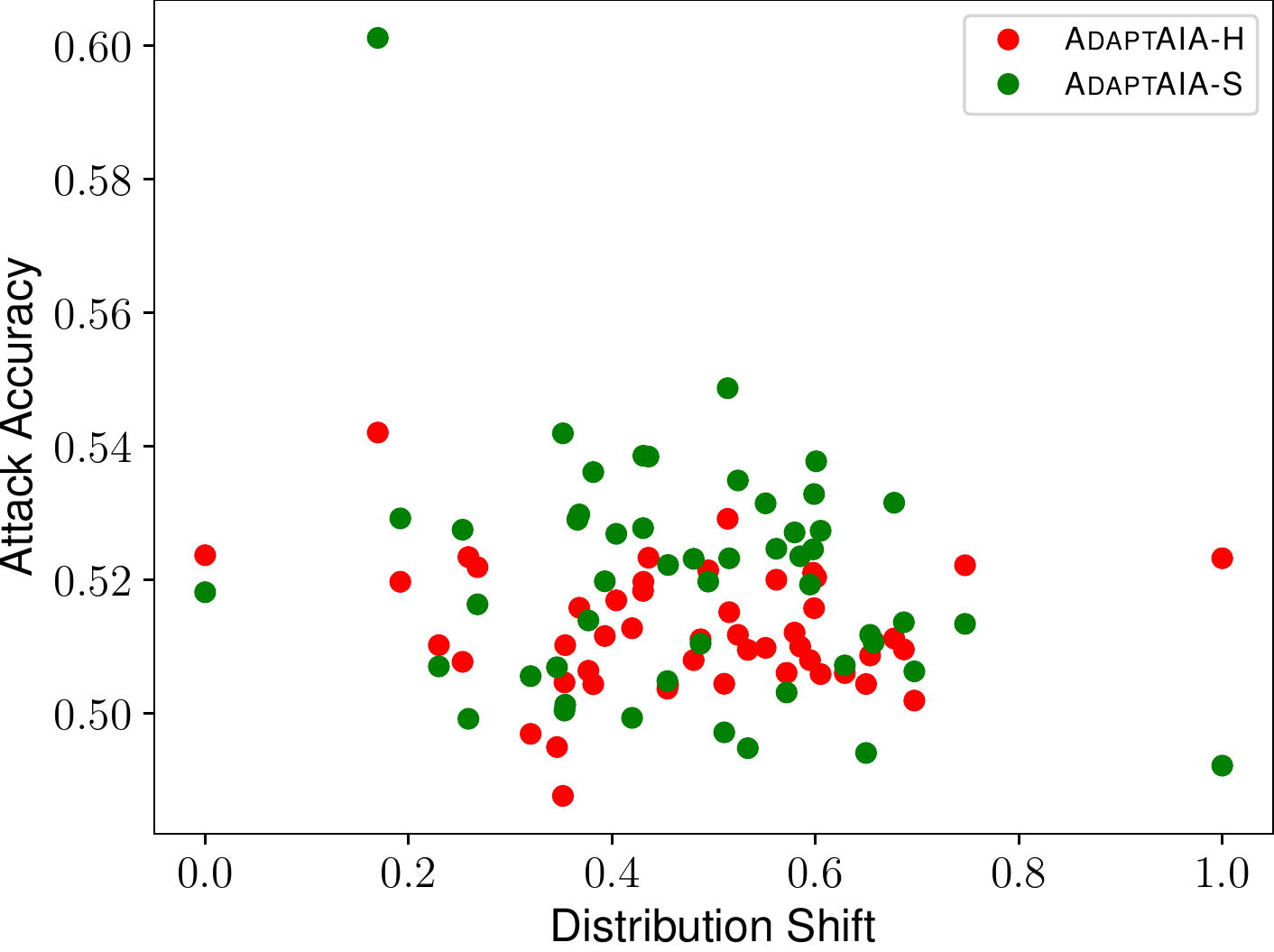}}
    \caption{Impact on attack success of distribution shift between the training set of the target model, $\traindata$, and the auxiliary knowledge used by the adversary to conduct the attack, $\auxdata$ (\census dataset).}
    %Impact of distribution shift on attack success on the \census dataset.}
    \label{fig:shift}
\end{figure}

We present our results in Figure~\ref{fig:shift}. As expected, we observe that \egd and \advdebias are still successful in mitigating \adaptiveAIAHard and \adaptiveAIASoft regardless of \adv's $\auxdata$ distribution.

%% file: New/proofs/proof_egd_dp.tex
\begin{customthm}{\ref{th:dpgood}}
The maximum balanced accuracy achievable by AH is equal to $\frac{1}{2}(1+\text{\dempar\_level of }\targetmodel)$ . 
\end{customthm}
\begin{proof}
The set $B$ of function from $\{0,1\}$ to $\{0,1\}$ contains four elements: $b_0=0$, $b_1=id$, $b_2=1-id$ and $b,3=1$, where $\forall x, id(x) = x$.
For every $b\in B$ the balanced \aia accuracy is 
$BA(b) = \frac{1}{2}(P(b\circ \hat{Y}=0|S=0) + P(b\circ \hat{Y}=1|S=1))$.
We have $BA(b_0) = BA(b_3) = \frac{1}{2}$, hence, we can discard those elements when solving the attack optimisation problem.
This problem writes $\text{max}_{b\in B}B(A(b)) = \text{max}(BA(b_1), BA(b_2))$.
We remark that $b_1\circ \hat{Y}=\hat{Y}$ and $b_2\circ \hat{Y}=1 - \hat{Y}$.
Hence,
{\footnotesize
\begin{align*}
    BA(b_1) &= \frac{1}{2}(P(\hat{Y}=0|S=0) + P(\hat{Y}=1|S=1))\\
    &=\frac{1}{2}(1+P(\hat{Y}=1|S=1) - P(\hat{Y}=1|S=0))\\
    BA(b_2)&=\frac{1}{2}(1+P(\hat{Y}=1|S=0) - P(\hat{Y}=1|S=1))
\end{align*}
}
Thus,
{\footnotesize
\begin{align*}
    &\text{max}_{b\in B}BA(b) \\
    = &\frac{1}{2}\left(1+\text{max}\left(
    \begin{matrix}
        P(\hat{Y}=0|S=0) -P(\hat{Y}=1|S=1)\\ 
        P(\hat{Y}=1|S=0) -P(\hat{Y}=0|S=1)
    \end{matrix}
    \right)\right)\\
    =&\frac{1}{2}(1+|P(\hat{Y}=1|S=1) - P(\hat{Y}=1|S=0)|)
\end{align*}
}
\end{proof}

%% file: New/figures/fig_egd_utility.tex
\begin{figure}[!htb]
    \centering
    \begin{minipage}[b]{1\linewidth}
    \centering
    \subfigure[\compas]{
    \includegraphics[width=0.49\linewidth]{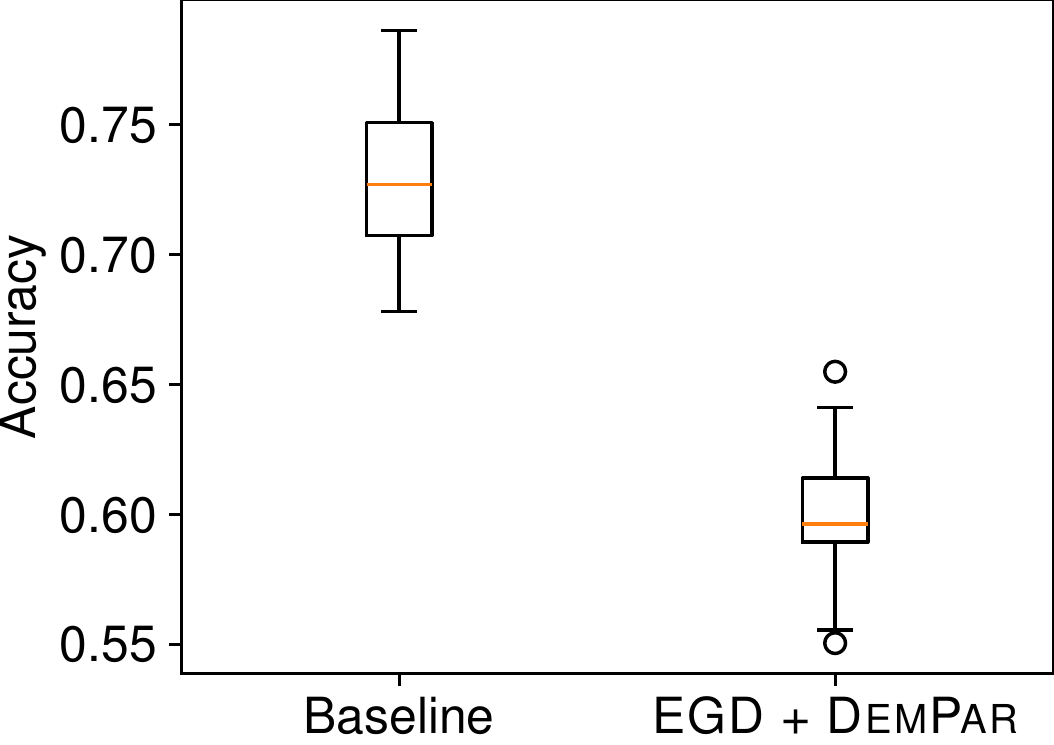}
    }%
    %\subfigure[\census]{
    %\includegraphics[width=0.49\linewidth]{New/figures/egd/census/census_egd_utility.pdf}
    %}%
    \subfigure[\lfw]{
    \includegraphics[width=0.49\linewidth]{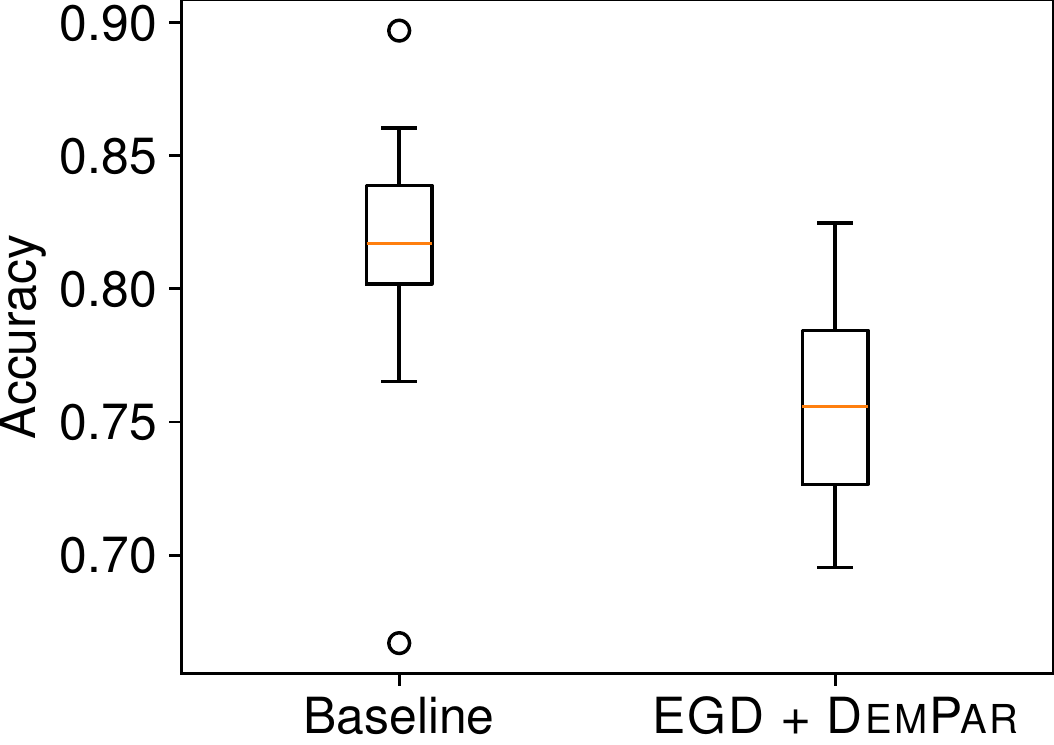}
    }
    %\subfigure[\compas]{
    %\includegraphics[width=0.49\linewidth]{New/figures/egd/compas/compas_egd_utility.pdf}
    %}
    \end{minipage}\\
    %\begin{minipage}[b]{1\linewidth}
    %\centering
    %\subfigure[\meps]{
    %\includegraphics[width=0.49\linewidth]{New/figures/egd/meps/meps_egd_utility.pdf}
    %}%
    %\subfigure[\lfw]{
    %\includegraphics[width=0.49\linewidth]{New/figures/egd/lfw/lfw_egd_utility.pdf}
    %}
    %\end{minipage}
    \caption{Utility degradation for \egd: We observe a statistically significant drop in $\targetmodel$'s accuracy on using \egddp which matches the observation from prior work~\cite{reductions}.}
    \label{fig:utilityEGD}
\end{figure}

%% file: New/figures/fig_egd_fair.tex
\begin{figure}[!htb]
    \centering
%    \begin{minipage}[b]{1\linewidth}
%    \centering
%    \subfigure[\census (\race)]{
%    \includegraphics[width=0.49\linewidth]{New/figures/egd/census/census_egd_dp_lvl_race.pdf}
%    }%
%    \subfigure[\census (\sex)]{
%    \includegraphics[width=0.49\linewidth]{New/figures/egd/census/census_egd_dp_lvl_sex.pdf}
%    }
%    \end{minipage}\\

    \begin{minipage}[b]{1\linewidth}
    \centering
    \subfigure[\compas (\race)]{
    \includegraphics[width=0.49\linewidth]{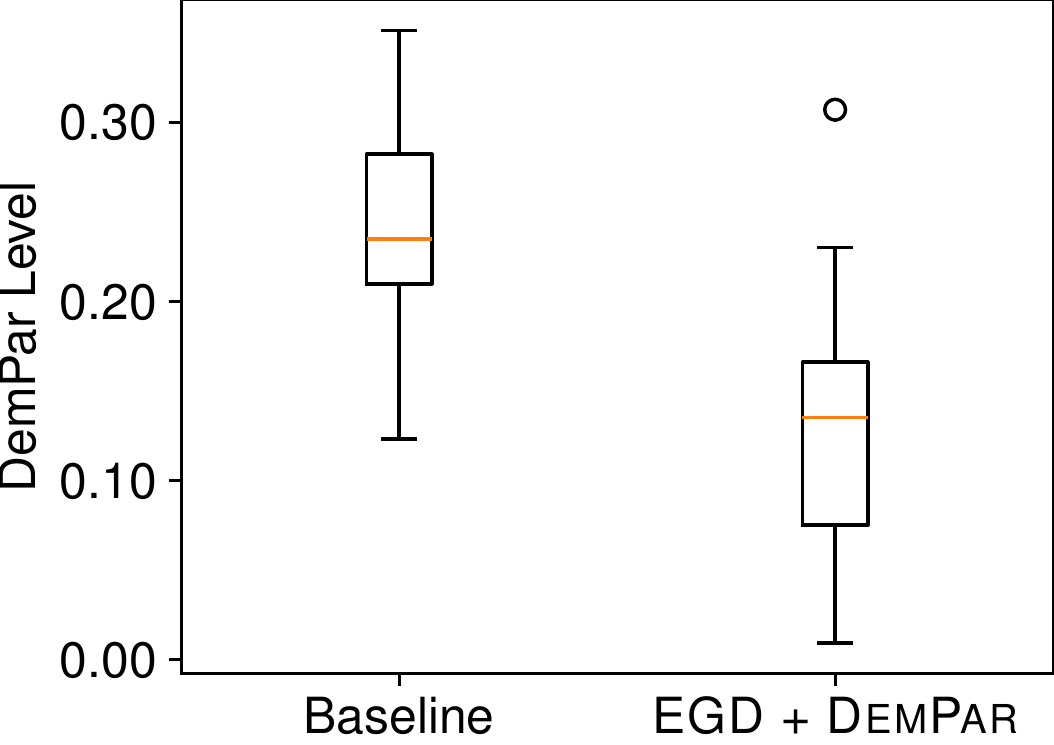}
    }%
    \subfigure[\compas (\sex)]{
    \includegraphics[width=0.49\linewidth]{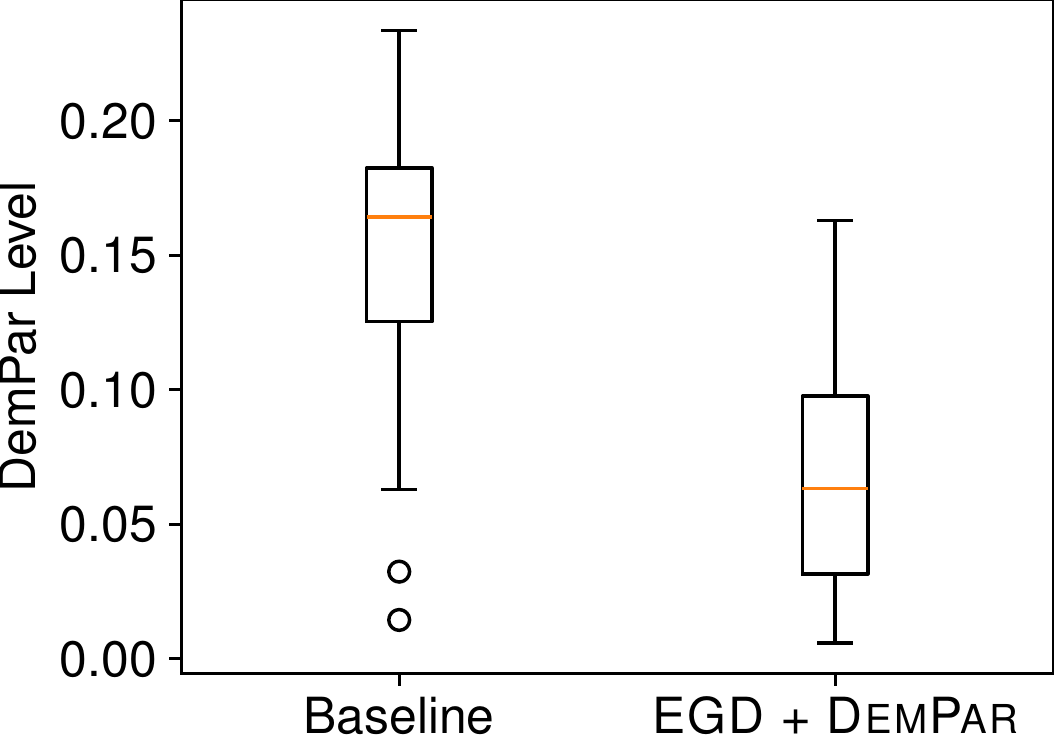}
    }
    \end{minipage}\\

%    \begin{minipage}[b]{1\linewidth}
%    \centering
%    \subfigure[\meps (\race)]{
%    \includegraphics[width=0.49\linewidth]{New/figures/egd/meps/meps_egd_dp_lvl_race.pdf}
%    }%
%    \subfigure[\meps (\sex)]{
%    \includegraphics[width=0.49\linewidth]{New/figures/egd/meps/meps_egd_dp_lvl_sex.pdf}
%    }
%    \end{minipage}\\

    \begin{minipage}[b]{1\linewidth}
    \centering
    \subfigure[\lfw (\race)]{
    \includegraphics[width=0.49\linewidth]{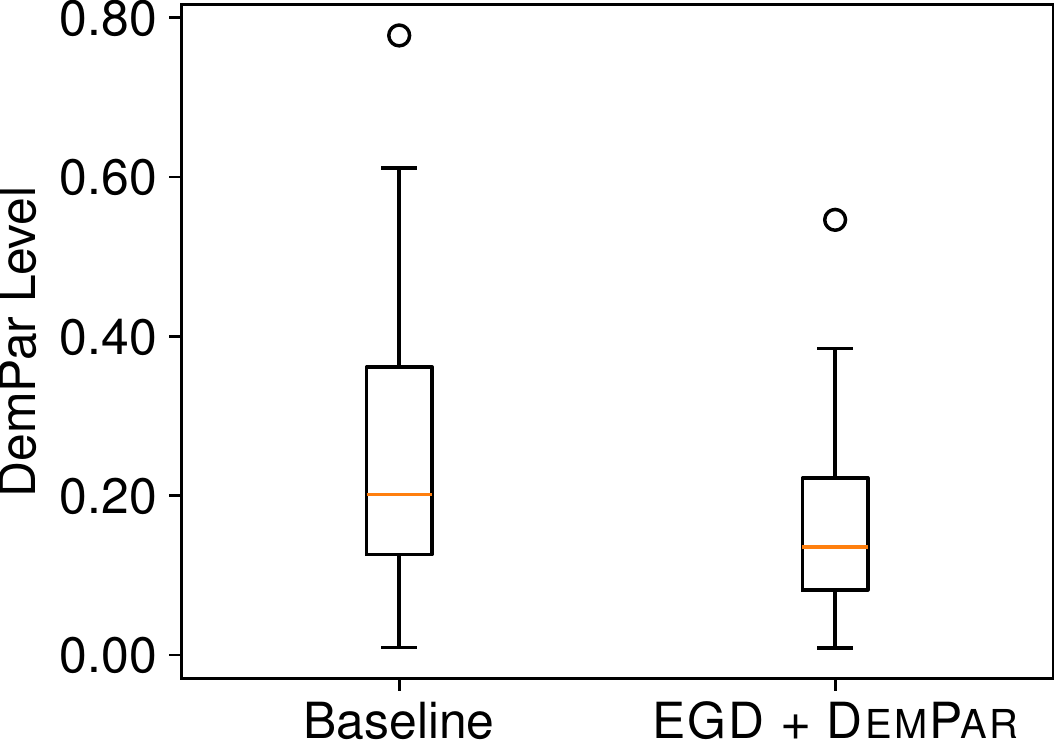}
    }%
    \subfigure[\lfw (\sex)]{
    \includegraphics[width=0.49\linewidth]{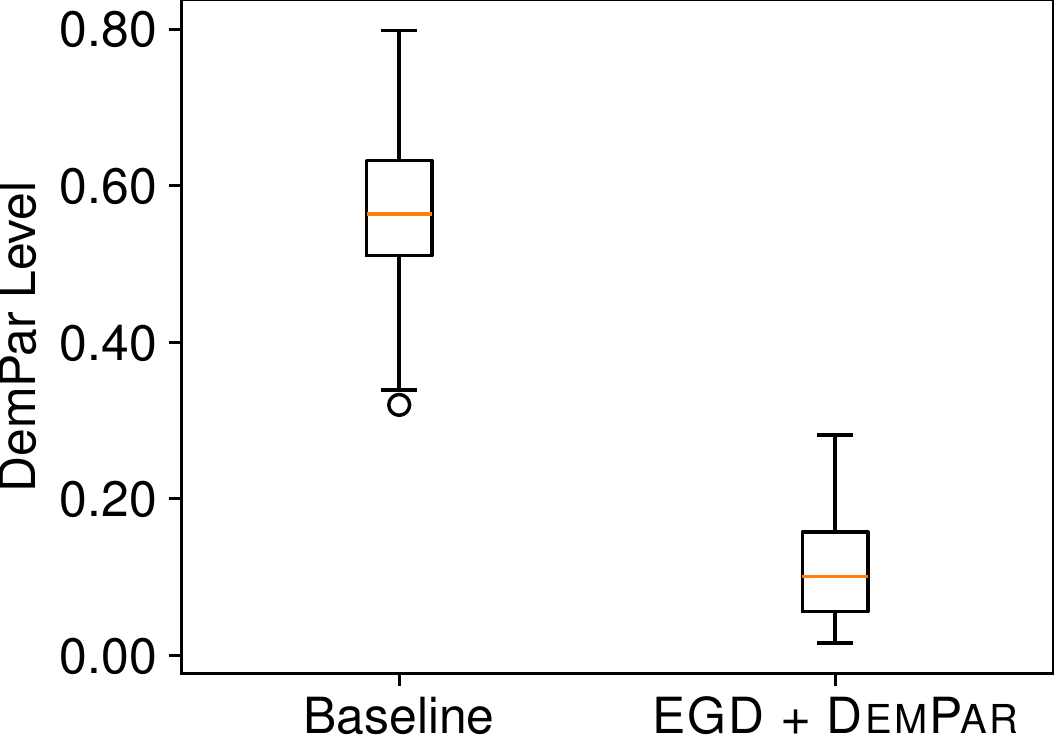}
    }
    \end{minipage}

    \caption{\dempar-Level for \egd: We observe that \dempar-Level is lower for \egd than the baseline indicating $\targetmodel$ is fair after \egd.}
    \label{fig:DemParegd2}
\end{figure}

%% file: New/proofs/proof_advdebias.tex
We remark that the definition~\ref{def:dp} of \dempar can be written synthetically as the following property:
$P_{\hat{Y},S}=P_{\hat{Y}}\otimes P_{S}$.
Where $P_{\hat{Y}}\otimes P_{S}$ is the product measure defined as the unique measure on 
$\mathcal{P}(\mathcal{Y})\times\mathcal{P}(\mathcal{S})$ such that
$\forall y\in\mathcal{P}(\mathcal{Y})\forall s\in\mathcal{P}(\mathcal{S})\quad P_{\hat{Y}}\otimes P_{S}(y\times s) = P_{\hat{Y}}(y)P_{S}(s)$.
This definition of \dempar is equivalent to definition \ref{def:dp} for binary labels and sensitive attribute but more general because when $\hat{Y}$ is not binary as in soft labels, this new definition is well defined.
We write formally 
\begin{definition}
\label{def:dps}
    $\hat{Y}$ satisfies extended \dempar for $S$ if and only if: $P_{\hat{Y},S}=P_{\hat{Y}}\otimes P_{S}$.
\end{definition}
We remark that we can not derive a quantity similar to \dempar-level with this definition but this extended \dempar assures indistinguishably of the sensitive attribute when looking at the soft labels.
We have the following theorem:
\begin{customthm}{\ref{th:advdebias}}
    The following propositions are equivalent 
    \begin{itemize}
        \item $\hat{Y}$ satisfies extended \dempar for $S$
        \item The balanced accuracy of \adaptiveAIASoft is $\frac{1}{2}$
    \end{itemize}
\end{customthm}
\begin{proof}
    \begin{align*}
            &\forall a~P(\hat{Y}\in a^{-1}(\{0\})|S=0)+P(\hat{Y}\in a^{-1}(\{1\})|S=1) = 1\\
            \Leftrightarrow&\forall a~P(\hat{Y}\in a^{-1}(\{0\})|S=0)=P(\hat{Y}\in a^{-1}(\{0\})|S=1)\\
            \Leftrightarrow&\forall A~P(\hat{Y}\in A)|S=0)=P(\hat{Y}\in A|S=1) \\
            \Leftrightarrow &P_{\hat{Y},S}=P_{\hat{Y}}\otimes P_{S}
        \end{align*}
\end{proof}
Adversarial Debiasing aims to imposing $P_{\hat{Y},S}=P_{\hat{Y}}\otimes P_{S}$ hence Adversarial Debiasing mitigates \adaptiveAIASoft.
%In conclusion, with extended \dempar we can not unconditionally bound the balanced accuracy of the attack without introducing distances in the space of distributions, but it gives us a condition to protect the sensitive attribute in case of an adversary gaining access to soft labels (AS). 

%% file: New/figures/fig_advdebias_utility.tex
\begin{figure}[!htb]
    \centering
    \begin{minipage}[b]{1\linewidth}
    \centering
    \subfigure[\compas]{
    \includegraphics[width=0.49\linewidth]{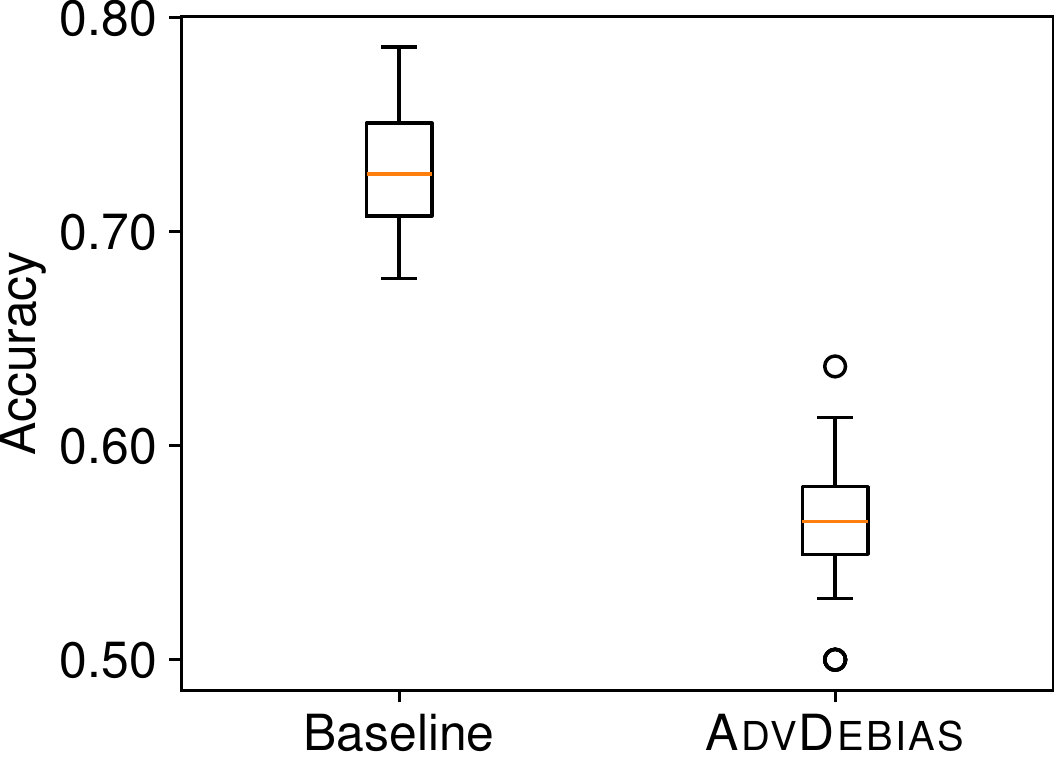}
    }%
    \subfigure[\lfw]{
    \includegraphics[width=0.49\linewidth]{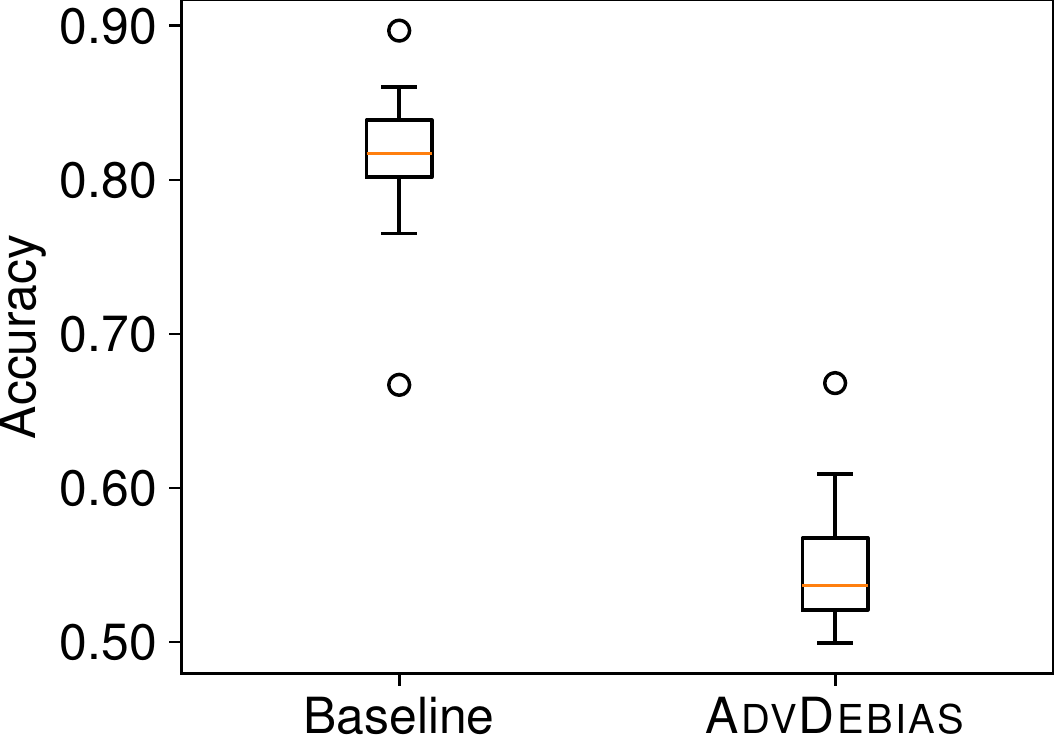}
    }
    \end{minipage}\\
%    \begin{minipage}[b]{1\linewidth}
%    \centering
%    \subfigure[\meps]{
%    \includegraphics[width=0.49\linewidth]{New/figures/advdebias/meps/meps_advdeb_utility.pdf}
%    }%
%    \subfigure[\lfw]{
%    \includegraphics[width=0.49\linewidth]{New/figures/advdebias/lfw/lfw_advdeb_utility.pdf}
%    }
%    \end{minipage}
    \caption{Utility degradation for \advdebias: We observe a statistically significant drop in $\targetmodel$'s accuracy on using \advdebias.}
    \label{fig:utilityAdvDebias2}
\end{figure}

%% file: New/figures/fig_advdebias_fair.tex
\begin{figure}[!htb]
    \centering
    \begin{minipage}[b]{1\linewidth}
    \centering
    \subfigure[\compas (\race)]{
    \includegraphics[width=0.49\linewidth]{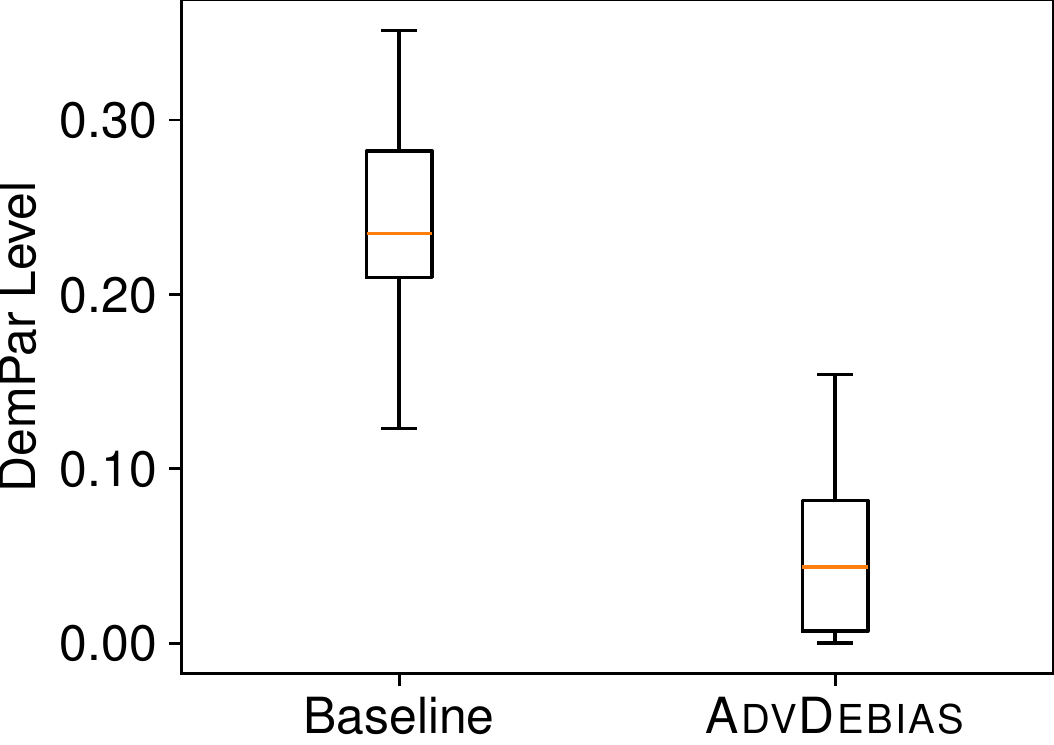}
    }%
    \subfigure[\compas (\sex)]{
    \includegraphics[width=0.49\linewidth]{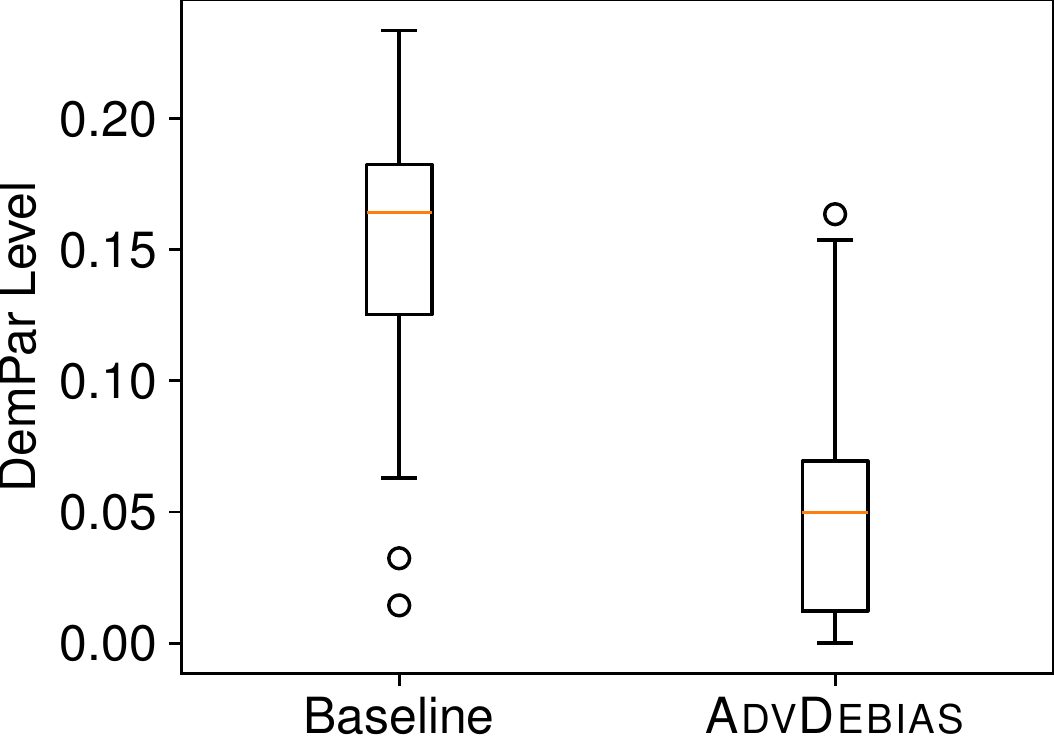}
    }
    \end{minipage}\\

    \begin{minipage}[b]{1\linewidth}
    \centering
    \subfigure[\lfw (\race)]{
    \includegraphics[width=0.49\linewidth]{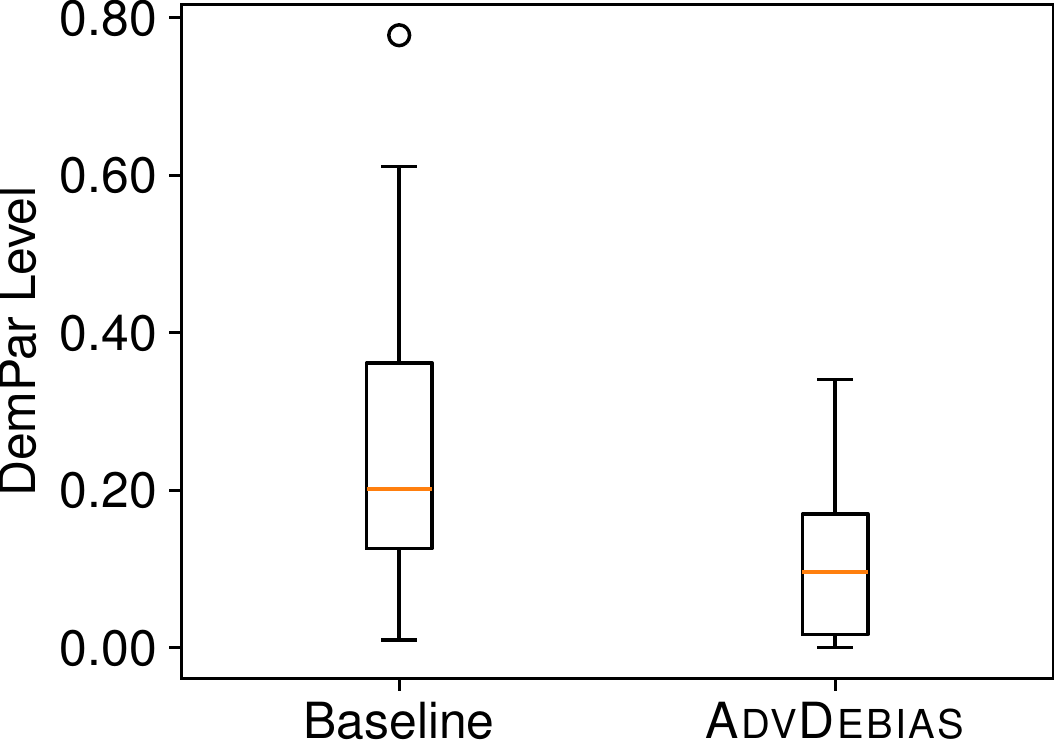}
    }%
    \subfigure[\lfw (\sex)]{
    \includegraphics[width=0.49\linewidth]{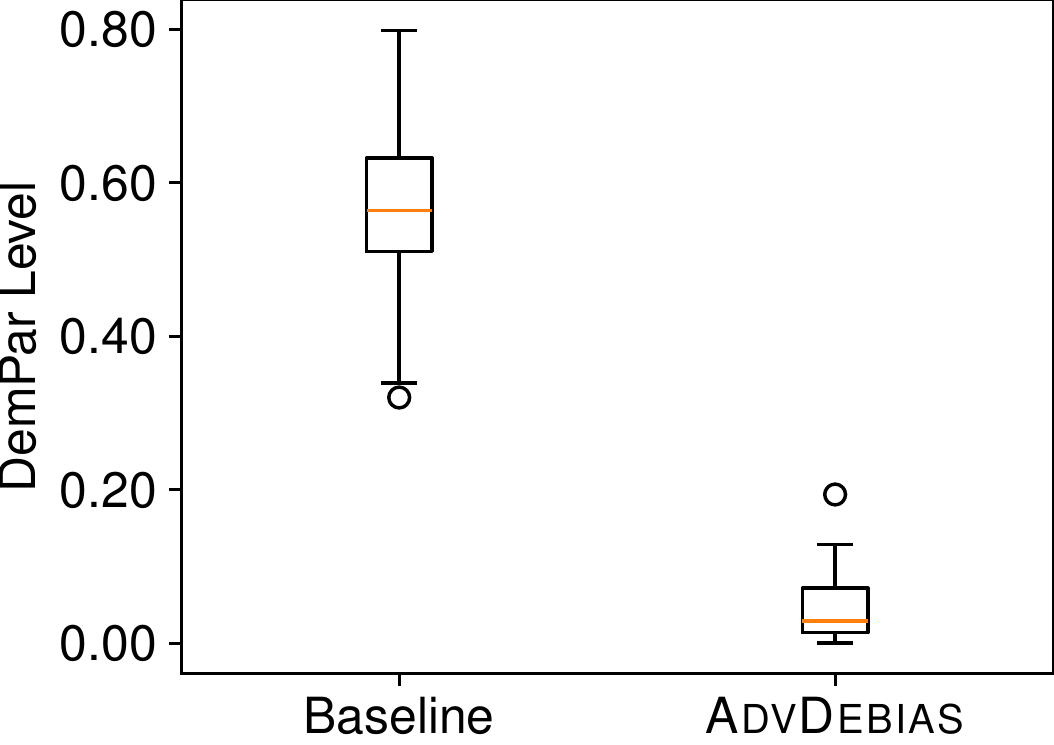}
    }
    \end{minipage}

    \caption{\dempar-Level for \advdebias: We observe that \dempar-Level is lower for \advdebias indicating $\targetmodel$ is fair.}
    \label{fig:DemParAdvDebias2}
\end{figure}

%% file: New/proofs/proof_egd_eo.tex
\begin{customthm}{\ref{th:eoo}}
If $\hat{Y}$ satisfies \eo for $Y$ and $S$ then the balanced accuracy of \adaptiveAIAHard is $\frac{1}{2}$ if and only if $Y$ is independent of $S$ or $\hat{Y}$ is independent of $Y$.
\end{customthm}
\begin{proof}
Let $\attackmodel$ be the attack model trained for AS: $\hat{S}=a\circ \hat{Y}$.
By the total probability formula
{\footnotesize
\begin{align*}
P(\hat{S}=0|S=0)=&P(\hat{S}=0|S=0Y=0)P(Y=0|S=0)\\
+&P(\hat{S}=0|S=0Y=1)P(Y=1|S=0)
\end{align*}
}
and as well
{\footnotesize
\begin{align*}
P(\hat{S}=1|S=1)=&P(\hat{S}=1|S=1Y=0)P(Y=0|S=1)\\
        +&P(\hat{S}=1|S=1Y=1)P(Y=1|S=1)
\end{align*}
}
Then we substitute those terms in the definition of the balanced accuracy of the target model.
\begin{align*}
        &\frac{P(\hat{S}=0|S=0)+P(\hat{S}=1|S=1)}{2}\\
        =&\frac{1}{2}+\frac{1}{2}\left(P(Y=0|S=0)-P(Y=0|S=1)\right)\\
        &\left(P(\hat{Y}\in \attackmodel^{-1}(\{1\})|S=1Y=0) -
        P(\hat{Y}\in \attackmodel^{-1}(\{1\})|S=1Y=1)\right)
\end{align*}
The balanced accuracy is equal to 0.5 if and only if $P(Y=0|S=0)=P(Y=0|S=1)$
or $\forall \attackmodel~P(\hat{Y}\in \attackmodel^{-1}(\{1\})|S=1Y=0)=P(\hat{Y}\in \attackmodel^{-1}(\{1\})|S=1Y=1)$.
The first term indicates that $Y$ is independent of $S$ and the second term indicates that $S=1$ the $\targetmodel$ random guess utility.
We can do the same computing for $S=0$ and obtain a similar conclusion. 
\end{proof}